\PassOptionsToPackage{unicode}{hyperref}
\PassOptionsToPackage{hyphens}{url}
\PassOptionsToPackage{dvipsnames,svgnames,x11names}{xcolor}
\documentclass[
  12pt]{article}

\usepackage{subfigure}
\usepackage{setspace}

\usepackage{amsmath, amssymb, amsthm}
\usepackage{iftex}
\ifPDFTeX
  \usepackage[T1]{fontenc}
  \usepackage[utf8]{inputenc}
  \usepackage{textcomp} 
\else 
  \usepackage{unicode-math}
  \defaultfontfeatures{Scale=MatchLowercase}
  \defaultfontfeatures[\rmfamily]{Ligatures=TeX,Scale=1}
\fi
\usepackage{lmodern}
\ifPDFTeX\else  
\fi
\IfFileExists{upquote.sty}{\usepackage{upquote}}{}
\IfFileExists{microtype.sty}{
  \usepackage[]{microtype}
  \UseMicrotypeSet[protrusion]{basicmath} 
}{}
\makeatletter
\@ifundefined{KOMAClassName}{
  \IfFileExists{parskip.sty}{%
    \usepackage{parskip}
  }{
    \setlength{\parindent}{0pt}
    \setlength{\parskip}{6pt plus 2pt minus 1pt}}
}{
  \KOMAoptions{parskip=half}}
\makeatother
\usepackage{xcolor}
\makeatletter
\ifx\paragraph\undefined\else
  \let\oldparagraph\paragraph
  \renewcommand{\paragraph}{
    \@ifstar
      \xxxParagraphStar
      \xxxParagraphNoStar
  }
  \newcommand{\xxxParagraphStar}[1]{\oldparagraph*{#1}\mbox{}}
  \newcommand{\xxxParagraphNoStar}[1]{\oldparagraph{#1}\mbox{}}
\fi
\ifx\subparagraph\undefined\else
  \let\oldsubparagraph\subparagraph
  \renewcommand{\subparagraph}{
    \@ifstar
      \xxxSubParagraphStar
      \xxxSubParagraphNoStar
  }
  \newcommand{\xxxSubParagraphStar}[1]{\oldsubparagraph*{#1}\mbox{}}
  \newcommand{\xxxSubParagraphNoStar}[1]{\oldsubparagraph{#1}\mbox{}}
\fi
\makeatother

\usepackage{longtable,booktabs,array}
\usepackage{calc} 
\usepackage{etoolbox}
\makeatletter
\patchcmd\longtable{\par}{\if@noskipsec\mbox{}\fi\par}{}{}
\makeatother
\IfFileExists{footnotehyper.sty}{\usepackage{footnotehyper}}{\usepackage{footnote}}
\makesavenoteenv{longtable}
\usepackage{graphicx}
\makeatletter
\def\maxwidth{\ifdim\Gin@nat@width>\linewidth\linewidth\else\Gin@nat@width\fi}
\def\maxheight{\ifdim\Gin@nat@height>\textheight\textheight\else\Gin@nat@height\fi}
\makeatother
\setkeys{Gin}{width=\maxwidth,height=\maxheight,keepaspectratio}
\makeatletter
\def\fps@figure{htbp}
\makeatother

\makeatletter
\@ifpackageloaded{caption}{}{\usepackage{caption}}
\AtBeginDocument{%
\ifdefined\contentsname
  \renewcommand*\contentsname{Table of contents}
\else
  \newcommand\contentsname{Table of contents}
\fi
\ifdefined\listfigurename
  \renewcommand*\listfigurename{List of Figures}
\else
  \newcommand\listfigurename{List of Figures}
\fi
\ifdefined\listtablename
  \renewcommand*\listtablename{List of Tables}
\else
  \newcommand\listtablename{List of Tables}
\fi
\ifdefined\figurename
  \renewcommand*\figurename{Figure}
\else
  \newcommand\figurename{Figure}
\fi
\ifdefined\tablename
  \renewcommand*\tablename{Table}
\else
  \newcommand\tablename{Table}
\fi
}
\@ifpackageloaded{float}{}{\usepackage{float}}
\floatstyle{ruled}
\@ifundefined{c@chapter}{\newfloat{codelisting}{h}{lop}}{\newfloat{codelisting}{h}{lop}[chapter]}
\floatname{codelisting}{Listing}

\makeatother
\makeatletter
\@ifpackageloaded{caption}{}{\usepackage{caption}}
\@ifpackageloaded{subcaption}{}{\usepackage{subcaption}}
\makeatother

\ifLuaTeX
  \usepackage{selnolig}  
\fi
\usepackage[]{natbib}
\usepackage{bookmark}

\IfFileExists{xurl.sty}{\usepackage{xurl}}{} 
\hypersetup{
  pdftitle={Online simultaneous inference for quantiles via smoothed stochastic gradient descent},
  pdfauthor={Likai Chen; Georg Keilbar; Wei Biao Wu},
  pdfkeywords={Online inference; simultaneous confidence bands; multiplier bootstrap; Bahadur representation; quantile crossing.},
  colorlinks=true,
  linkcolor={blue},
  filecolor={Maroon},
  citecolor={Blue},
  urlcolor={Blue},
  pdfcreator={LaTeX via pandoc}}

\def\A{\mathcal A}

\def\G{\mathcal G}
\def\F{\mathcal F}

\def\B{\mathcal {B}}

\def\I{\mathrm{I}}

\def\EE{\mathbb E}
\def\PP{\mathbb P}
\def\RR{\mathbb{R}}

\def\d{\mathrm{d}}
\def\NN{\mathbb{N}}

\def\1{\textbf{1}}

\def\argmin{\mathrm{argmin} }
\def\var{\mathrm{Var}}
\def\cov{\mathrm{Cov}}

\newtheorem{assumption}{\sc Assumption}
\newtheorem{theorem}{Theorem}[section]
\newtheorem{corollary}[theorem]{Corollary}
\newtheorem{lemma}[theorem]{Lemma}
\newtheorem{proposition}[theorem]{Proposition}
\newtheorem{remark}{\sc Remark}

\newcommand{\anon}{1}

\begin{document}

\def\spacingset#1{\renewcommand{\baselinestretch}%
{#1}\small\normalsize} \spacingset{1}


\if1\anon
{
  \title{\bf Online simultaneous inference for quantiles via smoothed stochastic gradient descent}
  \author{Likai Chen\thanks{Likai Chen is supported by the U.S. National Science Foundation under Awards 2515927 and 2222403.}\hspace{.2cm}\\
    Department of Statistics and Data Science, Washington University in St.\ Louis\\
    and \\
    Georg Keilbar\thanks{
    Georg Keilbar is funded by the Deutsche Forschungsgemeinschaft (DFG, German Research Foundation) - Project number 513634041.} \\
    Chair of Statistics, Humboldt-Universit\"at zu Berlin\\
    and \\
    Wei Biao Wu\thanks{Wei Biao Wu is supported by the U.S. National Science Foundation under Awards 2311249.} \\
    Department of Statistics, University of Chicago}
  \maketitle
} \fi

\if0\anon
{
  \bigskip
  \bigskip
  \bigskip
  \begin{center}
    {\LARGE\bf Online simultaneous inference for quantiles via smoothed stochastic gradient descent}
\end{center}
  \medskip
} \fi

\bigskip
\begin{abstract}
This paper considers the estimation of quantiles via a smoothed version of the stochastic gradient descent (SGD) algorithm. By smoothing the score function with a bandwidth tied to the learning rate, we obtain estimates that are monotone in the quantile level at every iteration, while retaining the memory and computational efficiency required for streaming data. We establish non-asymptotic tail probability bounds for the smoothed estimate with and without Polyak-Ruppert averaging, which are sub-exponential with a multi-regime structure. For the averaged estimate we further derive a Bahadur representation that is uniform in the quantile level and across coordinates, and a resulting Gaussian approximation by the maximum of Brownian bridges, with the dimension $p$ allowed to grow exponentially in the sample size. This yields simultaneous inference across coordinates and quantile levels. As an alternative that avoids estimating the sparsity function, we propose an online multiplier bootstrap that preserves monotonicity, runs in a single pass and is asymptotically valid. Extending the theory to a localized recursion, we obtain online nonparametric conditional quantile estimates with uniform bands over design points and quantile levels. Simulations confirm accurate finite-sample coverage, and we illustrate the method on conditional value-at-risk curves.
\end{abstract}

\noindent%
\vfill

\newpage
\spacingset{1.8} 

\section{Introduction}\label{sec:introduction}

Modern data often arrive sequentially and at a scale at which storing the full sample is infeasible and repeated recomputation is prohibitively expensive. The stochastic gradient descent (SGD) algorithm \citep{robbins1951stochastic, kiefer1952stochastic} addresses both constraints by updating the parameter recursively along the gradient of the objective function, using constant memory and a fixed number of operations per observation \citep{KushnerYin2003, moulines2011non}. \citet{ruppert1988efficient} and \citet{polyak1992acceleration} show that averaging the iterates yields an asymptotically normal estimate at the optimal rate. Quantiles are an important example in which these constraints bind. \citet{munro1980selection} showed that exact computation of the sample quantile requires $\Omega(n)$ memory, which has motivated a large literature on approximate quantile computation for streaming data \citep{greenwald2001space, cormode2021relative}. In many streaming applications the target is not a single quantile but entire quantile curves of a large number $p$ of series at once, such as return distributions across hundreds of assets or latency across servers.

A major drawback of the classical SGD quantile algorithm is that the resulting estimate is not monotone in the quantile level, which is a consequence of the non-smoothness of the piecewise-constant quantile score function. Quantile crossing is a long-standing concern \citep{Koenker2005}, and the available solutions are either based on restrictions \citep{he1997quantile} or on rearrangement of already computed quantiles \citep{chernozhukov2010quantile}, which is incompatible with a single-pass algorithm. In this paper, we propose a modified, smoothed SGD recursion in which the indicator in the update is replaced by a Lipschitz continuous approximation whose smoothing parameter shrinks with the iteration. The resulting estimate is monotone in the quantile level by construction, in every iteration and for every sample size, while retaining the memory and computational efficiency of the original algorithm. The construction is closely connected to the convolution-type smoothed quantile regression (\emph{conquer}) method of \citet{fernandes2021smoothing} and \citet{he2023smoothed}. However, instead of fixing a single smoothing bandwidth for the entire sample, we tie the bandwidth to the learning rate so that the smoothing vanishes automatically as the recursion proceeds.

We first establish non-asymptotic tail probability bounds for the smoothed recursion with and without Polyak-Ruppert averaging, sharpening and generalizing the bounds of \citet{chen2023recursive} for the unsmoothed algorithm. The bounds are sub-exponential with a multi-regime structure and hold uniformly in the quantile level and across the $p$ coordinates. The monotonicity of the smoothed SGD recursion is the key factor that allows us to pass from a grid of quantile levels to the supremum over an interval. We then derive a Bahadur representation for the averaged estimate that is uniform in the quantile level and across coordinates, together with a Gaussian approximation. The maximum of the normalized process over coordinates and quantile levels is approximated by the maximum of Brownian bridges, with the dimension $p$ allowed to grow exponentially in the sample size. The limit is the same as for the batch empirical quantile process, while using constant memory per coordinate and quantile level. Critical values obtained from maxima of Brownian bridges yield confidence bands that hold simultaneously over all coordinates and quantile levels, as well as tests for hypotheses on entire quantile curves.

Inference based on the Gaussian approximation requires the estimation of the density evaluated at the quantile, as well as estimation of the covariance structure of the limiting process. As an alternative method for simultaneous inference, we propose an online multiplier bootstrap that perturbs the updating step by independent multipliers with mean and variance one. It is density-free, so that neither the critical value nor the band require estimating any densities, and it reproduces the dependence across coordinates automatically. Since the multipliers are bounded, the perturbed recursion again preserves monotonicity, so that the bootstrap quantile curves do not cross, and the replications can be updated jointly with the original recursion in a single online pass. We show the validity of the bootstrap procedure. A related online bootstrap was proposed and studied by \citet{fang2018online}, but its analysis requires convex, twice continuously differentiable losses, which excludes the quantile loss.

Finally, we extend the theory to conditional quantiles. Instead of imposing a linear specification, we consider a nonparametric local constant version of the recursion, in which each design point carries its own trajectory that is updated only by those observations whose covariate falls into the corresponding window. The learning rate is indexed by the number of active observations. The effective sample sizes are therefore random and specific to the design point, which is the main new technical difficulty. Monotonicity continues to hold, and the resulting Gaussian approximation is simultaneous over design points and quantile levels, delivering uniform confidence bands for entire conditional quantile curves in a single online pass. We illustrate the procedure on daily S\&P~500 returns with the lagged volatility index as covariate, so that the estimated conditional quantiles are one-day-ahead value-at-risk curves across volatility regimes. The simultaneous bands let us reject a location-scale specification of the return distribution jointly across all volatility states and quantile levels.

Our work relates to a growing literature on online inference for SGD-based quantile estimation. Most closely, \citet{wei2026central} establish a central limit theorem for the last iterate of the constant learning-rate SGD quantile recursion via a Markov-chain argument. In contrast, we study the decaying-rate averaged scheme and obtain a Bahadur representation and Gaussian approximation that are uniform in the quantile level and across coordinates. Online bootstrap procedures that perturb the SGD update by independent multipliers have been proposed by \citet{fang2018online} and, for nonconvex objectives, by \citet{zhong2023online}, while \citet{zhu2024high} exploit parallel stochastic optimization for high-confidence inference. Our online multiplier bootstrap is of the same type but preserves monotonicity and targets inference simultaneously uniform in $i$ and $\tau$. \citet{samsonov2024gaussian}, \citet{sheshukova2025gaussian} and \citet{wei2026gaussian} recently derived Gaussian approximation and concentration results, together with multiplier-bootstrap validity, for averaged linear stochastic approximation and for SGD with decaying and constant learning rates, and \citet{li2026statistical} established high-probability guarantees for high-dimensional SGD. Alternatively, online inference can be based on the limiting distribution together with a recursive covariance estimate \citep{chen2020statistical, zhu2023online}.

The main contributions are threefold. Methodologically, we show that a suitably smoothed recursion makes monotonicity in the quantile level automatic at every stage of the algorithm, at no cost in memory or computation. Theoretically, we move from pointwise to simultaneous inference for stochastic approximation and provide two complementary routes to critical values. The first is directly based on the Gaussian approximation, whereas the second is based on an online multiplier bootstrap, which is free of nuisance estimation at the price of running additional recursions in parallel. Both routes are shown to have accurate finite-sample coverage in simulations. Finally, we extend both the algorithm and the inference to nonparametric conditional quantiles, where the effective sample sizes are random and specific to each design point.

Section~\ref{sec:smoothed} introduces the smoothed SGD algorithm and shows its monotonicity, Sections~\ref{sec:tailprobability} and \ref{sec:bahadur} develop the tail bounds, the Bahadur representation and the Gaussian approximation, Section~\ref{sec:bootstrap} the online multiplier bootstrap and Section~\ref{sec:conditional} the conditional quantile case. Section~\ref{sec:numerical} reports a Monte Carlo study and an empirical illustration, and Section~\ref{sec:conclusion} concludes. All proofs and additional numerical results are given in the Supplementary Material.

\emph{Notation:} Throughout this paper, we define $|g|_{\infty}:=\sup_x |g(x)|$ for a function $g$. For two sequences of positive numbers, $(a_n)$ and $(b_n)$, we write $a_n\lesssim b_n$ or $a_n=O(b_n)$ if there exists a positive constant $C$ such that $a_n/b_n\leq C$ for all large $n$. Similarly, we write $a_n=o(b_n)$ if $a_n/b_n\to0$ as $n$ goes to infinity. For two sequences of random variables, $(X_n)$ and $(Y_{n})$, we write $X_n=O_{\PP}(Y_n)$ if for all $\epsilon>0$, there exists $C>0$ such that $\PP(|X_n/Y_n|\le C)\geq1-\epsilon$ for all large $n$. Finally, we write $X_n=o_{\mathbb{P}}(Y_{n})$ if $X_n/Y_{n}\to0$ in probability as $n$ goes to infinity.

\section{A Smoothed SGD Recursion for Quantiles}\label{sec:smoothed}

Let $(X_{1,k}, X_{2, k}, \ldots, X_{p, k})^\top, k = 1, 2, \ldots$, be i.i.d.\ random vectors which are distributed as $(X_{1}, X_{2}, \ldots, X_{p})^\top$. Let $F_i(x)$ be the distribution function of $X_{i, k}$. The $\tau$-quantile of $X_i$ is defined as the minimizer of the expected quantile loss,
\begin{align*}
    Q_i(\tau):=\argmin_{x\in\mathbb{R}}\EE\left\{\rho_\tau(X_i-x)\right\},
\end{align*}
where $\rho_\tau(u)=u(\tau-\1_{u<0})$ is the quantile loss function at quantile level $\tau$. If $F_i$ is smooth, then we have that $F_i(Q_i(\tau))=\tau$. A natural empirical estimator takes the form,
\begin{align*}
    \widehat{Q}_i(\tau):=\argmin_{x\in\mathbb{R}}\frac{1}{n}\sum_{k=1}^n\rho_\tau(X_{i,k}-x).
\end{align*}
The statistical properties of $\widehat{Q}_i(\tau)$ are well studied. However, the estimator is neither computationally nor memory-efficient and might be infeasible for large-scale streaming data. An alternative estimator which avoids these issues can be obtained via the stochastic gradient descent (SGD) algorithm. Starting from a constant initial value $Y_{i,1}(\tau)=y_{i}$,
with $\max_{1\leq i\leq p}|y_i|:=c_y<\infty$, estimates are updated recursively by the following generating mechanism:
\begin{align*}
Y_{i,k+1}(\tau)=Y_{i,k}(\tau)+\gamma_k\Big(\tau-\1_{\{Y_{i,k}(\tau)-X_{i,k+1}\geq 0\}}\Big),    
\end{align*}
where $\gamma_k=c_\gamma k^{-\beta}$ is the learning rate and $c_\gamma>0$ is a constant. If the learning rate fulfills the condition $1/2<\beta<1$, the SGD solution converges almost surely to the true $\tau$-quantile. \citet{chen2023recursive} developed non-asymptotic theory on the sub-exponential convergence of $Y_{i,k}(\tau)$.
\begin{figure}[tbp]
        \centering
            \includegraphics[width=0.6\textwidth]{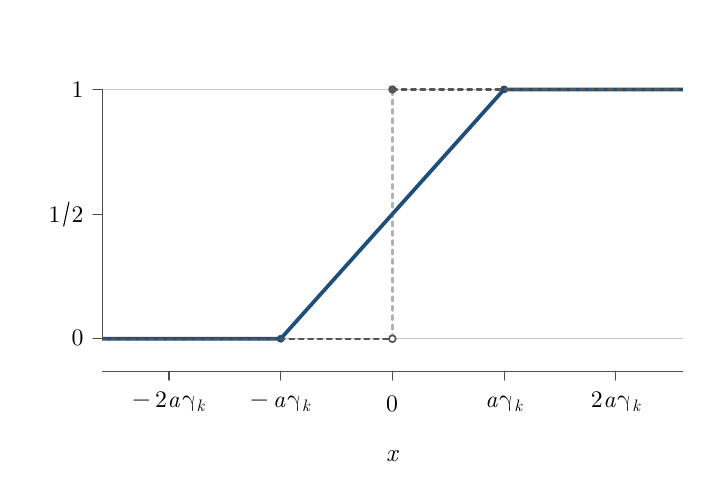}
        \caption{\footnotesize Visualization of $g_{a\gamma_k}(x)$ (solid line) vs. the nonsmooth indicator function (dashed line).}
        \label{figure:g}
\end{figure}

\begin{figure}[tbp]
        \centering
        \begin{subfigure}
            \centering
            \includegraphics[width=0.49\textwidth]{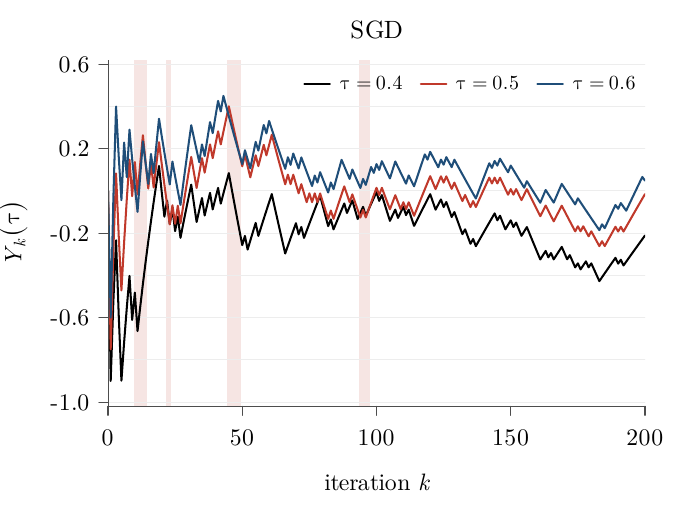}
        \end{subfigure}
        \hfill
        \begin{subfigure}
            \centering
            \includegraphics[width=0.49\textwidth]{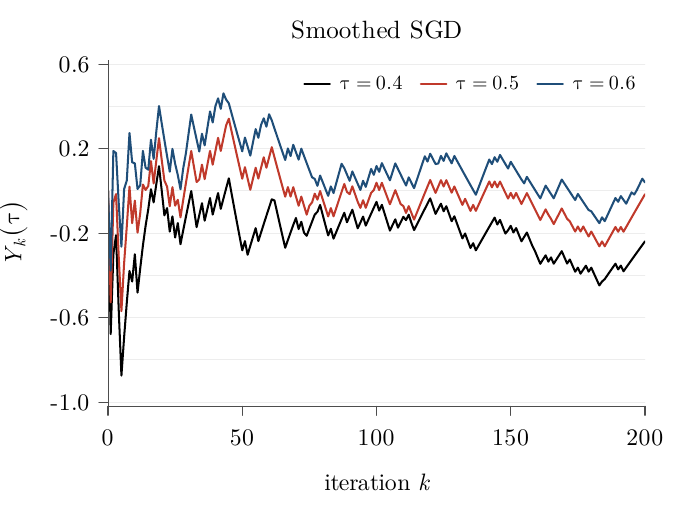}
        \end{subfigure}
        \caption{\footnotesize Comparison of the SGD algorithm (left panel) and the smoothed SGD algorithm (right panel) for $\tau=0.4$ (black line), $\tau=0.5$ (red line) and $\tau=0.6$ (blue line). Shaded areas indicate quantile crossings.}
        \label{figure:monotone}
\end{figure}

A crucial problem of the standard SGD quantile estimation algorithm is that the generated $Y_{i,k}(\tau)$ are not monotonically increasing with respect to $\tau$ due to the non-smoothness of the indicator function. Figure~\ref{figure:monotone} shows that quantile curves can cross, a highly undesirable shortcoming. Consider the following modification, which replaces the indicator function with a piecewise linear function, 
\begin{align*}
g(x)=\begin{cases}
1 \quad &\textrm{if  } x\geq 1,\\
(x+1)/2&\textrm{if  } -1\leq x<1,\\
0       &\textrm{if  } x<-1.
\end{cases}
\end{align*}
See Figure~\ref{figure:g} for a visualization. Further, we write $g_k(x)=g(x/k)$. The adjusted SGD scheme for any $1\leq i\leq p$ is $Y_{i,1}(\tau)=y_i$ and 
\begin{align}
\label{eq:sgdcont}
Y_{i,k+1}(\tau)=Y_{i,k}(\tau)+\gamma_k\big(\tau-g_{a\gamma_k}(Y_{i,k}(\tau)-X_{i,k+1})\big),
\end{align}
where the constant $a\geq 1.$
Throughout the paper, we take
\begin{align*}
\gamma_k=c_\gamma k^{-\beta},
\quad c_\gamma>0,\quad \textrm{and}\quad\frac12<\beta<1.    
\end{align*}

In contrast to the existing SGD algorithm, the unpleasant quantile crossing issue can be avoided by the new smoothed gradient algorithm. This issue is illustrated in Figure~\ref{figure:monotone}. Lemma~\ref{lem:monotone} establishes its monotonicity. The proof is given in
Appendix~\ref{sec:proof_sec3}.

\begin{lemma}[Monotonicity]
\label{lem:monotone}
For any \(1\le i\le p\), \(\tau\ge \tau'\), and \(k\in\mathbb N\), we have
\[
Y_{i,k}(\tau)\ge Y_{i,k}(\tau').
\]
\end{lemma}
We want to point out the deep connection between our smoothed SGD algorithm and the convolution-type smoothed quantile regression (\emph{conquer}) method proposed by \citet{fernandes2021smoothing} and \citet{he2023smoothed}. Instead of smoothing the indicator function in the quantile loss function, as proposed by \citet{horowitz1998bootstrap}, the method is based on the following modified version of the quantile loss,
\begin{align*}
    \ell_{\tau,h}(u)=\left(\rho_\tau*K_h\right)(u)=\int_{-\infty}^{\infty}\rho_{\tau}(v)K_h(v-u)dv,
\end{align*}
where $*$ is the convolution operator, $K_h(u)=h^{-1}K(u/h)$ and $K(u)$ is a kernel function which integrates to one with bandwidth parameter $h$. Our SGD generating scheme in (\ref{eq:sgdcont}) is based on this modified loss function with a uniform kernel and an adaptively chosen bandwidth. To be more precise, $K(u)=\frac{1}{2}\1_{|u|\leq1}$ and the bandwidth parameter decreases with each iteration exactly corresponding to the learning rate, $h_k=a\gamma_k$. Consequently, $h_k$ approaches zero when the number of steps goes to infinity, allowing fast online updating and computation. This is one of the fundamental differences of our approach compared to the \emph{conquer} method, which chooses only one bandwidth parameter for the entire sample. An iteratively chosen bandwidth is attractive in online settings, especially in the presence of an unknown time horizon $n$.

Due to favorable asymptotic properties, we consider the averaged version of the algorithm, commonly known as Polyak-Ruppert averaging \citep{ruppert1988efficient, polyak1992acceleration}. The final estimate takes the form,
\begin{align*}
    \bar{Y}_{i,n}(\tau)=\frac{1}{n}\sum_{k=1}^{n}Y_{i,k}(\tau),
\end{align*}
which allows recursive updating via $\bar Y_{i,k+1}(\tau)=k\bar Y_{i,k}(\tau)/(k+1)+Y_{i,k+1}(\tau)/(k+1)$. The monotonicity property remains unaffected by applying this additional averaging step.

\section{Non-asymptotic Tail Probability Bounds for Smoothed SGD}\label{sec:tailprobability}

In this section, we derive non-asymptotic tail probability bounds for the smoothed SGD algorithm with Polyak-Ruppert averaging. We are interested in the tail probability
\begin{align*}
    \PP\left(\max_{1\leq i\leq p}\sup_{\tau_0\leq\tau\leq\tau_1}\left|\bar Y_{i,n}(\tau)-Q_i(\tau)\right|>x\right),\text{ for }x>0.
\end{align*}
For this purpose, let $\tau_0<\tau_1$ be some constants in $(0,1)$, and we are interested in examining the performance of $Y_{i,k}(\tau)-Q_i(\tau)$ for $\tau\in [\tau_0,\tau_1]$. We impose the following smoothness condition on the density of $X_{i,k}$, which ensures the uniqueness of the theoretical quantile.

\begin{assumption}
\label{asmp:condf}
Assume that \(X_{i,k}\) has distribution function \(F_i(\cdot)\) and density \(f_i(\cdot)\).
For each \(1\le i\le p\), assume that \(f_i\) is differentiable and that
\[
\max_{1\le i\le p}
\left(
|f_i|_\infty+|f_i'|_\infty
\right)
\le c_f .
\]
Moreover, assume
\[
c_L:=\inf_{1\le i\le p}\inf_{\tau\in[\tau_0,\tau_1]}
f_i(Q_i(\tau))>0
\qquad\text{and}\qquad
\max_{1\le i\le p}
\sup_{\tau_0\le \tau\le \tau_1}|Q_i(\tau)|\le M .
\]    
\end{assumption}

\begin{theorem}
\label{thm:singletail}
Under Assumption~\ref{asmp:condf}, for any $0\leq t\leq cn^{\beta(1-\beta)}$ and
$\tau_0\leq \tau\leq \tau_1,$ we have
\begin{align}
\label{eq:mgfbdd}
\EE(e^{t|Y_{i,n}(\tau)-Q_i(\tau)|})\leq c'n^\beta,    
\end{align}
where $c,c'>0$ are some positive constants independent of $n,p.$ 
\end{theorem}

Theorem~\ref{thm:singletail} provides an upper bound on the moment generating function of $|Y_{i,n}(\tau)-Q_i(\tau)|$. It implies that, for any $x>0,$
\begin{align}
\label{eq:singletailbdd}
\PP\big(|Y_{i,n}(\tau)-Q_i(\tau)|> x\big)\leq c'n^\beta \exp\big\{-cn^{\beta(1-\beta) }x\big\}.    
\end{align}
The above bound generalizes and adapts Theorem 3.1 of \citet{chen2023recursive} to the smoothed recursion, while the previous result was based on the SGD algorithm without smoothing the score function. It turns out that a sharper upper bound exists by using a more refined argument. Theorem~\ref{thm:tailsingleonerefine} below provides a sharper sub-exponential tail probability bound for the smoothed SGD solution for quantile estimation. The proof of Theorem~\ref{thm:tailsingleonerefine} is highly nontrivial as it involves martingale approximation, martingale concentration inequalities, a careful analysis of the recursive algorithm and an asymptotic linear approximation; see Appendix~\ref{sec:proof_sec3} of the Supplementary Material. 

\begin{theorem}
\label{thm:tailsingleonerefine}
Under Assumption~\ref{asmp:condf}, we have  for any $x > 0$ that
\begin{align}\label{3terms}
\PP\Big(|Y_{i,n}(\tau)-Q_i(\tau)|>x\Big)
\lesssim \min\Big\{e^{-c n^\beta x^2}+e^{-c'n^{\min\{2(1-\beta),\beta\}}}, e^{-c''n^{\beta(1-\beta)}x}\Big\},
\end{align}
where $c,c',c''$ are positive constants independent of $p,n,i,\tau.$ In particular when $\beta\leq 2/3$, we have
\begin{align}
\label{eq:simplifiedbdd}
\PP\Big(|Y_{i,n}(\tau)-Q_i(\tau)|>x\Big)
\lesssim \min\Big\{e^{-c n^\beta x^2}+e^{-c'n^\beta}, e^{-c''n^{\beta(1-\beta)}x}\Big\}.    
\end{align}
\end{theorem}

\begin{remark}
Theorem~\ref{thm:tailsingleonerefine} reveals an interesting multi-regime
phenomenon in the upper bound. Considering the exponential terms in
\eqref{eq:simplifiedbdd} up to constants in the exponents, the dominating exponents are
\[
\begin{cases}
- n^{\beta(1-\beta)}x,
& 0<x\lesssim n^{-\beta^2},\\[3pt]
- n^\beta x^2,
& n^{-\beta^2}\lesssim x<1,\\[3pt]
- n^\beta,
& 1\le x\lesssim n^{\beta^2},\\[3pt]
- n^{\beta(1-\beta)}x,
& x\gtrsim n^{\beta^2}.
\end{cases}
\]
Compared with \eqref{eq:singletailbdd} in Theorem~\ref{thm:singletail},
Theorem~\ref{thm:tailsingleonerefine} gives a sharper bound in the range
\(n^{-\beta^2}\lesssim x\lesssim n^{\beta^2}\). In particular, it gives a
Gaussian-type improvement \(e^{-c n^\beta x^2}\) for
\(n^{-\beta^2}\lesssim x<1\), and a fixed-threshold improvement
\(e^{-c n^\beta}\) for \(1\le x\lesssim n^{\beta^2}\).
\end{remark}

We now derive in the following theorem non-asymptotic probability bounds for the smoothed SGD algorithm with Polyak-Ruppert averaging. In addition, the bounds in Theorems~\ref{thm:singletail} and \ref{thm:tailsingleonerefine} hold for a fixed quantile level $\tau$. In Theorem~\ref{thm:tail} we therefore extend this result to hold uniformly over $\tau_0\leq\tau\leq\tau_1$.

\begin{theorem}
\label{thm:tail}
Under Assumption~\ref{asmp:condf}, for $\tau_0\leq \tau\leq \tau_1$ we have
\begin{align}
\label{eq:singletailbaryintau}
\PP\big(|\bar Y_{i,n}(\tau)-Q_i(\tau)|>x\big)\lesssim &\exp(-c_1nx^2/2)+ 
n\exp(-c_2E(x))+n\exp(-c_3E'(x)),
\end{align}
where $\alpha=\min\{2(1-\beta), \beta\}$ with
\begin{align*}
E(x)=\max\Big\{\min\{(nx)^{\alpha/(\alpha+\beta^2)}, n^{2-\beta}x^2, n^\alpha\}, n^{1-\beta^2}x \Big\}
\end{align*}
and
\begin{align*}
E'(x)=\max\Big\{\min\{ (nx)^{\alpha/(2\alpha+1-2\beta(1-\beta))}, n^\beta x\}, n^{\beta(1-\beta)}x^{1/2} \Big\}.    
\end{align*}
Consequently, we have the uniform upper bound
\begin{align}
\label{eq:maxsupitaubardiff}
&\PP\Big(\max_{1\leq i\leq p}\sup_{\tau_0\leq \tau\leq \tau_1}|\bar Y_{i,n}(\tau)-Q_i(\tau)|> x\Big)\nonumber\\
\lesssim &p(1+x^{-1})\Big(
\exp(-c_1nx^2/2)+ 
n\exp(-c_2E(x))+n\exp(-c_3E'(x))\Big).
\end{align}
Here the constants in $\lesssim,$ $c_1,c_2$ and $c_3$ are independent of $i,\tau, n, p.$ 
\end{theorem}

In the special case $\beta\le 2/3$, we have $\alpha=\beta$. Moreover, the
$E(x)$-term can be absorbed into the $E'(x)$-term. Hence, \eqref{eq:singletailbaryintau} becomes
\begin{align}
&\PP\Big(|\bar Y_{i,n}(\tau)-Q_i(\tau)|> x\Big)\nonumber\\
\lesssim & \exp(-c_1 nx^2/2)+n\min\left\{
\exp\left[-c_2\min\{(nx)^{\beta/(1+2\beta^2)},\, n^\beta x\}\right], \exp\left(-c_3n^{\beta(1-\beta)}x^{1/2}\right)
\right\}.
\end{align}

A key step for deriving this uniform bound is the monotonicity of both the theoretical quantile, $Q_i(\tau)$, and the smoothed SGD solution, $\bar Y_{i,n}(\tau)$, which we showed in Lemma~\ref{lem:monotone}. 


We now compare our bound to previous results in the literature. A closely related result is the exponential bound for the averaged SGD algorithm without smoothing in Theorem 3.2 of \citet{chen2023recursive}. An important difference is that our bound is valid uniformly over the quantile level whereas the above result is limited to the fixed $\tau$ case. Another related paper is \citet{cardot2017online} who provide a tail probability bound for the SGD estimate for the geometric median which is a generalization of the classical median to the multivariate case. However, their bound is only valid for dimensions larger than 2. Theorem 3.4 of \citet{chen2023recursive} extends the result to the univariate case. This bound is still only decreasing algebraically in $x$ and it is only valid for a fixed quantile level $\tau$.

\section{Bahadur Representation and Gaussian Approximation}\label{sec:bahadur}

In this section, we introduce a uniform Bahadur representation which shows the asymptotic linearity of the smoothed SGD procedure for quantile estimation. Based on this Bahadur representation, we further present a Gaussian approximation result which allows us to conduct simultaneous inference on the estimated quantiles. To provide some historical background, for the (non-recursive) sample quantile estimate, \citet{bahadur1966note} showed the asymptotic linearization result
\begin{align*}
    \widehat{Q}_i(\tau)-Q_i(\tau)=\frac{\tau - F_{i,n}(Q_i(\tau))}{f_i(Q_i(\tau))}+O_{\mathrm{a.s.}}(n^{-3/4}(\log n)^{1/2}(\log\log n)^{1/4}),
\end{align*}
where $F_{i,n}(u) = n^{-1} \sum_{j=1}^n {\1}_{\{X_{i, j} \le u\}}$ denotes the empirical cdf of $X_i$. We refer to \citet{kiefer1967bahadur} and \citet{einmahl1996short} for further details. Such Bahadur-type representations are useful in a variety of statistical applications, including inference on quantile regression coefficients \citep{koenker1978regression}, sample quantiles under dependent data \citep{sen1968asymptotic, wu2005bahadur}, and general M-estimation problems \citep{he1996general}. Recently, \citet{he2023smoothed} derived a Bahadur-type representation to analyze the asymptotic properties of their \emph{conquer} method for linear quantile regression.

We first need to introduce some notation. We can rewrite the SGD iteration scheme \eqref{eq:sgdcont} into $Y_{i,k+1}(\tau)=Y_{i,k}(\tau)+\gamma_k Z_{i,k+1}(\tau)$, where
\begin{align*}
Z_{i,k+1}(\tau)=\tau-g_{a\gamma_k}(Y_{i,k}(\tau)-X_{i,k+1}).
\end{align*}
Further, for $\F_{i,k}=\sigma(X_{i,k}, X_{i,k-1},\ldots)$ let
\begin{align}
\label{eq:defxiiktau}
\bar\xi_{i,n}(\tau)=n^{-1} \sum_{k=1}^n \xi_{i,k+1}(\tau), \mbox{ where }
\xi_{i,k}(\tau)=Z_{i,k}(\tau)-\EE(Z_{i,k}(\tau)|\F_{i,k-1}).    
\end{align} 
Then $\xi_{i,k+1}(\tau)$, $k\geq 1,$ are martingale differences with respect to $\F_{i,k}.$


The following theorem establishes the Bahadur representation for the averaged smoothed SGD algorithm which holds uniformly over the quantile level and over $i$.
\begin{theorem}[Uniform Bahadur representation]
\label{thm:unifbahadur}
Under Assumption~\ref{asmp:condf}, we have for any $x > 0$ that
\begin{align}
\label{eq:tailforunifbahadur}
&\PP\Big(\max_{1\leq i\leq p}\sup_{\tau_0\leq \tau\leq \tau_1} \big|\bar Y_{i,n}(\tau)-Q_i(\tau)-\bar\xi_{i,n}(\tau)/f_i(Q_i(\tau))\big| >x\Big)\nonumber\\
\lesssim& np(1+n/x)\Big(
\exp(-c_1E(x))+\exp(-c_2E'(x))\Big),
\end{align}
where $E(x)$ and $E'(x)$ are the same as in \eqref{eq:singletailbaryintau}.
\end{theorem}

\begin{remark}
As a direct consequence of Theorem~\ref{thm:unifbahadur}, letting \(L_n=\log(np)\), we have
\begin{align}
&\max_{1\le i\le p}\sup_{\tau_0\le \tau\le \tau_1}
\left|
\bar Y_{i,n}(\tau)-Q_i(\tau)
-\bar \xi_{i,n}(\tau)/f_i(Q_i(\tau))
\right|\nonumber\\
=&O_{\mathbb P}\left(
n^{-\beta}L_n
+n^{-(2-\beta)/2}L_n^{1/2}
+n^{-1}L_n^{s}
\right),
\label{J27706p}
\end{align}
where $\alpha=\min\{2(1-\beta),\beta\}$ and $s=(2\alpha+1-2\beta(1-\beta))/\alpha$.
\end{remark}

In the low-dimensional case with \(p\asymp n^v\), \(v\ge 0\), up to a
multiplicative logarithmic factor, the dominating term in \eqref{J27706p} is
\(n^{-\beta}\) when \(\beta<2/3\), and \(n^{-(2-\beta)/2}\) when
\(\beta\ge 2/3\). 
In the high-dimensional case with \(\log p\asymp n^v\), \(v>0\), the three
terms in \eqref{J27706p} are respectively of polynomial orders
\begin{align*}
n^{-\beta+v},\qquad
n^{-(2-\beta)/2+v/2},\qquad
n^{-1+vs}.    
\end{align*}
Hence the dominating term depends jointly on \(\beta\) and \(v\), and is
given by the largest of these three polynomial orders.

Based on the above Bahadur representation for the smoothed SGD solution with Polyak-Ruppert averaging, we obtain the following Gaussian approximation result, which is useful for performing simultaneous inference of quantiles across coordinates and quantile levels.

\begin{theorem}
\label{thm:main}
Under Assumption~\ref{asmp:condf}, we have
\begin{align}\label{J27}
&\sup_{u\in\RR}\Big|\PP\Big(\max_{1\leq i\leq p}\sup_{\tau_0 \leq \tau\leq \tau_1}n^{1/2}|f_i(Q_i(\tau))(\bar Y_{i,n}(\tau)-Q_i(\tau))|\leq u \Big)\cr
&\qquad -\PP\Big(\max_{1\leq i\leq p}\sup_{\tau_0\leq \tau\leq \tau_1}|\B_i(\tau)| \leq u\Big)\Big|\nonumber\\
&\qquad\lesssim L_n^{7/6}n^{-1/6}+L_n^{3/2}n^{-r}+L_n^{s+1/2}n^{-1/2},
\end{align}
where $L_n=\log(np)$, $r=\min\{(1-\beta)/2, \beta-1/2\}$,
$s=(2\alpha+1-2\beta(1-\beta))/\alpha$, $\alpha=\min\{2(1-\beta),\beta\}$ and
$(\B_i(t))_{i, t}$ is a centered Gaussian process with covariance function
\begin{equation*}
    \cov(\B_i(t), \B_j(s)) =\PP\big(X_{i,k+1}\leq Q_i(t), X_{j,k+1}\leq Q_j(s)\big)-t s.
\end{equation*}
In particular, for each $i$, $\B_i(t)$ is a Brownian bridge on $[0,1]$.
\end{theorem}

Theorem~\ref{thm:main} is the key tool for simultaneous inference on the quantiles $Q_i(\tau)$, jointly over the coordinates $1\le i\le p$ and the quantile levels $\tau\in[\tau_0,\tau_1]$. As a consequence, critical values can be obtained by means of simulation. We study the finite-sample properties of our Gaussian approximation result in Section~\ref{sec:numerical}.

\begin{remark}[Allowed dimension of $p$ relative to $n$]\label{remark:dimension}
We want to highlight that the dimension $p$ is allowed to be of exponential
order in $n$, and the admissible order depends on the learning rate parameter $\beta$.
Assume $\log(p)\asymp n^v$ with \(v>0\). Then the three terms in the upper
bound in Theorem~\ref{thm:main} are of polynomial orders
\[
n^{-1/6+7v/6},
\qquad
n^{-r+3v/2}
\qquad\text{and}\qquad
n^{-1/2+v(s+1/2)},
\]
so that the Gaussian approximation error converges to zero provided
\begin{align*}
v<\min\Big\{\frac17,\ \frac{2r}{3},\ \frac{1}{2s+1}\Big\}.
\end{align*}
The second quantity is always the smallest. Indeed, $r\leq 1/6$ gives
$2r/3<1/7$. Moreover, for $1/2<\beta\leq 2/3$ we have $r=\beta-1/2$,
$\alpha=\beta$ and $s=1/\beta+2\beta$, while for $2/3\leq\beta<1$ we have
$r=(1-\beta)/2$, $\alpha=2(1-\beta)$ and $s=1+(1-\beta)+\{2(1-\beta)\}^{-1}$.
In both cases $2r(2s+1)<3$, that is, $2r/3<1/(2s+1)$. Therefore the
Gaussian approximation error converges to zero provided $v<2r/3$,
and the right-hand side is maximized at $\beta=2/3$. At this value $r=1/6$, and
hence the admissible range is $v<1/9$.
\end{remark}

\section{Online Multiplier Bootstrap Procedure}\label{sec:bootstrap}
In this section, we propose an online bootstrap procedure for uniform inference on the estimated quantiles as an alternative to explicitly relying on the Gaussian approximation result in Theorem~\ref{thm:main}. For this purpose, we consider a randomly perturbed version of the smoothed SGD procedure in \eqref{eq:sgdcont}.
Recall $a\geq1$, and let $W_1,W_2,\ldots,W_n$ be i.i.d.\ random
variables with mean one and variance one, satisfying
\[
0\leq\omega_1\leq W_k\leq\omega_2\leq2a.
\]
The constant $a$ thus plays a dual role: while $a\geq1/2$ is sufficient
for monotonicity of the original recursion in Lemma~\ref{lem:monotone},
the stronger condition $a\geq1$ ensures that the admissible multiplier
class is nonempty and that the bound $W_k\leq2a$ preserves monotonicity
of the perturbed recursion. In particular, the two-point distribution
$\PP(W=0)=\PP(W=2)=1/2$ is admissible for every $a\geq1$ and attains
the upper bound when $a=1$. We further assume that $(W_k)_k$ is
independent of $(X_k)_k$. Consider new sequences $Y_{i,k}^*(\tau)$ with $Y_{i,1}^*(\tau)=0$ and 
\begin{align}
\label{eq:bootstrap}
Y_{i,k+1}^*(\tau)=Y_{i,k}^*(\tau)+\gamma_kW_{k+1}\Big(\tau-g_{a\gamma_k}(Y_{i,k}^*(\tau)-X_{i,k+1})\Big).    
\end{align}
A similar online bootstrap procedure based on perturbing the SGD updating step was proposed in \citet{fang2018online}. However, that work requires convex, twice continuously differentiable losses, which excludes the quantile loss. Additionally, our setting is more demanding since we target inference that is uniform in $i$ and in the quantile level $\tau$. The following lemma shows that the perturbed version of the smoothed SGD algorithm retains the monotonicity property.

\begin{lemma}[Monotonicity of $Y_{i,k}^*(\tau)$]
\label{lem:monotone_bootstrap}
For any $\tau\geq \tau'$ and $k\in\NN,$ we have $Y_{i,k}^*(\tau)\geq Y_{i,k}^*(\tau').$
\end{lemma}

The scheme outlined in \eqref{eq:bootstrap} is an online version of the multiplier bootstrap. The goal of reweighting the update by an independent multiplier $W_{k+1}$ is to generate a perturbed averaged trajectory $\bar Y_{i,n}^*(\tau)$ whose fluctuation around $\bar Y_{i,n}(\tau)$ mimics that of $\bar Y_{i,n}(\tau)$ around $Q_i(\tau)$. The conditions $\EE(W_k)=1$ and $\var(W_k)=1$ ensure that the bootstrap process inherits the same covariance structure as in Theorem~\ref{thm:main}. The boundedness condition $W_k\leq 2a$ is needed to preserve the monotonicity in the quantile level and to keep the martingale differences bounded.
The perturbed recursion satisfies bootstrap analogues of the tail bounds
and the uniform Bahadur representation established for the original
recursion. The required decomposition and technical details are given in
Appendix~\ref{sec:proof_sec5}.
\begin{proposition}
\label{prop:tailsingleonerefine_bootstrap}
Under Assumption~\ref{asmp:condf}, for the perturbed recursion \eqref{eq:bootstrap} we have for any $x>0$ and $\tau_0\leq\tau\leq\tau_1$ that
\begin{align}
\label{eq:3terms_bootstrap}
\PP\Big(|Y_{i,n}^*(\tau)-Q_i(\tau)|>x\Big)
\lesssim \min\Big\{e^{-c n^\beta x^2}+e^{-c'n^{\min\{2(1-\beta),\beta\}}}, e^{-c''n^{\beta(1-\beta)}x}\Big\},
\end{align}
where $c,c',c''$ and the constant in $\lesssim$ are independent of $p,n,i,\tau$. In particular, when $\beta\leq 2/3$, the bound \eqref{eq:simplifiedbdd} holds with $Y_{i,n}(\tau)$ replaced by $Y_{i,n}^*(\tau)$.
\end{proposition}

\begin{corollary}
\label{cor:tail_bootstrap}
Under Assumption~\ref{asmp:condf}, Theorem~\ref{thm:tail} holds with $\bar Y_{i,n}(\tau)$ replaced by $\bar Y_{i,n}^*(\tau)$. Specifically, the pointwise bound \eqref{eq:singletailbaryintau} and the uniform bound \eqref{eq:maxsupitaubardiff} hold with the same $E(x)$ and $E'(x)$.
\end{corollary}

Define the unstudentized sampling and bootstrap statistics
\begin{align*}
T
&=
\max_{1\le i\le p}
\sup_{\tau_0\le\tau\le\tau_1}
n^{1/2}
\left|\bar Y_{i,n}(\tau)-Q_i(\tau)\right|,\\
T^*
&=
\max_{1\le i\le p}
\sup_{\tau_0\le\tau\le\tau_1}
n^{1/2}
\left|\bar Y^*_{i,n}(\tau)-\bar Y_{i,n}(\tau)\right|.
\end{align*}
The following theorem establishes unconditional validity of the
unstudentized bootstrap. 

\begin{theorem}\label{thm:bootstrap}
Assume Assumption~\ref{asmp:condf}. Recall the definitions of
$L_n$, $r$, and $s$ from Theorem~\ref{thm:main}, and let
\begin{align*}
R_n
=
L_n^{7/6}n^{-1/6}
+
L_n^{3/2}n^{-r}
+
L_n^{s+1/2}n^{-1/2}.
\end{align*}
Then
\begin{align*}
\sup_{u\in\RR}
\left|
\PP(T^*\le u)-\PP(T\le u)
\right|
\lesssim R_n.
\end{align*}
\end{theorem}
For the conditional result, let
$\mathcal D_n=\sigma\{X_k:1\le k\le n+1\}$
denote the $\sigma$-field generated by the data. 
\begin{theorem}
\label{thm:bootstrap_cond}
Assume Assumption~\ref{asmp:condf}. Then there exist constants
$C,C'>0$, independent of $n$ and $p$, such that
\begin{align*}
\PP\left\{
\sup_{u\in\RR}
\left|
\PP\left(T^*\le u\mid\mathcal D_n\right)
-
\PP(T\le u)
\right|
\le CR_n
\right\}
\ge 1-C'(np)^{-1/2}.
\end{align*}
Consequently we have $
\sup_{u\in\RR}
\left|
\PP\left(T^*\le u\mid\mathcal D_n\right)
-
\PP(T\le u)
\right|
=O_{\PP}(R_n).$
\end{theorem}

The conditional validity result provides a density-estimation-free
simultaneous confidence band. For a significance level $\eta\in(0,1)$, let
$
c^*_{1-\eta}
=
\inf\left\{
u\in\RR:
\PP\left(T^*\le u\mid\mathcal D_n\right)\ge 1-\eta
\right\}.$
By Theorem~\ref{thm:bootstrap_cond} and Gaussian
anti-concentration, the resulting critical value satisfies
\[
\left|
\PP\left(T\le c^*_{1-\eta}\right)
-
(1-\eta)
\right|
\lesssim R_n+(np)^{-1/2}.
\]
Consequently, if $R_n\to0$, the simultaneous confidence band is
\begin{align*}
\PP\Bigg(
Q_i(\tau)\in
\left[
\bar Y_{i,n}(\tau)-\frac{c^*_{1-\eta}}{\sqrt n},
\,
\bar Y_{i,n}(\tau)+\frac{c^*_{1-\eta}}{\sqrt n}
\right]
\ \text{for all }\
1\le i\le p,\ 
\tau_0\le\tau\le\tau_1
\Bigg)
=
1-\eta+o(1).
\end{align*}
Therefore, the simultaneous confidence band is
\begin{align}
\label{eq:density_free_bootstrap_band}
\left[
\bar Y_{i,n}(\tau)-\frac{c^*_{1-\eta}}{\sqrt n},
\,
\bar Y_{i,n}(\tau)+\frac{c^*_{1-\eta}}{\sqrt n}
\right],
\qquad
1\le i\le p,\quad
\tau_0\le\tau\le\tau_1.
\end{align}
Neither the critical value nor the confidence band requires estimation of
$f_i(Q_i(\tau))$.
For practical implementation, one can run $B$ perturbed versions of the
smoothed SGD algorithm together with the original recursion in a single online
pass. Let $\mathcal G\subset[\tau_0,\tau_1]$ be a grid of quantile levels. For
each bootstrap replication $b=1,\ldots,B$, compute
\begin{align*}
T_{\mathcal G}^{*(b)}
=
\max_{1\le i\le p}
\max_{\tau\in\mathcal G}
n^{1/2}
\left|
\bar Y_{i,n}^{*(b)}(\tau)-\bar Y_{i,n}(\tau)
\right|.
\end{align*}
Let $\widehat c_{1-\eta}^*$ be the empirical $(1-\eta)$-quantile of
$\{T_{\mathcal G}^{*(b)}:1\le b\le B\}$.
Conditional on $\mathcal D_n$, the bootstrap replications are i.i.d., and the
Dvoretzky--Kiefer--Wolfowitz inequality implies that their empirical
distribution approximates the conditional bootstrap distribution uniformly
with error $O_{\PP}(B^{-1/2})$. Hence the empirical bootstrap band has coverage
$1-\eta+o(1)$ whenever $R_n\to0$ and $B\to\infty$, retaining the approximation
rate $R_n$ requires $B^{-1/2}=O(R_n)$.
The resulting simultaneous
confidence band over the grid is
\begin{align*}
\left[
\bar Y_{i,n}(\tau)-\frac{\widehat c_{1-\eta}^*}{\sqrt n},
\,
\bar Y_{i,n}(\tau)+\frac{\widehat c_{1-\eta}^*}{\sqrt n}
\right],
\qquad
1\le i\le p,\quad \tau\in\mathcal G.
\end{align*}

\section{Nonparametric Online Quantile Regression}
\label{sec:conditional}

The theory developed above is useful not only for marginal quantiles but also
for conditional quantile curves. Linear quantile regression is a standard
tool in econometrics \citep{koenker1978regression}. Here we avoid imposing a
linear specification and instead consider a nonparametric conditional
quantile. Let $(Z_k,Y_k)$ be i.i.d.\ observations, where $Z_k$ is a scalar
covariate with density $f_Z$. Write $F(y\mid z)$ and $f(y\mid z)$ for the
conditional cdf and density of $Y$ given $Z=z$, and let
$q(z,\tau)=F^{-1}(\tau\mid z)$.

The classical nonparametric literature develops kernel, nearest-neighbor, and
local-polynomial conditional-quantile estimators
\citep{BhattacharyaGangopadhyay1990,Chaudhuri1991,FanHuTruong1994,YuJones1998}.
Uniform Bahadur representations are available for local-polynomial methods
\citep{KongLintonXia2010,guerre2012uniform}. Uniform inference over the
covariate, the quantile index, or both has likewise been developed
\citep{hardle2010confidence,Sabbah2014,quyoon2015,belloni2019conditional}.
These procedures are designed for batch data. Related methods address
memory-constrained or streaming quantile regression
\citep{chen2019quantile,jiang2022renewable,wang2025real,xie2025statistical}. The closest recursive nonparametric predecessor is
\citet{LabopinRichardEtAl2019}, who combine nearest-neighbor localization with
the quantile recursion of Robbins and Monro and establish convergence and
non-asymptotic mean-square bounds.

Our objective is an online local estimator that preserves monotonicity in the
quantile level, as established in Section~\ref{sec:smoothed}, and supports
simultaneous inference over design points $z_1,\ldots,z_p$ and quantile levels
$\tau\in[\tau_0,\tau_1]$, using only constant memory for each combination of
design point and quantile level.

For a specified conditional quantile function $q_0(z,\tau)$, consider the null hypothesis
\[
H_0:q(z_i,\tau)=q_0(z_i,\tau)
\quad\text{for every }1\le i\le p
\text{ and }\tau_0\le\tau\le\tau_1.
\]

To motivate the online construction, let $K$ be a symmetric, nonnegative
kernel supported on $[-1,1]$, let $K_{h_n}(u)=h_n^{-1}K(u/h_n)$, and consider
the usual batch local constant estimator
\[
 \widehat q^{\mathrm{batch}}(z,\tau)
 =\argmin_{u\in\RR}\sum_{k=1}^n
 \rho_\tau(Y_k-u)K_{h_n}(z-Z_k).
\]
We focus on the uniform kernel. At a design point $z_i$, the batch estimator is
then simply the empirical $\tau$-quantile of those responses for which
$|z_i-Z_k|\le h_n$. This observation motivates the basic idea of our online procedure. I.e., we feed only
those active responses into a smoothed SGD recursion and index its learning
rate by the number of active observations.
Accordingly, put
\[
 N_{i,k}=\sum_{j=1}^k\1_{\{|z_i-Z_j|\le h_n\}}.
\]
Following the unconditional algorithm in Section~\ref{sec:smoothed}, we start from
$\widehat q_{i,1}(\tau)=y_i$, where
$\max_{1\le i\le p}|y_i|\le c_y<\infty$.
Observation $(Z_k,Y_k)$ produces the next
iterate according to
\begin{equation}
\widehat q_{i,k+1}(\tau)=
\begin{cases}
\widehat q_{i,k}(\tau)+\gamma_{N_{i,k}}
 \left[\tau-g_{a\gamma_{N_{i,k}}}
 \{\widehat q_{i,k}(\tau)-Y_k\}\right],
 & |z_i-Z_k|\le h_n,\\
\widehat q_{i,k}(\tau), & |z_i-Z_k|>h_n,
\end{cases}
\label{eq:condsgdN}
\end{equation}
for $k=1,\ldots,n$, where $a\ge1$ and
$\gamma_r=c_\gamma r^{-\beta}$ as before. The argument for
Lemma~\ref{lem:monotone}, with $\gamma_k$ replaced by
$\gamma_{N_{i,k}}$, also shows that
$\tau\mapsto\widehat q_{i,k}(\tau)$ is nondecreasing.  See
Subsection~\ref{sec:mono_lnqr} in Appendix~\ref{app:local-sgd-details} for the proof. For $N_{i,n}\ge1$,
define the averaged estimator
\begin{equation}
 \bar q_{i,n}(\tau)=\frac{1}{N_{i,n}}
 \sum_{k=1}^n\1_{\{|z_i-Z_k|\le h_n\}}
 \widehat q_{i,k}(\tau).
\label{eq:local-average}
\end{equation}
When $N_{i,n}=0$, set $\bar q_{i,n}(\tau)=y_i$.

Conditional on the assignment to a certain window, the trajectory at
$z_i$ is the ordinary smoothed SGD recursion applied to an i.i.d.\ sample of
random size $N_{i,n}$, consisting of responses whose covariates satisfy
$|Z_k-z_i|\le h_n$. Disjoint windows make these samples conditionally
independent across design points. The two new ingredients are therefore the
random effective sample sizes and the local smoothing bias. 

The population quantile targeted by the local recursion
\eqref{eq:condsgdN} is the quantile of $Y$ conditional on $Z$ falling in a
neighborhood of $z_i$, rather than the pointwise quantile.
Let
\[
A_{i,h}=[z_i-h_n,z_i+h_n],
\qquad
\pi_{i,h}=\PP(Z\in A_{i,h}).
\]
Its cdf and density are
\begin{align}
F_{i,h}^{\mathrm{win}}(y)
&=\PP(Y\le y\mid Z\in A_{i,h})
=\pi_{i,h}^{-1}\int_{z_i-h_n}^{z_i+h_n}
F(y\mid z)f_Z(z)\,\d z,
\label{eq:window-cdf}\\
f_{i,h}^{\mathrm{win}}(y)
&=\pi_{i,h}^{-1}\int_{z_i-h_n}^{z_i+h_n}
f(y\mid z)f_Z(z)\,\d z.
\label{eq:window-density}
\end{align}
Define the window and pointwise quantiles and their corresponding density
values by
\[
\begin{aligned}
q_{i,h}(\tau)
&=\{F_{i,h}^{\mathrm{win}}\}^{-1}(\tau),
&
d_{i,h}(\tau)
&=f_{i,h}^{\mathrm{win}}\{q_{i,h}(\tau)\},\\
q_i(\tau)
&=q(z_i,\tau),
&
d_i(\tau)
&=f\{q_i(\tau)\mid z_i\}.
\end{aligned}
\]

\begin{assumption}[Local design and regularity of conditional distribution]
\label{asm:local-quantile}
Let $0<\tau_0<\tau_1<1$. The observations $(Z_k,Y_k)$ are i.i.d., and the
design points may depend on $n$. The constants $c_f$, $c_L$, $M$, and $c_y$
retain the same roles as in the preceding sections.
\begin{enumerate}
\item[(i)] The points $z_i$ lie in a fixed compact interval $\mathcal Z$ whose closed
$\eta_Z$-neighborhood is contained in the interior of the support of $Z$.
On that neighborhood, $f_Z$ is twice continuously differentiable and, for
some $0<c_Z<C_Z<\infty$,
\[
 c_Z\le f_Z(z)\le C_Z,
 \qquad |f_Z'(z)|+|f_Z''(z)|\le C_Z.
\]
Moreover, $h_n\downarrow0$, $h_n<\eta_Z$, and
$\min_{i\ne j}|z_i-z_j|>2h_n.$

\item[(ii)] For $z$ in the neighborhood from part (i) and all
$y\in\mathbb R$, $F(y\mid z)$ is continuously differentiable in $y$ with
$\partial_yF(y\mid z)=f(y\mid z)$, and $f(y\mid z)$ is continuously
differentiable in $y$ with
\[
 f(y\mid z)+|\partial_yf(y\mid z)|\le c_f.
\]
For $y$ in a fixed neighborhood of
$\{q(z,\tau):z\in\mathcal Z,\ \tau_0\le\tau\le\tau_1\}$, the functions
$F(y\mid z)$ and $f(y\mid z)$ are twice continuously differentiable in $z$,
with uniformly bounded first two derivatives and uniformly continuous second
derivatives. Finally,
\[
\inf_{z\in\mathcal Z}\inf_{\tau_0\le\tau\le\tau_1}
f\{q(z,\tau)\mid z\}\ge c_L>0,
\qquad
\sup_{z\in\mathcal Z}\sup_{\tau_0\le\tau\le\tau_1}
|q(z,\tau)|\le M.
\]
\end{enumerate}
\end{assumption}

This is an interior-point condition for a local constant estimator. Boundary
windows generally have first-order bias. Local linear quantile methods can
remove it \citep{YuJones1998,guerre2012uniform}.
Under Assumption~\ref{asm:local-quantile}, uniformly in $i$, we have
$\pi_{i,h}\asymp h_n$, the window distributions satisfy the regularity
conditions in Assumption~\ref{asmp:condf}, and
\[
\sup_{\tau_0\le\tau\le\tau_1}
\left\{
|q_{i,h}(\tau)-q_i(\tau)|
+|d_{i,h}(\tau)-d_i(\tau)|
\right\}
\lesssim h_n^2.
\]
These properties are stated and proved in
Lemma~\ref{lem:window-bias} in Appendix~\ref{app:local-sgd-details}.

Recall $r$ and $s$ from Theorem~\ref{thm:main} and define
\begin{equation}
 L_{n,h}=\log(e p n h_n),
 \qquad \lambda_p=1+\log p,
\label{eq:local-log-factors}
\end{equation}
and
\begin{equation}
 \mathcal R_{n,h,p}
 =L_{n,h}^{7/6}(nh_n)^{-1/6}
 +L_{n,h}^{3/2}(nh_n)^{-r}
 +L_{n,h}^{s+1/2}(nh_n)^{-1/2}.
\label{eq:local-GA-rate}
\end{equation}
Let $\B_1^{\mathrm{loc}},\ldots,\B_p^{\mathrm{loc}}$ be independent standard
Brownian bridges and put
\[
 G_p^{\mathrm{loc}}
 =\max_{1\le i\le p}\sup_{\tau_0\le\tau\le\tau_1}
 |\B_i^{\mathrm{loc}}(\tau)|.
\]
The superscript distinguishes these independent bridges from the generally
correlated process $\B_i$ in Theorem~\ref{thm:main}.

\begin{theorem}[Gaussian approximation at random local sample sizes]
\label{thm:local-window-gaussian}
Suppose Assumption~\ref{asm:local-quantile} holds and $nh_n\ge2$. Define
\begin{equation}
 T_n^{\mathrm{win}}
 =\max_{1\le i\le p}\sup_{\tau_0\le\tau\le\tau_1}
 N_{i,n}^{1/2}d_{i,h}(\tau)
 |\bar q_{i,n}(\tau)-q_{i,h}(\tau)|.
\label{eq:T-window}
\end{equation}
For constants $C,c>0$ independent of $n,p,h_n$,
\begin{equation}
 \sup_{u\in\RR}
 \left|\PP\{T_n^{\mathrm{win}}\le u\}
 -\PP(G_p^{\mathrm{loc}}\le u)\right|
 \le C\mathcal R_{n,h,p}+Cp\exp(-cnh_n).
\label{eq:window-GA}
\end{equation}
\end{theorem}

The disjointness condition in Assumption~\ref{asm:local-quantile}(i) yields
the independent bridges in the reference distribution. The covariance for
overlapping windows is given in Appendix~\ref{app:local-sgd-details}.

For inference on the pointwise conditional quantiles, let
\begin{equation}
 T_n^{\mathrm{loc}}
 =\max_{1\le i\le p}\sup_{\tau_0\le\tau\le\tau_1}
 N_{i,n}^{1/2}d_i(\tau)
 |\bar q_{i,n}(\tau)-q_i(\tau)|.
\label{eq:T-local}
\end{equation}

\begin{corollary}[Gaussian approximation for the conditional quantile curves]
\label{cor:local-point-gaussian}
Under the conditions of Theorem~\ref{thm:local-window-gaussian},
\begin{align}
 \sup_{u\in\RR}
 \left|\PP\{T_n^{\mathrm{loc}}\le u\}
 -\PP(G_p^{\mathrm{loc}}\le u)\right|
 \lesssim{}&\mathcal R_{n,h,p}+p\exp(-cnh_n)\notag\\
 &+\{nh_n^5\lambda_p\}^{1/2}
 +h_n^2(L_{n,h}\lambda_p)^{1/2}.
\label{eq:local-point-GA}
\end{align}
Consequently, the approximation error tends to zero if
\begin{equation}
 \mathcal R_{n,h,p}\to0,
 \qquad \frac{\lambda_p}{nh_n}\to0,
 \qquad nh_n^5\lambda_p\to0.
\label{eq:local-rate-conditions}
\end{equation}
The last condition is the simultaneous-inference undersmoothing condition.
\end{corollary}

For $\beta=2/3$ and fixed $p$, the choice
$h_n\asymp n^{-1/4}$ satisfies the sufficient rate conditions. In a fixed
compact design interval, disjointness also requires
$p\lesssim h_n^{-1}$. The detailed rate calculation is given in
Appendix~\ref{app:local-sgd-details}.

Let $c_{1-\alpha,p}^{\mathrm{loc}}$ be the $(1-\alpha)$ quantile of
$G_p^{\mathrm{loc}}$, which can be simulated from $p$ independent Brownian
bridges on the algorithm's quantile grid.

\begin{corollary}[Feasible simultaneous confidence bands]
\label{cor:local-feasible-bands}
Let $\widehat d_i(\tau)$ be positive estimators such that
\begin{equation}
 \Delta_{d,n}=\max_{1\le i\le p}
 \sup_{\tau_0\le\tau\le\tau_1}
 \left|\frac{\widehat d_i(\tau)}{d_i(\tau)}-1\right|,
 \qquad \lambda_p\Delta_{d,n}=o_P(1).
\label{eq:density-relative-rate}
\end{equation}
On the event $\min_iN_{i,n}>0$, define
\begin{equation}
 \mathcal C_{i,n}(\tau)=\left[
 \bar q_{i,n}(\tau)\ -\
 \frac{c_{1-\alpha,p}^{\mathrm{loc}}}
 {N_{i,n}^{1/2}\widehat d_i(\tau)},
 \bar q_{i,n}(\tau)\ +\
 \frac{c_{1-\alpha,p}^{\mathrm{loc}}}
 {N_{i,n}^{1/2}\widehat d_i(\tau)}\right].
\label{eq:local-confidence-band}
\end{equation}
Set $\mathcal C_{i,n}(\tau)=\RR$ for every $i,\tau$ on the complementary
event. Under \eqref{eq:local-rate-conditions},
\[
 \PP\{q(z_i,\tau)\in\mathcal C_{i,n}(\tau)
 \text{ for every }i\le p,\ \tau_0\le\tau\le\tau_1\}
 =1-\alpha+o(1).
\]
The event $\min_iN_{i,n}>0$ has probability $1-o(1)$. A uniformly
consistent kernel or quantile-spacing estimator can be used for
$\widehat d_i$. An estimator of $d_{i,h}$ is equally valid under
\eqref{eq:local-rate-conditions}.
\end{corollary}

\section{Simulation Study and Empirical Illustration}\label{sec:numerical}

\subsection{Simulation Study}

In this section, we present a Monte Carlo study of the Gaussian approximation
in Theorem~\ref{thm:main} and the online bootstrap procedure in
Section~\ref{sec:bootstrap}. We assume that $X_1,\ldots,X_n$ are i.i.d.\
draws from either the standard normal distribution or the $t$-distribution
with $10$ degrees of freedom. Quantiles are estimated using the averaged
smoothed SGD algorithm over the grid
$\mathcal G=\{0.1,0.2,\ldots,0.9\}$.

For inference based on the Brownian-bridge approximation, we use the
density-studentized statistic
\begin{align*}
\max_{\tau\in\mathcal G}
n^{1/2}
\left|
f(Q(\tau))
\{\bar Y_n(\tau)-Q(\tau)\}
\right|,
\end{align*}
with critical values obtained by simulating
$\max_{\tau\in\mathcal G}|\mathcal B(\tau)|$. In contrast, the online
bootstrap uses the unstudentized statistic
\begin{align*}
T_{\mathcal G}
=
\max_{\tau\in\mathcal G}
n^{1/2}
\left|
\bar Y_n(\tau)-Q(\tau)
\right|,
\end{align*}
with critical values obtained from the raw bootstrap differences. Therefore,
the online bootstrap does not require the density or a density estimator.

We first report the simulation results based on the critical values from the asymptotic distribution, i.e., from the maximum of Brownian bridges. Throughout, we set the hyperparameters of the algorithm to $a=1$ and $c_\gamma=3$, with the starting value $y=0$. All results are averaged over $2000$ Monte Carlo iterations. The empirical size for Gaussian data and different sample sizes and significance levels is given in Table~\ref{table:simulation_gaussian}. The results show that the empirical size crucially depends on the learning rate parameter $\beta$. For $\beta=2/3$, the empirical size is close to the nominal level across all sample sizes and significance levels. This choice for the learning rate parameter also provided the optimal bound in Remark~\ref{remark:dimension}. In contrast, the choice $\beta=1/2$, which is not covered by our theory, leads to an oversized test for large $n$. Choosing $\beta=3/4$ results in over-rejection for small $n$, with the distortion vanishing as $n$ grows. Similar observations can be made for the simulation results based on $t$-distributed data in Table~\ref{table:simulation_t} in Section~\ref{sec:numerical_app} in the supplement.

In the previous results, the sparsity function, i.e., the density evaluated at the $\tau$-quantile, $f(Q(\tau))$, is assumed to be known. As a robustness check, we consider the case in which this quantity is estimated via kernel density estimation. The results in Table~\ref{table:simulation_kde} of the supplement show that for large $n$ the coverage is still close to the nominal level.
Additionally, we report QQ plots to visualize how well the empirical distribution of the test statistic is approximated by its asymptotic distribution. See Figures~\ref{figure:qq} and \ref{figure:qq2} in Appendix~\ref{sec:numerical_app} for standard normal and $t$-distributed data under known as well as estimated sparsity. The results show that the asymptotic distribution matches the empirical distribution of the test statistic very well.

\begin{table}[tbp]
\footnotesize
\centering
\begin{tabular}{r|rrrr|rrrr|rrrr}
	\hline
  \multicolumn{1}{r}{} & \multicolumn{4}{c}{$\beta=1/2$} & \multicolumn{4}{c}{$\beta=2/3$} & \multicolumn{4}{c}{$\beta=3/4$} \\
  $n$ & $15\%$ & $10\%$ & $5\%$ & $1\%$ & $15\%$ & $10\%$ & $5\%$ & $1\%$ & $15\%$ & $10\%$ & $5\%$ & $1\%$ \\
  \hline
  250 & 0.152 & 0.096 & 0.044 & 0.009 & 0.162 & 0.102 & 0.052 & 0.007 & 0.178 & 0.117 & 0.062 & 0.014 \\ 
  500 & 0.158 & 0.100 & 0.048 & 0.010 & 0.175 & 0.116 & 0.059 & 0.010 & 0.192 & 0.135 & 0.068 & 0.020 \\ 
  1000 & 0.161 & 0.104 & 0.054 & 0.011 & 0.154& 0.103 & 0.051 & 0.014 & 0.172 & 0.111 & 0.056 & 0.016 \\ 
  2000 & 0.186 & 0.117 & 0.064 & 0.010 & 0.162 & 0.112 & 0.058 & 0.012 & 0.172 & 0.116 & 0.056 & 0.006 \\ 
  4000 & 0.174 & 0.112 & 0.052 & 0.010 & 0.161 & 0.102 & 0.050 & 0.011 & 0.169 & 0.108 & 0.065 & 0.017 \\ 
  8000 & 0.196 & 0.130 & 0.068 & 0.011 & 0.151 & 0.096 & 0.046 & 0.009 & 0.152 & 0.100 & 0.056 & 0.013 \\ 
 \hline
\end{tabular}
\caption{\footnotesize Empirical size of the asymptotic test for standard normal data, at nominal levels $\alpha\in\{15\%,10\%,5\%,1\%\}$ (columns), for learning rates $\beta\in\{1/2,2/3,3/4\}$ and increasing sample size $n$ (rows). Critical values are taken from the maximum of the approximating Brownian bridges, and the sparsity $f(Q(\tau))$ is treated as known. Results are averaged over $2000$ Monte Carlo replications.}
\label{table:simulation_gaussian}
\end{table}

Next, we examine the finite-sample performance of the density-estimation-free
online bootstrap. For each Monte Carlo sample and bootstrap replication
$b=1,\ldots,B$, we compute
\begin{align*}
T_{\mathcal G}^{*(b)}
=
\max_{\tau\in\mathcal G}
n^{1/2}
\left|
\bar Y_n^{*(b)}(\tau)-\bar Y_n(\tau)
\right|.
\end{align*}
Let $\widehat c_{1-\eta}^*$ denote the empirical $(1-\eta)$-quantile of
$\{T_{\mathcal G}^{*(b)}:1\le b\le B\}$. The null hypothesis is rejected
whenever $T_{\mathcal G}>\widehat c_{1-\eta}^*$.

We set $c_\gamma=2$, $a=4$, $y=0$, and $\beta=2/3$. The multiplier weights
are i.i.d.\ draws from the two-point distribution
$\PP(W_k=0)=\PP(W_k=2)=1/2$. The results are based on $B=1000$ bootstrap
replications and $1000$ Monte Carlo iterations.
Table~\ref{table:simulation_bootstrap_uniform} reports the corresponding empirical size results for both Gaussian and $t_{10}$-distributed data. Overall, the bootstrap performs well for large sample sizes, whereas the empirical size is above the nominal level for small sample sizes. A further aspect to keep in mind is the computational cost from running $B$ perturbed versions of the smoothed SGD algorithm in parallel.


\begin{table}[tbp]
\centering
\begin{tabular}{r|rrrr|rrrr}
	\hline 
  \multicolumn{1}{r}{} & \multicolumn{4}{c}{Normal distribution} & \multicolumn{4}{c}{$t_{10}$-distribution} \\
  $n$ & $15\%$ & $10\%$ & $5\%$ & $1\%$ & $15\%$ & $10\%$ & $5\%$ & $1\%$ \\
  \hline
  250 & 0.289 & 0.194 & 0.100 & 0.016 & 0.226 & 0.139 & 0.062 & 0.006  \\ 
  500 & 0.266 & 0.165 & 0.087 & 0.019 & 0.222 & 0.144 & 0.065 & 0.013 \\ 
  1000 & 0.237 & 0.153 & 0.079 & 0.018 & 0.201 & 0.138 & 0.068 & 0.014 \\ 
  2000 & 0.196 & 0.146 & 0.081 & 0.018 & 0.160 & 0.113 & 0.050 & 0.012 \\ 
  4000 & 0.158 & 0.114 & 0.060 & 0.017 & 0.164 & 0.110 & 0.059 & 0.014 \\ 
  8000 & 0.157 & 0.103 & 0.041 & 0.008 & 0.164 & 0.114 & 0.053 & 0.010  \\ 
 \hline 
\end{tabular}
\caption{\footnotesize Empirical size of the online multiplier bootstrap test, at nominal levels $\alpha\in\{15\%,10\%,5\%,1\%\}$ (columns), for standard normal and $t_{10}$ data (column blocks) and increasing sample size $n$ (rows). Critical values are the empirical $(1-\alpha)$-quantiles of $B=1000$ bootstrap replications, with $\beta=2/3$. Results are averaged over $1000$ Monte Carlo replications.}
\label{table:simulation_bootstrap_uniform}
\end{table}

We also assess the conditional quantile procedure of
Section~\ref{sec:conditional} in an additional simulation. The simulation
design and QQ plots are reported in Appendix~\ref{sec:numerical_app}. 
Figure~\ref{figure:cond} indicates that the Gaussian approximation is accurate,
particularly at the larger sample size.
\subsection{Empirical Illustration}

We want to further illustrate our nonparametric online quantile regression procedure of Section~\ref{sec:conditional} on financial data. We take the daily closing value of the S\&P~500 index and of the CBOE volatility index (VIX) from the FRED database of the Federal Reserve Bank of St.\ Louis over the ten-year period ending on August 10, 2026, yielding $n=2510$ trading days. The response is the daily log return in percent, and the covariate is the lagged volatility index, so that $Z_k$ is predetermined and $q(z,\tau)$ for small $\tau$ is a one-day ahead value-at-risk conditional on the current volatility state. The localization is carried out using grid points $z_i\in\{12,16,\ldots,28\}$, which cover calm as well as elevated regimes of the sample period. We consider $\tau\in\mathcal{G}=\{0.1,\ldots,0.9\}$, with hyperparameters $\beta=2/3$, $a=1$, $c_\gamma=3$, and $h_n=1.9n^{-1/4}$. This bandwidth choice ensures that the windows are disjoint as required by Assumption~\ref{asm:local-quantile}(i). Since the distribution of the volatility index is right-skewed, the effective sample sizes $N_{i,n}$ are not balanced across the design points and the bands are wider at the elevated states. As in the simulations under estimated sparsity, $d_i(\tau)$ is obtained by kernel density estimation on the observations of the $i$-th window, evaluated at $\bar q_{i,n}(\tau)$.
 
We test the location-scale hypothesis,
\begin{equation}
 H_0:\quad q(z_i,\tau)=\mu(z_i)+\sigma(z_i)\Phi^{-1}(\tau),
 \qquad i\le p,\ \tau\in\mathcal{G},
\label{eq:empirical-null}
\end{equation}
which lets the scale of the return distribution depend on the volatility state but forces the same Gaussian shape at every state. Being a restriction at all design points and all quantile levels simultaneously, \eqref{eq:empirical-null} is of the form covered by Corollary~\ref{cor:local-feasible-bands}, and we reject for large values of
\[
 T_n=\max_{i\le p}\max_{\tau\in\G}
 N_{i,n}^{1/2}\widehat d_i(\tau)
 \bigl|\bar q_{i,n}(\tau)-\widehat\mu(z_i)-\widehat\sigma(z_i)\Phi^{-1}(\tau)\bigr|,
\]
where $\widehat\mu(z_i)$, $\widehat\sigma(z_i)$ are estimates of the conditional mean and volatility within window $i$.
 
Table~\ref{table:deviations} in the Supplementary Materials reports the studentized absolute deviations for each $i$ and $\tau$. The maximum value is $T_n = 1.920$ and the critical value at the $5\%$ significance level is $1.444$. The null hypothesis \eqref{eq:empirical-null} can thus be rejected at the $5\%$ significance level with a $p$-value of $0.002$. See Figure~\ref{figure:cond_tau_bands} in the Supplementary Material for the estimated curves with uniform $95\%$-confidence bands \eqref{eq:local-confidence-band}. The effect of the volatility index seems not to be restricted to the scale of the return distribution. The volatility regime affects the shape as well as the dispersion of the distribution. Fixed Gaussian quantiles with a volatility forecast seem to be misspecified and the conditional quantile curves are better estimated directly. Two qualifications are in order. First, since $\widehat\mu$ and $\widehat\sigma$ converge at the same rate as $\bar q_{i,n}$, their estimation error is not asymptotically negligible, which is ignored by the Brownian-bridge critical values. Second, daily returns are likely to be serially dependent, whereas our Gaussian approximation result is based on i.i.d.\ data.

\section{Conclusion}\label{sec:conclusion}

We proposed a modified SGD algorithm for quantile estimation based on smoothing the quantile score function. The algorithm is computationally and memory efficient and, unlike the standard SGD quantile recursion, is monotone in the quantile level. For the averaged version we derived a Bahadur representation and a Gaussian approximation result which hold uniformly over the quantile level and over the coordinates, thereby allowing simultaneous inference even when the dimension grows exponentially in the sample size. We further provided sharpened non-asymptotic tail probability bounds for both the averaged and the non-averaged version. As an alternative that avoids estimating the sparsity function, we proposed an online multiplier bootstrap which again preserves monotonicity and is valid conditionally on the data. The theory was further extended to a localized recursion in which each design point is updated only by the observations falling into its window, yielding an online estimator of nonparametric conditional quantiles that is again monotone in the quantile level. The associated Gaussian approximation is simultaneous over design points and quantile levels and delivers feasible uniform confidence bands for the conditional quantile curves. Simulation results indicate a good finite-sample performance in both the unconditional and the conditional setting, and an application to
conditional value-at-risk curves rejected a location-scale specification of returns.

\subsection*{Data Availability Statement}\label{data-availability-statement}

The daily closing values of the S\&P~500 index (\texttt{SP500}) and of the CBOE volatility index (\texttt{VIXCLS}) used in the empirical illustration in Section~\ref{sec:numerical} were downloaded from the FRED database of the Federal Reserve Bank of St.\ Louis, \url{https://fred.stlouisfed.org/}.

{\footnotesize
\onehalfspacing
\bibliography{literature}
}


\newpage
\setcounter{page}{1}
\appendix

\phantomsection\label{supplementary-material}
\bigskip

\begin{center}

{\large\bf SUPPLEMENTARY MATERIAL}

\end{center}

\section{Proofs for Results in Sections~\ref{sec:smoothed} and \ref{sec:tailprobability}}
\label{sec:proof_sec3}

\subsection{Proof of Lemma~\ref{lem:monotone}}

\begin{proof}
We prove the result by induction on \(k\). For \(k=1\), the claim holds since $Y_{i,1}(\tau)=Y_{i,1}(\tau')=y_i$.
Now suppose that for all \(\tau\ge \tau'\) and some \(k\ge 1\),
$Y_{i,k}(\tau)\ge Y_{i,k}(\tau')$.
Then for $k+1,$ by the recursion \eqref{eq:sgdcont},
\begin{align}
&Y_{i,k+1}(\tau)-Y_{i,k+1}(\tau') \nonumber\\
&=
Y_{i,k}(\tau)-Y_{i,k}(\tau')
+\gamma_k(\tau-\tau') \nonumber\\
&\quad
-\gamma_k\Big[
g_{a\gamma_k}(Y_{i,k}(\tau)-X_{i,k+1})
-
g_{a\gamma_k}(Y_{i,k}(\tau')-X_{i,k+1})
\Big].
\label{eq:gkytau}
\end{align}
Since \(g_{a\gamma_k}\) is Lipschitz continuous with Lipschitz constant
\((2a\gamma_k)^{-1}\), and since
\(Y_{i,k}(\tau)\ge Y_{i,k}(\tau')\), we have
$$|g_{a\gamma_k}(Y_{i,k}(\tau)-X_{i,k+1})-g_{a\gamma_k}(Y_{i,k}(\tau')-X_{i,k+1})|\leq (2a\gamma_k)^{-1}(Y_{i,k}(\tau)-Y_{i,k}(\tau')).$$
Substituting this bound into \eqref{eq:gkytau} gives
\begin{align*}
Y_{i,k+1}(\tau)-Y_{i,k+1}(\tau')
&\ge
\gamma_k(\tau-\tau')
+
\left(1-\frac{1}{2a}\right)
(Y_{i,k}(\tau)-Y_{i,k}(\tau')).
\end{align*}
Because \(\tau\ge \tau'\), \(a\geq 1\), and by the induction hypothesis
\(Y_{i,k}(\tau)-Y_{i,k}(\tau')\ge 0\), the right-hand side is nonnegative.
Thus we complete the proof.
\end{proof}

\subsection{Proof of Theorem~\ref{thm:singletail}}

\begin{proof}
Recall the distribution function of $X_{i,k}$ is $F_i(\cdot).$ Then the moment generating function takes the form
\begin{align*}
&\EE(e^{t(Y_{i,k+1}(\tau)-Q_i(\tau))})\\
=&\EE\Big\{ e^{t(Y_{i,k}(\tau)-Q_i(\tau)+\gamma_k\tau)}\1_{Y_{i,k}(\tau)-X_{i,k+1}<-a\gamma_k} + 
e^{t(Y_{i,k}(\tau)-Q_i(\tau)-\gamma_k(1-\tau))}\1_{Y_{i,k}(\tau)-X_{i,k+1}\geq a\gamma_k } \\
&+e^{t\big(Y_{i,k}(\tau)-Q_i(\tau)+\gamma_k\tau-(Y_{i,k}(\tau)-X_{i,k+1}+a\gamma_k)/(2a)\big)}\1_{ -a\gamma_k\leq Y_{i,k}(\tau)-X_{i,k+1}<a\gamma_k} \Big\}\\
\leq &\EE\Big\{ e^{t(Y_{i,k}(\tau)-Q_i(\tau))}\big[ e^{t\gamma_k\tau}(1-F_i(Y_{i,k}(\tau)+a\gamma_k)) + e^{-t\gamma_k(1-\tau)}F_i(Y_{i,k}(\tau)-a\gamma_k )\\
&+e^{t\gamma_k(\tau+1)}\1_{ -a\gamma_k\leq Y_{i,k}(\tau)-X_{i,k+1}<a\gamma_k} \big]\Big\},
\end{align*}
where the last inequality is because $X_{i,k+1}$ is independent of $Y_{i,k}(\tau)$ and $|t(Y_{i,k}(\tau)-X_{i,k+1}+a\gamma_k)/(2a)|$ is bounded by $t\gamma_k$ when $-a\gamma_k\leq Y_{i,k}(\tau)-X_{i,k+1}<a\gamma_k$. Since $|f_i|_\infty\leq c_f,$
\begin{align*}
&\EE\Big(e^{t(Y_{i,k}(\tau)-Q_i(\tau))+t\gamma_k(\tau+1)}\1_{ -a\gamma_k\leq Y_{i,k}(\tau)-X_{i,k+1}<a\gamma_k} \Big)\\ 
\leq &\EE\Big(e^{t(Y_{i,k}(\tau)-Q_i(\tau))+t\gamma_k(\tau+1)}\big[F_i(Y_{i,k}(\tau)+a\gamma_k)-F_i(Y_{i,k}(\tau)-a\gamma_k)\big]\Big)\\
\leq &\EE\Big(e^{t(Y_{i,k}(\tau)-Q_i(\tau))+t\gamma_k(\tau+1)}\Big)2c_fa\gamma_k.
\end{align*}
Hence by Taylor's expansion, we further have
\begin{align}
\label{eq:eqforetykp1}
\EE(e^{t(Y_{i,k+1}(\tau)-Q_i(\tau))})
\leq &\EE\Big\{ e^{t(Y_{i,k}(\tau)-Q_i(\tau))}\Big[ e^{t\gamma_k\tau}(1-F_i(Y_{i,k}(\tau))) + e^{-t\gamma_k(1-\tau)}F_i(Y_{i,k}(\tau))\Big]\Big\}\nonumber\\
+&c_f\EE\big[e^{t(Y_{i,k}(\tau)-Q_i(\tau))}(e^{t\gamma_k\tau}+e^{-t\gamma_k(1-\tau)})\big]a\gamma_k\nonumber\\
&+2c_f\EE\big(e^{t(Y_{i,k}(\tau)-Q_i(\tau))}\big)e^{t\gamma_k(\tau+1)}a\gamma_k.
\end{align}
Since $e^{t\gamma_k\tau}-e^{-t\gamma_k(1-\tau)}\geq 0,$  the function
$$e^{t\gamma_k\tau}(1-F_i(x)) + e^{-t\gamma_k(1-\tau)}F_i(x)
=e^{t\gamma_k\tau}-F_i(x)\big(e^{t\gamma_k\tau} - e^{-t\gamma_k(1-\tau)}\big)
$$ 
is monotonic decreasing on $x\in\RR$ and is upper bounded by $e^{t\gamma_k\tau}.$ Take $c_1=c_f(1+2e+2e^2)a$, $c_1'=4c_1c_L^{-1}$ and $L=c_1'/t$. 
Hence by \eqref{eq:eqforetykp1}, for any $k\geq (c_\gamma t)^{1/\beta}$ we have that $t\gamma_k\leq 1$ and 
\begin{align}
\label{eq:temp1}
&\EE(e^{t(Y_{i,k+1}(\tau)-Q_i(\tau))})\nonumber\\
\leq &\EE\Big\{ e^{t(Y_{i,k}(\tau)-Q_i(\tau))}\big[ e^{t\gamma_k\tau}(1-F_i(L+Q_i(\tau))) + e^{-t\gamma_k(1-\tau)}F_i(L+Q_i(\tau)) \big]\1_{Y_{i,k}(\tau)\geq L+Q_i(\tau)}\nonumber\\
&\qquad+e^{tL+t\gamma_k\tau}\1_{Y_{i,k}(\tau)< L+Q_i(\tau)}\Big\}+c_1\gamma_k\EE(e^{t(Y_{i,k}(\tau)-Q_i(\tau))})\nonumber\\
\leq &\EE\Big\{ e^{t(Y_{i,k}(\tau)-Q_i(\tau))}\big[ e^{t\gamma_k\tau}(1-F_i(L+Q_i(\tau))) + e^{-t\gamma_k(1-\tau)}F_i(L+Q_i(\tau)) +c_1\gamma_k\big]\Big\}+
c_0,    
\end{align}
where $c_0=e^{c_1'+1}$. Since $t\gamma_k\tau<1,$ 
by Taylor's expansion, 
$e^{t\gamma_k\tau}\leq 1+t\gamma_k\tau+t^2\gamma_k^2\tau^2.$ Then we have
\begin{align}
\label{eq:bddforexptgtau}
&e^{t\gamma_k\tau}(1-F_i(L+Q_i(\tau))) + e^{-t\gamma_k(1-\tau)}F_i(L+Q_i(\tau))\nonumber\\
\leq &(1+t\gamma_k\tau+t^2\gamma_k^2)(1-F_i(L+Q_i(\tau)))+(1-t\gamma_k(1-\tau)+t^2\gamma_k^2)F_i(L+Q_i(\tau))\nonumber\\
\leq & 1-t\gamma_k\big[F_i(L+Q_i(\tau))-\tau\big]+t^2 \gamma_k^2\nonumber\\
\leq & 1-t\gamma_k\big(f_i(Q_i(\tau))-c_fL\big)L+t^2 \gamma_k^2,
\end{align}
where the last inequality is due to the fact that $\tau=F_i(Q_i(\tau))$ and 
\begin{equation*}
    F_i(L+Q_i(\tau))-F_i(Q_i(\tau))-f_i(Q_i(\tau))L \geq -|f_i'|_\infty L^2.
\end{equation*}
Recall that $L=c_1'/t$. Inserting above into \eqref{eq:temp1}, for
$t\geq 2c_fc_1'/c_L$, we have $$f_i(Q_i(\tau))-c_fL\geq c_L-c_fc_1'/t\geq c_L/2$$ 
and thus
\begin{align*}
&\EE(e^{t(Y_{i,k+1}(\tau)-Q_i(\tau))})\\ 
\leq &
\EE(e^{t(Y_{i,k}(\tau)-Q_i(\tau))})
[1-t\gamma_k\big(f_i(Q_i(\tau))-c_fL\big)L+t^2 \gamma_k^2+c_1\gamma_k]+c_0\\
\leq & \EE(e^{t(Y_{i,k}(\tau)-Q_i(\tau))})(1-c_1\gamma_k+t^2 \gamma_k^2)+c_0,
\end{align*}
where the last inequality is due to
$t(c_L/2)L-c_1=c_1'c_L/2-c_1=c_1.$ Recursively applying above inequality, for $k_t=\lceil (c_\gamma t)^{1/\beta} \rceil,$ we have
\begin{align}
\label{eq:temphaha}
 \EE(e^{t(Y_{i,n+1}(\tau)-Q_i(\tau))})
 \leq c_0\Big(1+\sum_{k=k_t+1}^n \phi_k\Big)+\phi_{k_t}\EE(e^{t(Y_{i,k_t}(\tau)-Q_i(\tau))}),
 \end{align}
 where
 \begin{align}
  \phi_k=\prod_{l=k}^n(1-c_1\gamma_l+t^2 \gamma_l^2).
\end{align}
Let $c_2=c_1c_\gamma(1-(1/2)^{1-\beta})/(1-\beta)$ and
\begin{align}
\label{eq:defofc}
c=\min\big\{ ((2\beta-1)c_2/2)^\beta/c_\gamma,  (c_1/c_\gamma)^{1/2}2^{-\beta-1/2}, ((1-\beta)c_2/4)^\beta/c_\gamma, c_2/(4(M+c_y))\big\}.    
\end{align}
Then by the same argument as for (18) in the proof of Theorem 3.1 in \citet{chen2023recursive}, for $2c_fc_1'/c_L\leq t\leq cn^{\beta(1-\beta)},$ we have
\begin{align}
\label{eq:midetyintaubdd}
\EE\big(e^{t|Y_{i,n}(\tau)-Q_i(\tau)|}\big)  \leq c'n^\beta,  
\end{align}
where 
\begin{align}
\label{eq:defofcprim}
c'=c_0c_3+(c_0+1)(c_0c_3),  \,\,  c_3=1/c_2+2/(c_1c_\gamma).
\end{align}
By Jensen's inequality, for any $0\leq t_1<t$,
\begin{align}
\label{eq:markovthm2}
 \EE\big(e^{t_1|Y_{i,n}(\tau)-Q_i(\tau)|}\big)
=\EE\big(e^{t|Y_{i,n}(\tau)-Q_i(\tau)|(t_1/t)}\big)\leq \big[\EE\big(e^{t|Y_{i,n}(\tau)-Q_i(\tau)|}\big)\big]^{t_1/t}.
\end{align}
Hence \eqref{eq:midetyintaubdd} holds for any $0\leq t\leq cn^{\beta(1-\beta)}.$ 

By Markov's inequality with $t=cn^{\beta(1-\beta)}$, we have
    \begin{align*}
    \PP( |Y_{i,n}(\tau)-Q_i(\tau)|\geq x )
    &\leq e^{-tx}\EE\Big[\exp\big\{t|Y_{i,n}(\tau)-Q_i(\tau)|\big\}\Big]\leq c'n^\beta e^{-cn^{\beta(1-\beta)}x}.
    \end{align*}
Thus we complete the proof.
\end{proof}

\subsection{Proof of Theorem~\ref{thm:tailsingleonerefine}}
We first record a tail bound, stated for the multiplier bootstrap
recursion~\eqref{eq:bootstrap}, that will be used in the proof of
Theorem~\ref{thm:tailsingleonerefine}.

\begin{lemma}
\label{lem:excursion}
Under Assumption~\ref{asmp:condf}, let $\eta_0=c_L/(4c_f).$ For any fixed $0<\eta\leq \eta_0,$ there exist constants $c_\eta, C_\eta>0$ independent of $n,p,i,\tau$ such that
\begin{align}
\label{eq:excursion}
\PP\Big(\max_{n/2\leq k\leq n}|Y^*_{i,k}(\tau)-Q_i(\tau)|>\eta\Big)
\leq C_\eta e^{-c_\eta n^{\min\{2(1-\beta),\beta\}}}.
\end{align}
Since the recursion \eqref{eq:sgdcont} is the special case $W_k\equiv 1$ of \eqref{eq:bootstrap}, the bound \eqref{eq:excursion} also holds with $Y^*_{i,k}(\tau)$ replaced by $Y_{i,k}(\tau).$
\end{lemma}
 
\begin{proof}
Write $e^*_k=Y^*_{i,k}(\tau)-Q_i(\tau).$ Since $W_{k+1}$ is independent of $(X_{i,k+1},\F^*_{i,k})$ and $\EE W_{k+1}=1,$ by \eqref{eq:bootstrap} and \eqref{eq:xiik1star},
\begin{align}
\label{eq:excursionincrement}
e^*_{k+1}-e^*_k=\gamma_k\big(\tau-G_{i,k}(Y^*_{i,k}(\tau))\big)+\gamma_k\xi^*_{i,k+1}(\tau),
\end{align}
where $\xi^*_{i,k+1}(\tau),$ $k\geq 1,$ are martingale differences with respect to $\F^*_{i,k}$ with $|\xi^*_{i,k+1}(\tau)|\leq 2a+1,$ and $|e^*_{k+1}-e^*_k|\leq 2a\gamma_k.$
 
By Assumption~\ref{asmp:condf}, $f_i(y)\geq f_i(Q_i(\tau))-c_f|y-Q_i(\tau)|\geq c_L/2$ for $|y-Q_i(\tau)|\leq 2\eta_0.$ Hence, for $0<\eta\leq\eta_0,$
\begin{align*}
F_i(Q_i(\tau)+\eta/2)-\tau\geq c_L\eta/4
\quad\textrm{and}\quad
\tau-F_i(Q_i(\tau)-\eta/2)\geq c_L\eta/4.
\end{align*}
Set $\kappa_\eta=c_L\eta/8.$ Since $\sup_{x\in\RR}|F_i(x)-G_{i,k}(x)|\leq 2ac_f\gamma_k,$ there exists $N_\eta\in\NN,$ depending only on $\eta,a,c_f,c_\gamma,\beta,c_L,$ such that for all $k\geq N_\eta$ we have $2a\gamma_k\leq \eta/4,$ $2ac_f\gamma_k\leq \kappa_\eta,$ and consequently
\begin{align}
\label{eq:driftsign}
&e^*_k\geq \eta/2 \;\Longrightarrow\; \tau-G_{i,k}(Y^*_{i,k}(\tau))\leq -\kappa_\eta,\nonumber\\
&e^*_k\leq -\eta/2 \;\Longrightarrow\; \tau-G_{i,k}(Y^*_{i,k}(\tau))\geq \kappa_\eta.
\end{align}
In the following, assume $n$ is large enough such that $n/2\geq 2N_\eta.$
 
Fix $n/2\leq m\leq n$ and consider the upper tail $\PP(e^*_m>\eta).$ The lower tail is handled symmetrically. Decompose according to the last time before $m$ at which the process is at most $\eta/2$ with
\begin{align*}
B_s&=\{e^*_s\leq \eta/2\}\cap\bigcap_{j=s+1}^{m}\{e^*_j>\eta/2\}\cap\{e^*_m>\eta\},
\qquad N_\eta<s\leq m-1,\\
B_{N_\eta}&=\bigcap_{j=N_\eta+1}^{m}\{e^*_j>\eta/2\}\cap\{e^*_m>\eta\},
\end{align*}
we have $\{e^*_m>\eta\}= \bigcup_{s=N_\eta}^{m-1}B_s.$ On $\{e^*_m>\eta\}$ let $s$ be the
largest $j\in[N_\eta,m]$ with $e^*_j\le\eta/2$, and $s:=N_\eta$ if no such $j$ exists.
Then $s\le m-1$ since $e^*_m>\eta$, and the event lies in $B_s$.

For $s>N_\eta,$ on $B_s$ we have $e^*_{s+1}\leq e^*_s+2a\gamma_s\leq 3\eta/4,$ and by \eqref{eq:excursionincrement} and \eqref{eq:driftsign}, for all $s+1\leq j\leq m-1$ the drift term is at most $-\kappa_\eta.$ Hence on $B_s,$
\begin{align*}
e^*_m\leq 3\eta/4+\sum_{j=s+1}^{m-1}\gamma_j\xi^*_{i,j+1}(\tau)-\kappa_\eta\Gamma_s,
\qquad
\Gamma_s=\sum_{j=s+1}^{m-1}\gamma_j,
\end{align*}
so that
\begin{align}
\label{eq:Bsinclusion}
B_s\subseteq\Big\{\sum_{j=s+1}^{m-1}\gamma_j\xi^*_{i,j+1}(\tau)\geq \eta/4+\kappa_\eta\Gamma_s\Big\}.
\end{align}
For $s=N_\eta,$ the same argument applies with $e^*_{s+1}$ bounded deterministically: since $Y^*_{i,1}(\tau)=0,$ $|e^*_1|\leq M$ and $|e^*_{N_\eta+1}|\leq M+2a\sum_{j\leq N_\eta}\gamma_j=:C'_\eta,$ so that \eqref{eq:Bsinclusion} holds with $\eta/4$ replaced by $\eta-C'_\eta.$ Since $\Gamma_{N_\eta}\gtrsim n^{1-\beta}\to\infty$ while $C'_\eta$ is a constant, the right-hand side is at least $\kappa_\eta\Gamma_{N_\eta}/2$ for all $n$ large, and this only changes constants below.

For each fixed $s,$ the right-hand side of \eqref{eq:Bsinclusion} concerns a martingale with bounded increments $|\gamma_j\xi^*_{i,j+1}(\tau)|\leq (2a+1)\gamma_j,$ so Azuma's inequality gives
\begin{align}
\label{eq:azumaBs}
\PP(B_s)\leq \exp\Big\{-\frac{(\eta/4+\kappa_\eta\Gamma_s)^2}{2(2a+1)^2\sum_{j=s+1}^{m-1}\gamma_j^2}\Big\}.
\end{align}
If $s\leq m/2,$ then $\Gamma_s\gtrsim m^{1-\beta}\gtrsim n^{1-\beta},$ while $\sum_{j=s+1}^{m-1}\gamma_j^2\leq \sum_{j\geq 1}\gamma_j^2<\infty$ since $\beta>1/2,$ and \eqref{eq:azumaBs} yields $\PP(B_s)\leq e^{-cn^{2(1-\beta)}}.$ If $s>m/2,$ write $L=m-s.$ For $L=1$ we have $B_s=\emptyset,$ since on $B_s,$ $e^*_m\leq e^*_{m-1}+2a\gamma_{m-1}\leq 3\eta/4<\eta.$ For $L\geq 2,$ the sums in \eqref{eq:Bsinclusion} have $L-1\geq L/2$ terms with $\gamma_j\asymp n^{-\beta},$ so $\Gamma_s\gtrsim Ln^{-\beta}$ and $\sum_{j=s+1}^{m-1}\gamma_j^2\lesssim Ln^{-2\beta},$ and the exponent in \eqref{eq:azumaBs} is bounded below by a constant times
\begin{align*}
\frac{(\eta+\kappa_\eta Ln^{-\beta})^2}{Ln^{-2\beta}}
\geq \max\Big\{\frac{\eta^2n^{2\beta}}{L},\;\kappa_\eta^2L\Big\}
\geq \eta\kappa_\eta n^{\beta},
\end{align*}
so that $\PP(B_s)\leq e^{-cn^{\beta}}.$ Summing over $N_\eta\leq s\leq m-1$ and $n/2\leq m\leq n$ produces a factor at most $n^2,$ which is absorbed into the constants since $n^2e^{-cn^{\min\{2(1-\beta),\beta\}}}\lesssim e^{-c'n^{\min\{2(1-\beta),\beta\}}}.$ This completes the proof.
\end{proof}

We proceed with the proof of Theorem~\ref{thm:tailsingleonerefine}.
\begin{proof}
The SGD iteration \eqref{eq:sgdcont} can be rewritten into
\begin{align*}
Y_{i,k+1}(\tau)=Y_{i,k}(\tau)+\gamma_k Z_{i,k+1}(\tau),   
\end{align*}
where
\begin{align}
\label{eq:zik1}
Z_{i,k+1}(\tau)=\tau-g_{a\gamma_k}(Y_{i,k}(\tau)-X_{i,k+1}).
\end{align}
The term $Z_{i,k+1}$ can be further decomposed into three parts: a martingale difference part, a linear part and the remainder. Let $\F_{i,k}=\sigma(X_{i,k}, X_{i,k-1},\ldots)$ and 
\begin{align}
\label{eq:martingalediffik1}
\xi_{i,k+1}(\tau)=Z_{i,k+1}(\tau)-\EE(Z_{i,k+1}(\tau)|\F_{i,k}).    
\end{align} 
Then $\xi_{i,k+1}(\tau)$, $k\geq 1,$ are martingale differences with respect to $\F_{i,k}.$  
Furthermore, let  
\begin{align}
\label{eq:phoikreprest}
\rho_{i,k}(\tau)=\EE(Z_{i,k+1}(\tau)|\F_{i,k})+f_i(Q_i(\tau))(Y_{i,k}(\tau)-Q_i(\tau)).
\end{align} 
Here $\rho_{i,k}(\tau)$ represents the remainder term of $\EE(Z_{i,k+1}(\tau)|\F_{i,k})$ after subtracting the linear part $-f_i(Q_i(\tau))(Y_{i,k}(\tau)-Q_i(\tau)).$
  Combining all the above, we have the desired decomposition
\begin{align}
\label{eq:decompyiktau}
&Y_{i,k+1}(\tau)-Q_i(\tau)\nonumber\\
=&Y_{i,k}(\tau)-Q_i(\tau)
+\gamma_kZ_{i,k+1}(\tau)\nonumber\\
=&Y_{i,k}(\tau)-Q_i(\tau)+\gamma_k\big[\xi_{i,k+1}(\tau)-f_i(Q_i(\tau))(Y_{i,k}(\tau)-Q_i(\tau))+\rho_{i,k}(\tau)\big].      
\end{align}
By \eqref{eq:decompyiktau}, we obtain the recursion,
\begin{align}
\label{eq:ykdecomposition1}
 &Y_{i,k+1}(\tau)-Q_i(\tau)\nonumber\\
=&\big[1-\gamma_kf_i(Q_i(\tau))\big](Y_{i,k}(\tau)-Q_i(\tau))+\gamma_k \big(\xi_{i,k+1}(\tau)+\rho_{i,k}(\tau)\big).    
\end{align}
Denote $a_{i,k}=1-\gamma_kf_i(Q_i(\tau))$. Since $\gamma_k=c_\gamma k^{-\beta}$ and
$f_i(Q_i(\tau))\leq c_f$, there exists a deterministic constant
$k_0$, independent of $n,p,i$, and $\tau$, such that
$\gamma_k f_i(Q_i(\tau))\leq 1/2$ for all $k\geq k_0$.
We may therefore restart recursion \eqref{eq:ykdecomposition1} at $k_0$ and absorb the
finitely many initial iterations into the initial condition, and $1>a_{i,k}>0$ for $k\geq k_0$. Let
\[
b_{i,k}:=\prod_{j=k_0}^{k}a_{i,j}.
\]
Recursively applying \eqref{eq:ykdecomposition1} leads to
\begin{align}
\label{eq:ynrecursive1}
Y_{i,n}(\tau)-Q_i(\tau)
&=\prod_{j=k_0}^{n-1}a_{i,j} (Y_{i,k_0}(\tau)-Q_i(\tau))+
\sum_{k=k_0}^{n-1}\Big(\prod_{j=k+1}^{n-1} a_{i,j}\Big)\gamma_k(\xi_{i,k+1}(\tau)+\rho_{i,k}(\tau))\nonumber\\
&=b_{i,n-1}\bigl(Y_{i,k_0}(\tau)-Q_i(\tau)\bigr)
+
\sum_{k=k_0}^{n-1}
\gamma_k b_{i,n-1}b_{i,k}^{-1}
\bigl(\xi_{i,k+1}(\tau)+\rho_{i,k}(\tau)\bigr).
\end{align}
Since $1+x\leq e^x$ for any $x\in\RR$, by Assumption~\ref{asmp:condf}, we have $a_{i,k}\leq \exp\{-c_L\gamma_k\}$ and thus
\begin{align}
\label{eq:bknbdd1}
b_{i,k}\leq \exp\Big\{-c_L \sum_{j=k_0}^k \gamma_j\Big\}  
\quad\textrm{and}\quad b_{i,n-1}b_{i,k}^{-1}\leq \exp\Big\{-c_L \sum_{j=k+1}^{n-1} \gamma_j\Big\}. 
\end{align}
By \eqref{eq:ynrecursive1}, we can decompose $|Y_{i,n}(\tau)-Q_i(\tau)|$ into three parts,
\begin{align*}
|Y_{i,n}(\tau)-Q_i(\tau)|
\leq &b_{i,n-1} |Y_{i,k_0}(\tau)-Q_i(\tau)|+
\Big|\sum_{k=k_0}^{n-1}\gamma_k b_{i,n-1}b_{i,k}^{-1}\xi_{i,k+1}(\tau)\Big|\\
+&\sum_{k=k_0}^{n-1}\gamma_k b_{i,n-1}b_{i,k}^{-1}|\rho_{i,k}(\tau)|
=\I_1+\I_2+\I_3.   
\end{align*}
In the following, we shall bound terms $\I_1$--$\I_3$ respectively.  
For part $\I_1,$ since
$|Y_{i,k_0}(\tau)-Q_i(\tau)|\leq c_y+M+\sum_{j=1}^{k_0-1}\gamma_j=:c_{y,0}$
is a constant independent of $n,p,i,\tau$, by \eqref{eq:bknbdd1} we obtain
\begin{align*}
\I_1\leq c_{y,0}b_{i,n-1}
\leq c_{y,0} \exp\big\{-c_Lc_\gamma(1-\beta)^{-1} \big((n-1)^{1-\beta}-k_0^{1-\beta}\big)\big\}.
\end{align*}

For part $\I_2,$ let $H_{i,k}(\tau)=\gamma_k b_{i,n-1}b_{i,k}^{-1}\xi_{i,k+1}(\tau)$. Since $\xi_{i,k+1}(\tau),$ $k_0\leq k\leq n-1,$ are martingale differences and are bounded in absolute value by $1$, by Freedman's concentration inequality,
\begin{align}
\label{eq:freedman1}
\PP\Big( \big|\sum_{k=k_0}^{n-1}H_{i,k}(\tau)\big| \geq z\Big) 
\leq 2\exp\Big\{-\frac{z^2}{2(c_Hz +V)}\Big\},
\end{align}
where 
\begin{align*}
c_H=\max_{k_0\leq k\leq n-1}|H_{i,k}(\tau)|,
\end{align*}
which by \eqref{eq:bknbdd1} is of order $O(n^{-\beta}),$ and $V$ is the upper bound for
\begin{align*}
\sigma^2&=\sum_{k=k_0}^{n-1}\EE\big(H_{i,k}^2(\tau) |\F_{i,k} \big)\leq \sum_{k=k_0}^{n-1}(\gamma_k b_{i,n-1}b_{i,k}^{-1})^2.
\end{align*}
We have $\sigma^2=O(n^{-\beta})$. Here the constants in $O(\cdot)$ and throughout this proof are independent of $n,p,\tau,i,$
Applying above to \eqref{eq:freedman1}, we obtain
\begin{align}
\label{eq:tij}
\PP(\I_2 \geq z)
\leq  2\Big(e^{-c_1n^{\beta}z}+e^{-c_1n^{\beta}z^2}\Big),
\end{align}
where $c_l, l\geq 1,$ in this proof are constants independent of $n,p,i,\tau.$

For part $\I_3,$ let
\begin{align}
G_{i,k}(x)=\EE\big(g_{a\gamma_k}(x-X_{i,k+1})\big).    
\end{align}
Then $\EE(Z_{i,k+1}|\F_{i,k})=\tau-G_{i,k}(Y_{i,k}(\tau))=F_i(Q_i(\tau))-G_{i,k}(Y_{i,k}(\tau))$ and
\begin{align}
\label{eq:gktau2}
\rho_{i,k}(\tau)&=F_i(Q_i(\tau))-G_{i,k}(Y_{i,k}(\tau))+f_i(Q_i(\tau))(Y_{i,k}(\tau)-Q_i(\tau))\nonumber\\
&=\big[F_i(Y_{i,k}(\tau)) -G_{i,k}(Y_{i,k}(\tau))\big]\nonumber\\
&\quad - \big[F_i(Y_{i,k}(\tau))-F_i(Q_i(\tau))-f_i(Q_i(\tau))(Y_{i,k}(\tau)-Q_i(\tau))\big].
\end{align}
Note that for any $x\in\RR,$
\begin{align*}
|F_i(x)-G_{i,k}(x)|
=\big|\EE\big[\1_{\{x-X_{i,k+1}\geq 0\}}-g_{a\gamma_k}(x-X_{i,k+1})\big]\big|
\leq 2a\gamma_k|f_i|_\infty.
\end{align*}
By Taylor's expansion 
\begin{align*}
|F_i(Y_{i,k}(\tau))-F_i(Q_i(\tau))-f_i(Q_i(\tau))(Y_{i,k}(\tau)-Q_i(\tau))|
\leq |f_i'|_\infty(Y_{i,k}(\tau)-Q_i(\tau))^2.
\end{align*}
Implementing above into \eqref{eq:gktau2} leads to
\begin{align}
\label{eq:rhoikbddtwoparts}
|\rho_{i,k}(\tau)|\leq 2ac_f\gamma_k+c_f(Y_{i,k}(\tau)-Q_i(\tau))^2. 
\end{align}
Let
\begin{align}
\label{eq:defofA}
A=\max\{1,(4-3\beta)/(c_Lc_\gamma)\},
\qquad
m_n=\lfloor n-An^\beta\log(n)\rfloor .
\end{align}
We further split $\I_3$ into two parts,
\begin{align*}
\sum_{k=k_0}^{n-1}\gamma_kb_{i,n-1}b_{i,k}^{-1}|\rho_{i,k}(\tau)|
=&\sum_{k=k_0}^{m_n}\gamma_kb_{i,n-1}b_{i,k}^{-1}|\rho_{i,k}(\tau)|
+\sum_{k=m_n+1}^{n-1}\gamma_kb_{i,n-1}b_{i,k}^{-1}|\rho_{i,k}(\tau)|\\
=&\I_{31}+\I_{32}.
\end{align*}
By \eqref{eq:sgdcont}, we have
\begin{align*}
|Y_{i,k}(\tau)-Q_i(\tau)|\leq |y_i-Q_i(\tau)|+\sum_{j=1}^{k-1}\gamma_j
\leq (c_y+M)+(1-\beta)^{-1}c_\gamma k^{1-\beta}.
\end{align*}
Together with \eqref{eq:rhoikbddtwoparts}, we obtain
\begin{align}
\label{eq:rhobddk}
|\rho_{i,k}(\tau)|\leq 2ac_f\gamma_k+c_f(Y_{i,k}(\tau)-Q_i(\tau))^2
\lesssim k^{2(1-\beta)},
\end{align}
where the constant in $\lesssim$ here and throughout this proof is independent
of $n,p,i,\tau$. Since $n^{1-\beta}-k^{1-\beta}\geq (1-\beta)(n-k)n^{-\beta}$
for $1\leq k\leq n$, \eqref{eq:bknbdd1} gives
\begin{align}
\label{eq:gammabbdd}
\gamma_kb_{i,n-1}b_{i,k}^{-1}
\lesssim & k^{-\beta}\exp\Big\{-c_Lc_\gamma(1-\beta)^{-1}\big(n^{1-\beta}-k^{1-\beta}\big)\Big\}\nonumber\\
\leq & k^{-\beta}\exp\big\{- c_Lc_\gamma(n-k)n^{-\beta}\big\}.
\end{align}
Hence by \eqref{eq:rhobddk}, and since $n-k\geq An^\beta\log(n)$ for $k\leq m_n$,
\begin{align*}
\I_{31}
\lesssim
\sum_{k=k_0}^{m_n}k^{2-3\beta}\exp\big\{-c_Lc_\gamma A\log(n)\big\}
\lesssim
n^{3-3\beta}n^{-c_Lc_\gamma A}
\leq n^{-1},
\end{align*}
by the choice of $A$ in \eqref{eq:defofA}. For the second part, writing
$\Delta=n-k$, we have by \eqref{eq:gammabbdd} that
\begin{align*}
\sum_{k=m_n+1}^{n-1}\gamma_kb_{i,n-1}b_{i,k}^{-1}
\lesssim
\sum_{\Delta=1}^{\lceil An^\beta\log(n)\rceil}
n^{-\beta}\exp\big\{-c_Lc_\gamma \Delta n^{-\beta}\big\}
\lesssim 1 .
\end{align*}
Therefore, using $\gamma_k\asymp n^{-\beta}$ for $k\geq m_n$ together with
\eqref{eq:rhobddk},
\begin{align*}
\I_{32}\lesssim n^{-\beta}
+\max_{m_n\leq k\leq n-1}|Y_{i,k}(\tau)-Q_i(\tau)|^2 .
\end{align*}
Combining $\I_1$-$\I_3$, we obtain that for any $1>x\geq  c n^{-\beta},$ some constant $c$ large, we have 
\begin{align}
\label{eq:sqrbdddisc}
&\PP\big(|Y_{i,n}(\tau)-Q_i(\tau)|>x\big)\nonumber\\
\leq&c_3\Big\{
\exp\big(-c_1n^{\beta}x^2\big)+
\PP\Big(\max_{m_n\leq k\leq n-1}|Y_{i,k}(\tau)-Q_i(\tau)|^2>x/c_2\Big)\Big\},
\end{align}
where $c_1,c_2,c_3$ are constants independent of $n$. Enlarging $c_3$ if
necessary, and we may assume $c_2\geq 1$ and that $c_3n^\beta\log(n)$ bounds the
number of indices in the window $[m_n,n-1]$. Set $x_0=x$ and
$x_{j+1}=(x_j/c_2)^{1/2}$, so that
\begin{align}
\label{eq:xjrec}
x_j=c_2^{-(1-2^{-j})}x^{2^{-j}}
\qquad\text{and}\qquad
x_{j+1}^2=x_j/c_2 .
\end{align}
Recursively apply the previous inequality $k_n+1$ times, where
$k_n=\log\log(n)/\log(2)$, leading to
\begin{align*}
&\PP\big(|Y_{i,n}(\tau)-Q_i(\tau)|>x\big)\\
\leq&c_3\Big(e^{-c_1n^\beta x^2}
+c_3n^\beta\log(n)e^{-c_1n^\beta x_1^2}+\ldots
+(c_3n^\beta\log(n))^{k_n}e^{-c_1n^\beta x_{k_n}^2}\Big)\\
&+(c_3n^\beta\log(n))^{k_n+1}\max_{n-A(k_n+1)n^\beta \log(n)\leq k\leq n}
\PP\big(|Y_{i,k}(\tau)-Q_i(\tau)|>x_{k_n+1}\big).
\end{align*}
Since $x<1$ and $c_2\geq 1$, \eqref{eq:xjrec} gives $x_j\geq c_2^{-1}x$ and
hence $x_{j+1}^2=x_j/c_2\geq c_2^{-2}x$. For $n^{-\beta/2}\lesssim x<1$ we have
$n^\beta x\geq n^{\beta/2}$, so each intermediate term satisfies, for
$1\leq j\leq k_n$,
\begin{align*}
(c_3n^\beta\log(n))^je^{-c_1n^\beta x_j^2}
\leq n^{c_4\log\log(n)}e^{-c_5n^\beta x}.
\end{align*}
Summing over $1\leq j\leq k_n$ and adding the term $j=0$ gives
\begin{align*}
c_3\sum_{j=0}^{k_n}(c_3n^\beta\log(n))^je^{-c_1n^\beta x_j^2}
\lesssim e^{-c_1n^\beta x^2}+\log\log(n)e^{-c_5n^\beta x}
\lesssim e^{-c_6n^\beta x^2}.
\end{align*}
For the residual term, $2^{k_n}=\log(n)$ gives $x^{2^{-k_n}}\geq e^{-\beta/2}$ and
hence $x_{k_n+1}\geq\lambda_0:=c_2^{-1}e^{-\beta/2}$. Moreover
$(k_n+1)An^\beta\log(n)=o(n)$, so every index appearing after $k_n+1$ steps lies
in $[n/2,n]$ for $n$ large. Hence by Lemma~\ref{lem:excursion} applied with
$\eta=\min\{\lambda_0,\eta_0\}$,
\begin{align*}
&(c_3n^\beta\log(n))^{k_n+1}\max_{n/2\leq k\leq n}
\PP\big(|Y_{i,k}(\tau)-Q_i(\tau)|>x_{k_n+1}\big)\\
&\qquad\lesssim n^{c_7\log\log(n)}e^{-cn^{\min\{2(1-\beta),\beta\}}}
\lesssim e^{-c'n^{\min\{2(1-\beta),\beta\}}}.
\end{align*}
Therefore, for $n^{-\beta/2}\lesssim x<1$,
\begin{align*}
\PP\big(|Y_{i,n}(\tau)-Q_i(\tau)|>x\big)
\lesssim e^{-cn^\beta x^2}+e^{-c'n^{\min\{2(1-\beta),\beta\}}}.
\end{align*}
If $x\leq n^{-\beta/2}$, then $e^{-cn^\beta x^2}\geq e^{-c}$ and the bound holds
trivially. If $x\geq 1$, then by Lemma~\ref{lem:excursion} applied with
$\eta=\min\{1,\eta_0\}$,
\begin{align*}
\PP\big(|Y_{i,n}(\tau)-Q_i(\tau)|>x\big)
\leq\PP\big(|Y_{i,n}(\tau)-Q_i(\tau)|>\min\{1,\eta_0\}\big)
\lesssim e^{-c'n^{\min\{2(1-\beta),\beta\}}}.
\end{align*}
Hence the first branch of \eqref{3terms} holds for all $x>0$. The second branch follows from Theorem~\ref{thm:singletail}, since the
polynomial factor can be absorbed into the exponential in the large-deviation
range, and the remaining range is covered by the first branch (and is
trivial when \(n^{\beta(1-\beta)}x=O(1)\)). This completes the proof.
\end{proof}

\subsection{Proof of Theorem~\ref{thm:tail}}

\begin{proof}
By \eqref{eq:decompyiktau}, we have
\begin{align*}
\gamma_k^{-1}(Y_{i,k+1}(\tau)-Y_{i,k}(\tau))
=\xi_{i,k+1}(\tau)-f_i(Q_i(\tau))(Y_{i,k}(\tau)-Q_i(\tau))+\rho_{i,k}(\tau).      
\end{align*}
Re-arranging and averaging above for $k$ from $1$ to $n$ leads to
\begin{align}
\label{eq:decompofbarwintauqitau}
f_i(Q_i(\tau))(\bar Y_{i,n}(\tau)-Q_i(\tau))
=-n^{-1}\sum_{k=1}^n \gamma_k^{-1}(Y_{i,k+1}(\tau)-Y_{i,k}(\tau))+\bar \xi_{i,n}(\tau)+\bar\rho_{i,n}(\tau),
\end{align}
where
\begin{equation*}
    \bar \xi_{i,n}=n^{-1}\sum_{k=1}^n\xi_{i,k+1} \mbox{ and } 
    \bar \rho_{i,n}=n^{-1}\sum_{k=1}^n \rho_{i,k}(\tau).
\end{equation*}
By the summation by parts formula on the first term of the right hand side of \eqref{eq:decompofbarwintauqitau}, we obtain
\begin{align*}
&f_i(Q_i(\tau))(\bar Y_{i,n}(\tau)-Q_i(\tau))\nonumber\\
= &-n^{-1}\gamma_{n}^{-1}(Y_{i,n+1}(\tau)-Y_{i,1}(\tau))+n^{-1}\sum_{k=1}^{n-1}(Y_{i,k+1}(\tau)-Y_{i,1}(\tau))(\gamma_{k+1}^{-1}-\gamma_k^{-1})\\
&+\bar\xi_{i,n}(\tau)+\bar\rho_{i,n}(\tau).
\end{align*}
Decompose $Y_{i,n+1}(\tau)-Y_{i,1}(\tau)$
into $(Y_{i,n+1}(\tau)-Q_i(\tau))-(Y_{i,1}(\tau)-Q_i(\tau)),$ then above equation can be rewritten into
\begin{align}
\label{eq:decompyinbartauqitau}
&f_i(Q_i(\tau))(\bar Y_{i,n}(\tau)-Q_i(\tau))\nonumber\\
=&-\Big[n^{-1}\gamma_{n}^{-1}(Y_{i,n+1}(\tau)-Q_i(\tau))-
n^{-1}\gamma_1^{-1}(Y_{i,1}(\tau)-Q_i(\tau))\Big]\nonumber\\
&+n^{-1}\sum_{k=1}^{n-1}(Y_{i,k+1}(\tau)-Q_i(\tau))(\gamma_{k+1}^{-1}-\gamma_k^{-1})+\bar\xi_{i,n}(\tau)+\bar\rho_{i,n}(\tau)\nonumber\\
=&-\I_1+\I_2+\I_3+\I_4.
\end{align}
Denote $\alpha=\min\{2(1-\beta), \beta\}$. 
For part $\I_1,$ $n^{-1}\gamma_1^{-1}|Y_{i,1}(\tau)-Q_i(\tau)|\leq (c_y+M)/(c_\gamma n).$
By Theorem~\ref{thm:tailsingleonerefine}, for $x\geq 2(c_y+M)/(c_\gamma n)$, we have
\begin{align*}
&\PP(|\I_1|> x)\\
\leq &\PP\big(n^{-1}\gamma_n^{-1}|Y_{i,n+1}(\tau)-Q_i(\tau)|>x/2\big)\\
\lesssim &
\min\Big\{\exp(-c_1 n^{2-\beta}x^2)
+\exp\big(-c_2n^\alpha\big),  
\exp\big(-c_3n^{1-\beta^2}x\big)\Big\},
\end{align*}
where $c_l,$ $l\geq 1,$ here and throughout this proof are independent of $i,\tau,n,p.$

For part $\I_2,$ take $H_2=\min\{n^{2-\beta}x^2, n^\alpha, (nx)^{\alpha/(\alpha+\beta^2)}\}$, let $a_k\propto H_2/(k^{1-\beta^2}nx)$ for $k\leq K_0$, and 
$a_k\propto \sqrt{H_2/(k^{2-\beta}n^2x^2)}$ for $k>K_0,$ with $K_0=H_2^{1/\alpha}$.
By Theorem~\ref{thm:tailsingleonerefine} we obtain
\begin{align*}
&\PP(|\I_2|>x)\\
\leq& \sum_{k=1}^{n-1}\PP\big(|Y_{i,k+1}(\tau)-Q_i(\tau)|> a_kk^{1-\beta}nx\big)\\
\lesssim& \sum_{k=1}^{n}
\min\left\{
\exp\left(-a_k^2 k^{2-\beta} n^2 x^2\right)
+
\exp\left(-k^\alpha\right),
\;
\exp\left(-a_k k^{1-\beta^2} n x\right)
\right\}\\
\lesssim &
n\min\Big\{\exp(-H_2), \exp(-n^{1-\beta^2}x)\Big\}.
\end{align*}

For part $\I_3$, $(\xi_{i,k})_k$ are martingale differences with respect to $\F_{i,k}$ and $|\xi_{i,k}|\leq 1.$ Thus by Azuma's concentration inequality, 
\begin{align*}
\PP(|\I_3|>x)\leq 2\exp\{-nx^2/2\}.    
\end{align*}

For $\I_4$ part, by \eqref{eq:rhoikbddtwoparts}, 
\begin{align}
\label{eq:rhokbdd}
\sum_{k=1}^n|\rho_{i,k}(\tau)|
\leq c_4n^{1-\beta}+c_f\sum_{k=1}^n (Y_{i,k}(\tau)-Q_i(\tau))^2.
\end{align} 
Let $H_4=\min\{ (nx)^{\alpha/(2\alpha+1-2\beta(1-\beta))}, n^\beta x\}$ and $K_0=H_4^{1/\alpha}.$ For $k\geq K_0$ let $a_k\propto H_4/(nxk^\beta)$ and for $k\leq K_0$ let $a_k\propto H_4^2/(nxk^{2\beta(1-\beta)})$, then by Theorem~\ref{thm:tailsingleonerefine}, for $x\geq  2c_4n^{-\beta},$
\begin{align*}
&\PP(|\I_4|>x)\\
\leq &\sum_{k=1}^n\PP\Big(c_f(Y_{i,k}(\tau)-Q_i(\tau))^2> na_kx/2\Big) \\
\lesssim & \sum_{k=1}^{n}
\min\left\{
\exp\left(-k^\beta n a_k x\right)
+
\exp\left(- k^\alpha\right),
\;
\exp\left(-k^{\beta(1-\beta)} (n a_k x)^{1/2}\right)
\right\}\\
\lesssim& n\min\left\{
\exp(-H_4), \exp\left(-n^{\beta(1-\beta)}x^{1/2}\right)
\right\}.
\end{align*}
Combining $\I_1$-$\I_4$, we have
\begin{align*}
&\PP\big(|\bar Y_{i,n}(\tau)-Q_i(\tau)|>x\big)\\  \lesssim &\exp(-nx^2/2)+ 
n\exp(-E_{12}(x))+n\exp(-E_4(x)),
\end{align*}
where 
\begin{align}
\label{eq:defe12e4}
E_{12}(x)=\max\Big\{H_2, n^{1-\beta^2}x \Big\}\quad\textrm{and}
\quad 
E_4(x)=\max\Big\{H_4, n^{\beta(1-\beta)}x^{1/2} \Big\}.    
\end{align}
Then we complete the proof of \eqref{eq:singletailbaryintau}. 
Let $\delta=c_L x/2$. Consider a $\delta$-net $\A_\delta=\{t_0,t_1,\ldots,t_K\}$ of the interval $[\tau_0,\tau_1]$ with $t_0=\tau_0,$ $t_j=\tau_0+j\delta$ for $1\leq j\leq K-1$, and $t_K=\tau_1$. 
Since $\min_{1\leq i\leq p}\inf_{\tau_0\leq \tau\leq \tau_1}f_i(Q_i(\tau))\geq c_L,$ we have 
\begin{align}
\label{eq:qidifftj}
|Q_i(t_j)-Q_i(t_{j+1})|\leq c_L^{-1}\delta=x/2.    
\end{align}
Note that both $\bar Y_{i,n}(\tau)$ and $Q_i(\tau)$ are monotonic increasing, by \eqref{eq:qidifftj} the deviation $|\bar Y_{i,n}(\tau)- Q_i(\tau)| $ can be bounded for any $\tau\in[t_j,t_{j+1}]$ as follows, 
\begin{align}
\label{eq:bdddifynbarqtau}
|\bar Y_{i,n}(\tau)- Q_i(\tau)| 
&\leq \max_{0\leq j\leq K}|\bar Y_{i,n}(t_j)- Q_i(t_j)|
+\max_{0\leq j\leq K-1}|Q_i(t_j)-Q_i(t_{j+1})|\nonumber\\
&\leq \max_{0\leq j\leq K}|\bar Y_{i,n}(t_j)- Q_i(t_j)|+x/2.
\end{align}
Then \eqref{eq:singletailbaryintau} and \eqref{eq:bdddifynbarqtau} lead to
\begin{align*}
&\PP\Big(\max_{1\leq i\leq p}\sup_{\tau_0\leq \tau\leq \tau_1}|\bar Y_{i,n}(\tau)- Q_i(\tau)|>x\Big)\\    
\leq &\PP\Big(\max_{1\leq i\leq p}\max_{0\leq j\leq K}|\bar Y_{i,n}(t_j)- Q_i(t_j)|>x/2 \Big)\\
\lesssim  & Kp\Big[\exp(-nx^2/2)+ 
n\exp(-E_{12}(x))+n\exp(-E_4(x))\Big].
\end{align*}
Result follows in view of $K=(\tau_1-\tau_0)/\delta=O(1+x^{-1}).$

\end{proof}

\section{Proofs for Results in Section~\ref{sec:bahadur}}\label{sec:proof_sec4}

\subsection{Proof of Theorem~\ref{thm:unifbahadur}}

\begin{proof} 
By \eqref{eq:gkytau}, for any $\tau_0\leq \tau<\tau'\leq \tau_1,$ we have
\begin{align*}
&Y_{i,k+1}(\tau')-Y_{i,k+1}(\tau)\\
\leq & Y_{i,k}(\tau')-Y_{i,k}(\tau)+
\gamma_k \big((\tau'-\tau)-g_{a\gamma_k}(Y_{i,k}(\tau')-X_{i,k+1})+g_{a\gamma_k}(Y_{i,k}(\tau)-X_{i,k+1}) \big)\\
\leq & Y_{i,k}(\tau')-Y_{i,k}(\tau)+
\gamma_k(\tau'-\tau), 
\end{align*}
where the last inequality is due to the monotonicity of $Y_{i,k}(\tau)$ and $g_{a\gamma_k}(\cdot)$. By induction
\begin{align}
\label{eq:bddyktau}
0\leq Y_{i,k}(\tau')-Y_{i,k}(\tau)
\leq \sum_{j=1}^{k-1} \gamma_j(\tau'-\tau)
\leq c_\gamma(1-\beta)^{-1} k^{1-\beta}(\tau'-\tau).
\end{align}
Since function $g_{a\gamma_k}(\cdot)$ is Lipschitz continuous with Lipschitz constant $(2a\gamma_k)^{-1}$, 
by \eqref{eq:bddyktau},
\begin{align}
\label{eq:xitautaupri}
|\xi_{i,k+1}(\tau)-\xi_{i,k+1}(\tau')|
&\leq |\tau-\tau'|+(2a\gamma_k)^{-1}|Y_{i,k}(\tau)-Y_{i,k}(\tau')|\nonumber\\
&\leq  \big[1+(2a)^{-1}(1-\beta)^{-1} k\big]|\tau'-\tau|.
\end{align}
By \eqref{eq:bddyktau},
for any $\delta>0$ and $|s-t|\leq \delta,$
\begin{align*}
&\big|(\bar Y_{i,n}(s)-Q_i(s))-(\bar Y_{i,n}(t)-Q_i(t))\big|
\leq c_\gamma (1-\beta)^{-1}n^{1-\beta}\delta+c_L^{-1}\delta\lesssim n^{1-\beta}\delta,
\end{align*}
where the constant in $\lesssim$ here and in the rest of this proof are some positive constants independent of $n,p,i,s,t.$ 
By \eqref{eq:xitautaupri},
\begin{align*}
&|\bar\xi_{i,n}(s)/f_i(Q_i(s))-\bar\xi_{i,n}(t)/f_i(Q_i(t))|\\
\leq&|\bar\xi_{i,n}(s)-\bar\xi_{i,n}(t)|/f_i(Q_i(s))
+|\bar\xi_{i,n}(t)||1/f_i(Q_i(s))-1/f_i(Q_i(t))|\\
\lesssim &n\delta.      
\end{align*}
Denote $$l_i(\tau)=\bar Y_{i,n}(\tau)-Q_i(\tau)-\bar\xi_{i,n}(\tau)/f_i(Q_i(\tau)).$$ Then for any $|s-t|\leq \delta,$ we have 
$|l_i(t)-l_i(s)|\leq c_1n\delta,$ some constant $c_1>0.$

Take $\delta=x/(2c_1n).$
Consider a $\delta$-net $\A_\delta=\{t_0,t_1,\ldots,t_K\}$ of the interval
$[\tau_0,\tau_1]$ with $t_0=\tau_0,$ $t_j=\tau_0+j\delta$ for $1\leq j\leq K-1$,
and $t_K=\tau_1$, so that
\begin{align}
\label{eq:netcard}
K+1\leq 2+(\tau_1-\tau_0)/\delta\lesssim 1+n/x .
\end{align}
Hence 
\begin{align}
\label{eq:discretemaxijbar}
\max_{1\leq i\leq p}\sup_{\tau_0\leq \tau \leq \tau_1} |l_i(\tau)|\leq 
\max_{1\leq i\leq p}\max_{0\leq j\leq K} |l_i(t_j)|+x/2.
\end{align}
By \eqref{eq:decompyinbartauqitau}, we have the decomposition 
\begin{align}
\label{eq:decomi1234}
f_i(Q_i(\tau))l_i(\tau)
=&-\Big[n^{-1}\gamma_{n}^{-1}(Y_{i,n+1}(\tau)-Q_i(\tau))-
n^{-1}\gamma_1^{-1}(Y_{i,1}(\tau)-Q_i(\tau))\Big]\nonumber\\
&+n^{-1}\sum_{k=1}^{n-1}(Y_{i,k+1}(\tau)-Q_i(\tau))(\gamma_{k+1}^{-1}-\gamma_k^{-1})+\bar\rho_{i,n}(\tau).
\end{align}
By the argument of $\I_1,\I_2$ and $\I_4$ in the proof of Theorem~\ref{thm:tail} and $f_i(Q_i(\tau))\geq c_L$ for any $\tau_0\leq \tau\leq \tau_1$,
\begin{align*}
\PP(|l_i(\tau)|>x)
\lesssim
n\exp(-E(x))+n\exp(-E'(x)).
\end{align*}
Since
$$\PP\Big(\max_{1\leq i\leq p}\max_{0\leq j \leq K}|l_i(t_j)|>x/2\Big)
\leq \sum_{i=1}^p\sum_{j=0}^K \PP\big(|l_i(t_j)|>x/2\big),$$
the result follows by combining the above with \eqref{eq:discretemaxijbar} and
\eqref{eq:netcard}, the constants $c_1,c_2$ absorbing the factor $2$ in
$E(x/2)$ and $E'(x/2)$.
\end{proof}

\subsection{Proof of Theorem~\ref{thm:main}}

\begin{proof}
Take $\delta=n^{-2}.$ 
Consider a $\delta$-net $\A_\delta=\{t_0,t_1,\ldots,t_K\}$ of the interval $[\tau_0,\tau_1]$ with $t_0=\tau_0,$ $t_j=\tau_0+j\delta$, for $1\leq j\leq K-1$ and $t_K=\tau_1$. Let
$$l_{i,n}(t)=n^{1/2}f_i(Q_i(t))(\bar Y_{i,n}(t)-Q_i(t)).$$
By \eqref{eq:bddyktau}, $|Y_{i,k}(s)-Y_{i,k}(t)|\leq c_\gamma(1-\beta)^{-1} k^{1-\beta}|s-t|.$ Hence
for $|s-t|\leq \delta,$
\begin{align*}
|l_{i,n}(s)-l_{i,n}(t)| \leq c_1 n^{3/2-\beta} \delta,
\end{align*}
where $c_1=c_fc_\gamma(1-\beta)^{-1}+c_Lc_f$.
Let $\Delta_0=c_1n^{3/2-\beta}\delta$. By the uniform Bahadur representation in Theorem~\ref{thm:unifbahadur}, for any $\Delta_1>0$, we consider the decomposition
\begin{align}
\label{eq:decomptoseparts}
&\PP\Big(\max_{1\leq i\leq p}\sup_{\tau_0 \leq \tau\leq \tau_1}|l_{i,n}(\tau)|< u \Big)\nonumber\\
\leq&\PP\Big(\max_{1\leq i\leq p}\max_{0 \leq j \leq K}|l_{i,n}(t_j)|< u +\Delta_0 \Big)\nonumber\\
\leq &\PP\Big(\max_{1\leq i\leq p}\max_{0\leq j\leq K}\big|l_{i,n}(t_j)-n^{1/2}\bar\xi_{i,n}(t_j)\big| >\Delta_1\Big)\nonumber\\
&+\PP\Big(\max_{1\leq i\leq p}\max_{0\leq j\leq K}n^{1/2}|\bar\xi_{i,n}(t_j)|< u+\Delta_0+\Delta_1\Big)=\I_1+\I_2.
\end{align}
Then by taking 
\begin{align*}
\Delta_1
=
d_1\left(
n^{1/2-\beta}L_n
+
n^{-(1-\beta)/2}L_n^{1/2}
+
n^{-1/2}L_n^s
\right),   
\end{align*}
where $L_n=\log(np)$, $s=(2\alpha+1-2\beta(1-\beta))/\alpha$ and $d_1>0$ is some constant sufficiently large,  
Theorem~\ref{thm:unifbahadur} leads to
\begin{align*}
\I_1\lesssim 1/(np).
\end{align*}
For $\I_2$ part,
let $\xi_{i,k+1}^\diamond(\tau)$ be a process close to $\xi_{i,k+1}(\tau)$ with form
\begin{align}
\label{eq:defxidiamond}
\xi_{i,k+1}^\diamond(\tau)
=\EE\big(\1_{\{X_{i,k+1}\leq Q_i(\tau)\}}\big)-\1_{\{X_{i,k+1}\leq Q_i(\tau)\}}
=\tau-\1_{\{X_{i,k+1}\leq Q_i(\tau)\}},
\end{align}
For any $\Delta_2, \Delta_3>0$, part $\I_2$ can be further decomposed into
\begin{align}
\label{eq:i2decompparts}
\I_2\leq &\PP\Big(\max_{1\leq i\leq p}\max_{0\leq j\leq K}n^{1/2}\big| \bar \xi_{i,n}(t_j)-\bar\xi_{i,n}^\diamond(t_j)\big|>\Delta_2\Big)\nonumber\\
+&\sup_{z\in\RR}\Big|\PP\Big(\max_{1\leq i\leq p}\max_{0\leq j\leq K}n^{1/2}|\bar \xi_{i,n}^\diamond(t_j)|\leq z\Big)
-\PP\Big(\max_{1\leq i\leq p}\max_{0\leq j\leq K} |\B_i(t_j)| \leq z\Big)\Big|\nonumber\\
+&\PP\Big( \Big|\max_{1\leq i\leq p}\sup_{\tau_0\leq \tau\leq \tau_1}|\B_i(\tau)|- \max_{1\leq i\leq p}\max_{0\leq j\leq K} |\B_i(t_j)| \Big|>\Delta_3 \Big)\nonumber\\
+&\PP\Big(\max_{1\leq i\leq p}\sup_{\tau_0\leq \tau\leq \tau_1}|\B_i(\tau)| \leq u+\Delta_0+\Delta_1+\Delta_2+\Delta_3\Big)\nonumber\\
=&\I_{21}+\I_{22}+\I_{23}+\I_{24}.
\end{align}
Take
\begin{align*}
\Delta_2
=
d_2\left(
n^{-1/2}L_n+\sqrt{\frac{M_nL_n}{n}}
\right),   
\end{align*}
where $M_n$ is defined in Lemma~\ref{lem:diffxitilxi} and \(d_2>0\) is large. Since \(K\lesssim n^2\), by Lemma~\ref{lem:diffxitilxi},
\begin{align}
\label{eq:delta2bdd}
\I_{21}
&\lesssim 1/(np).
\end{align}
Note that $\xi_{i,k+1}^\diamond(\tau)$ are i.i.d.\ for $k=1,2,\ldots,n.$ And for any $\tau_0\leq s\leq t\leq \tau_1$, $$\cov(\xi_{i,k+1}^{\diamond}(t),\xi_{i,k+1}^{\diamond}(s) )=s(1-t),$$
which is the same as the Brownian Bridge $\B_i(\tau).$ By Proposition 2.1 in \citet{chernozhukov2017central}, 
\begin{align}
\label{eq:GA}
\I_{22}\lesssim \log^{7/6}(npK)n^{-1/6}.
\end{align}
For $t\in[0,1]$, consider
\begin{align*}
W_i(t)=\delta^{-1/2}(\B_i(\delta t+t_{j-1})-\B_i(t_{j-1})).
\end{align*}
Then for $s\leq t$, $\cov(W_i(t),W_i(s))=s(1-\delta t)$, and
$\EE(W_i(t)-W_i(s))^2=(t-s)(1-\delta(t-s)).$
By Theorem 1.1 in \citet{10.1214/aop/1176991419}
\begin{align*}
\PP\Big(\sup_{0\leq t\leq 1}|W_i(t)|> x\Big)
\lesssim x^{-1}e^{-x^2/2}.
\end{align*}
Since $\sup_{0\leq t\leq 1}|W_i(t)|=\delta^{-1/2}\sup_{t_{j-1}\leq \tau\leq t_j}|\B_i(\tau)-\B_i(t_{j-1})|$ and 
\begin{align*}
\sup_{\tau_0\leq \tau\leq \tau_1}|\B_i(\tau)|- \max_{0\leq j\leq K} |\B_i(t_j)| 
\leq \max_{1\leq j\leq K} \sup_{t_{j-1}\leq \tau\leq t_{j}}| \B_i(\tau)-\B_i(t_{j-1}) |,
\end{align*}
for any $\Delta_3>0,$ we have
\begin{align}
\label{eq:delta3bdd}
\I_{23}\leq &\sum_{i=1}^p\sum_{j=1}^K\PP\Big(\sup_{t_{j-1}\leq \tau\leq t_{j}}| \B_i(\tau)-\B_i(t_{j-1}) |>\Delta_3\Big)\nonumber \\
=&\sum_{i=1}^p\sum_{j=1}^K\PP\Big(\sup_{0\leq t\leq 1}\delta^{1/2}|W_i(t)|>\Delta_3\Big)\nonumber\\
\leq &pK\delta^{1/2}\Delta_3^{-1}e^{-\Delta_3^2/(2\delta)}.
\end{align}
Take $\Delta_3=d_3n^{-1}L_n,$ some constant $d_3>0$ large enough, then $\I_{23}\lesssim 1/(np).$ 
Combining $\I_{21}-\I_{23}$ into \eqref{eq:i2decompparts} leads to
\begin{align*}
\I_2&\lesssim 
1/(np)
+\log^{7/6}(npK)n^{-1/6}
+\PP\Big(\max_{1\leq i\leq p}\sup_{\tau_0\leq \tau\leq \tau_1}|\B_i(\tau)|\leq u+\Delta_0+\Delta_1+\Delta_2+\Delta_3\Big).
\end{align*}
By Theorem 3, the anti-concentration inequality in \citet{chernozhukov2015comparison},
\begin{align*}
&\PP\Big(\max_{1\leq i\leq p}\sup_{\tau_0\leq \tau\leq \tau_1}|\B_i(\tau)|\leq u+\sum_{l=0}^3\Delta_l\Big)-\PP\Big(\max_{1\leq i\leq p}\sup_{\tau_0\leq \tau\leq \tau_1}|\B_i(\tau)|\leq u\Big)\\
\leq &\sum_{l=0}^3\Delta_l\EE\Big(\max_{1\leq i\leq p}\max_{0\leq j\leq K}|\B_i(t_j)|\Big)\\
\lesssim&
\Big(\sum_{\ell=0}^3\Delta_\ell\Big)
\log^{1/2}(np).
\end{align*}
Combining the bounds for $\I_{21}, \I_{22}$ and
$\I_{23}$ into \eqref{eq:i2decompparts}, we obtain
\begin{align*}
&\mathbb P\left(
\max_{1\le i\le p}\sup_{\tau_0\le \tau\le \tau_1}
|l_{i,n}(\tau)|<u
\right)
-
\mathbb P\left(
\max_{1\le i\le p}\sup_{\tau_0\le \tau\le \tau_1}
|\mathcal B_i(\tau)|<u
\right)\nonumber\\
\lesssim&
L_n^{7/6}n^{-1/6}
+
(\Delta_0+\Delta_1+\Delta_2+\Delta_3)L_n^{1/2}\nonumber\\
\lesssim & 
L_n^{7/6}n^{-1/6}
+
n^{1/2-\beta}L_n^{3/2}
+
n^{-(1-\beta)/2}L_n\\
&+
n^{-1/2}L_n^{s+1/2}+
\min\left\{
n^{-\beta/4}L_n^{5/4},
\,
n^{-\beta(1-\beta)/2}L_n^{3/2}
\right\}.
\end{align*}
Repeating the preceding argument for the other direction yields the corresponding bound and hence establishes the claimed result.

\end{proof}

\subsection{Auxiliary unstudentized Gaussian approximation}

For the proofs of the bootstrap results, define
\begin{align}
\label{eq:defHprocess}
d_i(\tau)
=
f_i(Q_i(\tau)),
\qquad
\mathcal H_i(\tau)
=
\frac{\mathcal B_i(\tau)}{d_i(\tau)}.
\end{align}
The process $\mathcal H_i$ is used only as a theoretical device and does not
enter the implementation of the bootstrap.

\begin{lemma}
\label{lem:main_unstudentized_aux}
Under Assumption~\ref{asmp:condf},
\begin{align*}
\sup_{u\in\RR}
\Bigg|
&\PP\left(
\max_{1\le i\le p}
\sup_{\tau_0\le\tau\le\tau_1}
n^{1/2}
\left|\bar Y_{i,n}(\tau)-Q_i(\tau)\right|
\le u
\right)\\
&-
\PP\left(
\max_{1\le i\le p}
\sup_{\tau_0\le\tau\le\tau_1}
\left|\mathcal H_i(\tau)\right|
\le u
\right)
\Bigg|
\lesssim R_n.
\end{align*}
\end{lemma}

\begin{proof}
Assumption~\ref{asmp:condf} implies, uniformly in $i$ and
$\tau\in[\tau_0,\tau_1]$,
\begin{align}
\label{eq:dweights_regular}
c_L
\le d_i(\tau)
\le c_f,
\qquad
\left|d_i^{-1}(t)-d_i^{-1}(s)\right|
\lesssim |t-s|.
\end{align}
Indeed, the lower density bound gives
$|Q_i(t)-Q_i(s)|\le c_L^{-1}|t-s|$, while the assumed bound on $f_i'$
implies the second inequality.
Apply the proof of Theorem~\ref{thm:main} to the unstudentized process
\begin{align*}
\widetilde l_{i,n}(\tau)
=
n^{1/2}\left\{\bar Y_{i,n}(\tau)-Q_i(\tau)\right\}.
\end{align*}
Its leading variables are
\begin{align*}
\widetilde\xi_{i,k+1}^{\diamond}(\tau)
=
\frac{\xi_{i,k+1}^{\diamond}(\tau)}{d_i(\tau)}.
\end{align*}
Their covariance satisfies
\begin{align*}
&\operatorname{cov}\left(
\widetilde\xi_{i,k+1}^{\diamond}(t),
\widetilde\xi_{j,k+1}^{\diamond}(s)
\right)\\
&\qquad=
\frac{
\PP\{X_{i,k+1}\le Q_i(t),\,
      X_{j,k+1}\le Q_j(s)\}-ts
}{
d_i(t)d_j(s)
}
=
\operatorname{cov}\{\mathcal H_i(t),\mathcal H_j(s)\}.
\end{align*}
Moreover,
\begin{align*}
\frac{\tau_0(1-\tau_1)}{c_f^2}
\le
\operatorname{var}\{\mathcal H_i(\tau)\}
\le
\frac{1}{4c_L^2}.
\end{align*}
Thus, the variances remain uniformly bounded above and away from zero.
Although $\mathcal H_i$ is not itself a Brownian bridge,
\eqref{eq:dweights_regular} gives
\begin{align*}
|\mathcal H_i(t)-\mathcal H_i(s)|
\le
c_L^{-1}|\mathcal B_i(t)-\mathcal B_i(s)|
+
|\mathcal B_i(s)|
|d_i^{-1}(t)-d_i^{-1}(s)|.
\end{align*}
Moreover, by the standard maximal inequality for Brownian bridges and a union
bound, for a sufficiently large constant \(C>0\),
\[
\PP\left(
\max_{1\le i\le p}\sup_{\tau_0\le\tau\le\tau_1}
|\mathcal B_i(\tau)|>C L_n^{1/2}
\right)
\lesssim (np)^{-1}.
\]
Therefore, outside an event of probability \(O((np)^{-1})\), uniformly over
\(i\) and \(|t-s|\le n^{-2}\),
\[
|\mathcal B_i(s)|
\left|d_i^{-1}(t)-d_i^{-1}(s)\right|
\lesssim n^{-2}L_n^{1/2}.
\]
This term is dominated by the existing discretization error
\(\Delta_3\asymp n^{-1}L_n\). Consequently, replacing
\(\mathcal B_i\) by \(\mathcal H_i\) does not change the approximation rate.

The Bahadur remainder from Theorem~\ref{thm:unifbahadur} is divided by
$d_i(\tau)$ and therefore changes only by a constant factor. Consequently, all
remainder orders are unchanged, which gives the claimed bound.
\end{proof}

\subsection{Proof of Lemma~\ref{lem:diffxitilxi}}

\begin{lemma}
\label{lem:diffxitilxi}
Let $\xi_{i,k+1}^\diamond(\tau)$ be defined as in \eqref{eq:defxidiamond}. For any 
$z>0,$
\begin{align*}
&\max_{1\leq i\leq p}\sup_{\tau\in[\tau_0,\tau_1]}\PP\Big(\Big|\sum_{k=1}^n (\xi_{i,k+1}(\tau)-\xi_{i,k+1}^\diamond(\tau))\Big|> z\Big)\\
\lesssim &\exp(-cz)
+\exp(-c'z^2/M_n)
+(np)^{-4},
\end{align*}
where the constant in $\lesssim$ and $c,c'>0$ are independent of $n, p$ and
\begin{align*}
M_n
=
\min\left\{
n^{1-\beta/2}L_n^{1/2}
+
L_n^{(\alpha+1-\kappa)/\alpha}, n^{1-\kappa}L_n
\right\},    
\end{align*}
where
\begin{align*}
\alpha=\min\{2(1-\beta),\beta\},
\quad
\kappa=\beta(1-\beta),
\quad
L_n=\log(np).   
\end{align*}
\end{lemma}
\begin{proof}
Note that
$\xi_{i,k+1}(\tau)- \xi_{i,k+1}^\diamond(\tau),\ 1\leq k\leq n$, are martingale differences with respect to 
$\F_{i,k}=\sigma(X_{i,k},X_{i,k-1},\ldots)$ and are bounded by $2$. By Freedman's inequality for any $M>0,$
\begin{align}
\label{eq:feedman}
\PP\Big(\Big|\sum_{k=1}^n (\xi_{i,k+1}(\tau)-\xi_{i,k+1}^\diamond(\tau))\Big|> z\Big)
\leq 2\exp\Big(-\frac{z^2}{2z+M}\Big)+\PP(V\geq M),
\end{align}
where
\begin{align}
\label{eq:exressionofV}
V&=\sum_{k=1}^n \EE\big[(\xi_{i,k+1}(\tau)- \xi_{i,k+1}^\diamond(\tau))^2|\F_{i,k}\big].
\end{align}
Denote
\begin{align*}
h_{i,k}(x)=  \EE\big(\1_{\{x-X_{i,k+1}\geq 0\}}\big) -\1_{\{x-X_{i,k+1}\geq 0\}}.   
\end{align*}
Then $\xi_{i,k+1}^\diamond(\tau)=h_{i,k}(Q_i(\tau))$. For any $x,y$, we have
\begin{align}
\label{eq:bddgkxk2}
&\EE\big[(h_{i,k}(x)-h_{i,k}(y))^2|\F_{i,k}\big]\nonumber\\
=&F_i(x)-F_i^2(x)+F_i(y)-F_i^2(y)
-2(F_i(x\wedge y)-F_i(x)F_i(y))\nonumber\\
=& [F_i(x)+F_i(y)-2F_i(x\wedge y)]-[F_i(x)-F_i(y)]^2\nonumber\\
\leq &|f_i|_\infty|x-y|.
\end{align}
Let $\tilde \xi_{i,k+1}(\tau)$ be $\xi_{i,k+1}^\diamond(\tau)$ with $Q_i(\tau)$ therein replaced by $Y_{i,k}(\tau)$, that is 
$\tilde\xi_{i,k+1}(\tau)=h_{i,k}(Y_{i,k}(\tau)).$ 
By \eqref{eq:bddgkxk2}, we obtain that
\begin{align*}
\EE\big[(\xi_{i,k+1}^\diamond(\tau)-\tilde \xi_{i,k+1}(\tau))^2|\F_{i,k}\big]\leq c_f|Q_i(\tau)-Y_{i,k}(\tau)|.    
\end{align*}
Since $|\tilde \xi_{i,k+1}(\tau)-  \xi_{i,k+1}(\tau)|\leq 
\1_{ |Y_{i,k}(\tau)-X_{i,k+1}|\leq a\gamma_k}+\EE(\1_{ |Y_{i,k}(\tau)-X_{i,k+1}|\leq a\gamma_k}|\F_{i,k}),$ we have
\begin{align*}
\EE\big[(\tilde \xi_{i,k+1}(\tau)-  \xi_{i,k+1}(\tau))^2|\F_{i,k}\big]
\leq 4|f_i|_\infty a\gamma_k.  
\end{align*}
Implementing above to \eqref{eq:exressionofV}, we obtain
\begin{align}
\label{eq:Vbdd}
V\lesssim \sum_{k=1}^n |Q_i(\tau)-Y_{i,k}(\tau)|
+\sum_{k=1}^n \gamma_k
\lesssim \sum_{k=1}^n |Q_i(\tau)-Y_{i,k}(\tau)|+n^{1-\beta},
\end{align}
where the constant in $\lesssim$ and throughout this proof is independent of $n,p,i,\tau.$
For any $M>0$, by Theorem~\ref{thm:tailsingleonerefine}, for any
\(a_k>0\) with \(\sum_{k=1}^n a_k\le 1\), we have
\begin{align}
\label{eq:bddvm}
&\mathbb P\left(
\sum_{k=1}^n |Q_i(\tau)-Y_{i,k}(\tau)|>M
\right)\nonumber\\
\leq &
\sum_{k=1}^n
\mathbb P\big(
|Q_i(\tau)-Y_{i,k}(\tau)|>a_kM
\big) \nonumber\\
\lesssim &
\sum_{k=1}^n
\min\left\{
\exp(-c k^\beta a_k^2M^2)+\exp(-c'k^\alpha),
\exp(-c'' k^\kappa a_kM)
\right\}.
\end{align}
Fix $H>0$ and set $K=H^{1/\alpha}$. For $k\le K$, use the second
branch in \eqref{eq:bddvm} and take
$a_k\propto H/(M k^\kappa).$
For \(k>K\), use the first branch in \eqref{eq:bddvm} and take
$a_k\propto H^{1/2}/(M k^{\beta/2}).$
Then we need
\begin{align*}
M\gtrsim H^{1+(1-\kappa)/\alpha}
+
n^{1-\beta/2}H^{1/2}.    
\end{align*}
Thus, taking \(H=C L_n\), we obtain
\[
\mathbb P\left(
\sum_{k=1}^n |Q_i(\tau)-Y_{i,k}(\tau)|>M
\right)
\lesssim (np)^{-4},
\]
whenever
$M\gtrsim
L_n^{(\alpha+1-\kappa)/\alpha}
+
n^{1-\beta/2}L_n^{1/2}.$
On the other hand, using only the second branch in \eqref{eq:bddvm} with
$a_k\propto k^{-\kappa}$, 
we reach the same bound whenever
$M\gtrsim n^{1-\kappa}L_n.$ Therefore 
by \eqref{eq:Vbdd} for some constant $c$ sufficiently large we have
$
\mathbb P(V\ge cM_n)\lesssim (np)^{-4}.  
$
By \eqref{eq:feedman} and above we have
\begin{align}
\mathbb P\Big(
\Big|
\sum_{k=1}^n
\left\{
\xi_{i,k+1}(\tau)-\xi^\diamond_{i,k+1}(\tau)
\right\}
\Big|>z
\Big) 
&\lesssim
\exp\Big(-\frac{c z^2}{2z+M_n}\Big)
+(np)^{-4}\nonumber \\
&\lesssim
\exp(-c_1z)
+
\exp\left(-c_2 z^2/M_n\right)
+(np)^{-4}.
\end{align}
Hence we complete the proof.
\end{proof}

\section{Proofs for Results in Section~\ref{sec:bootstrap}}\label{sec:proof_sec5}

\subsection{Auxiliary decomposition for the perturbed recursion}
The corresponding results for Theorems~\ref{thm:tailsingleonerefine} and \ref{thm:tail} also hold with $Y_{i,n}(\tau)$ replaced by $Y_{i,n}^*(\tau)$. To see this, 
let $\mathcal F^*_{i,k}
=
\sigma\{X_{i,k},X_{i,k-1},\ldots,W_k,W_{k-1},\ldots,W_1\}$
and $G_{i,k}(y)
=
\mathbb E[
g_{a\gamma_k}(y-X_{i,k+1})
].$
For the perturbed recursion, define
\begin{align}
\label{eq:xiik1star}
\xi^*_{i,k+1}(\tau)=
&\;
W_{k+1}
\left\{
\tau
-
g_{a\gamma_k}\big(Y^*_{i,k}(\tau)-X_{i,k+1}\big)
\right\} -
\left\{
\tau
-
G_{i,k}\big(Y^*_{i,k}(\tau)\big)
\right\},
\end{align}
and
\begin{align}
\label{eq:rhoikstartau}
\rho^*_{i,k}(\tau)
=
f_i(Q_i(\tau))
\big(Y^*_{i,k}(\tau)-Q_i(\tau)\big)
+
\tau
-
G_{i,k}\big(Y^*_{i,k}(\tau)\big).    
\end{align}
Since \(W_{k+1}\) is independent of the past and satisfies
\(\mathbb E W_{k+1}=1\), \(\xi^*_{i,k+1}(\tau)\) is a martingale difference.
Moreover, from the bootstrap recursion,
\[
\gamma_k^{-1}
\big(Y^*_{i,k+1}(\tau)-Y^*_{i,k}(\tau)\big)
=
W_{k+1}
\left\{
\tau
-
g_{a\gamma_k}\big(Y^*_{i,k}(\tau)-X_{i,k+1}\big)
\right\}.
\]
Therefore,
\[
f_i(Q_i(\tau))
\big(Y^*_{i,k}(\tau)-Q_i(\tau)\big)
=
-\gamma_k^{-1}
\big(Y^*_{i,k+1}(\tau)-Y^*_{i,k}(\tau)\big)
+
\xi^*_{i,k+1}(\tau)
+
\rho^*_{i,k}(\tau).
\]
This is the same decomposition as in the proof for the original recursion, with
\(Y_{i,k}\), \(\xi_{i,k+1}\), and \(\rho_{i,k}\) replaced by
\(Y^*_{i,k}\), \(\xi^*_{i,k+1}\), and \(\rho^*_{i,k}\), respectively. Since
\(0\le W_{k+1}\le 2a\), all boundedness and Taylor-expansion arguments remain
valid up to changes in constants.

\subsection{Proposition~\ref{eq:prop_bootstrap} (Bootstrap version of Theorem~\ref{thm:singletail})}

The following proposition shows that the perturbed estimate $Y_{i,n}^*(\tau)$ satisfies the same sub-exponential tail bound as its non-perturbed counterpart in Theorem~\ref{thm:singletail}.

\begin{proposition}
\label{eq:prop_bootstrap}
Under Assumption~\ref{asmp:condf}, for any $0\leq t\leq cn^{(1-\beta)\beta}$ and
$\tau_0\leq \tau\leq \tau_1,$ we have
\begin{align}
\EE(e^{t|Y_{i,n}^*(\tau)-Q_i(\tau)|})\leq c'n^\beta,    
\end{align}
where $c,c'>0$ are some constants independent of $n,p.$ 
Hence by Markov's inequality, 
for any $x>0,$
\begin{align}
\mathbb P\left(
|Y^*_{i,n}(\tau)-Q_i(\tau)|>x
\right)
\le
c'n^\beta
\exp\left\{-c n^{\beta(1-\beta)}x\right\}.    
\end{align}
\end{proposition}
The proof follows the same moment-generating-function argument as in the proof
of Theorem~\ref{thm:singletail}. 
Since \(W_{k+1}\) is independent of \(X_{i,k+1}\), satisfies
\(\mathbb E W_{k+1}=1\), and is bounded by \(2a\), the Taylor expansion used in
the proof of Theorem~\ref{thm:singletail} remains valid with modified constants. 

\subsection{Proof of Lemma~\ref{lem:monotone_bootstrap}}

\begin{proof}
Since $W_{k+1}\leq 2a$ and $g_{a\gamma_k}$ is Lipschitz continuous with Lipschitz constant $(2a\gamma_k)^{-1},$ for any $\tau, \tau',$ we have
\begin{align*}
\big|W_{k+1}\gamma_k\big(g_{a\gamma_k}(Y_{i,k}^*(\tau)-X_{i,k+1})-g_{a\gamma_k}(Y_{i,k}^*(\tau')-X_{i,k+1}) \big)\big|
\leq  |Y_{i,k}^*(\tau)-Y_{i,k}^*(\tau')|.
\end{align*}
Hence if $Y_{i,k}^*(\tau)\geq Y_{i,k}^*(\tau')$, then
\begin{align}
\label{eq:gkytau_bootstrap}
&Y_{i,k+1}^*(\tau)-Y_{i,k+1}^*(\tau')\nonumber\\
=&Y_{i,k}^*(\tau)-Y_{i,k}^*(\tau')+\gamma_kW_{k+1}\Big(\tau-\tau'-g_{a\gamma_k}(Y_{i,k}^*(\tau)-X_{i,k+1})+g_{a\gamma_k}(Y_{i,k}^*(\tau')-X_{i,k+1}) \Big)\nonumber\\
\geq & \gamma_kW_{k+1}(\tau-\tau').
\end{align}
Similar to the argument in the proof of Lemma~\ref{lem:monotone}, the desired result follows by induction.
\end{proof}

\subsection{Proof of Proposition~\ref{prop:tailsingleonerefine_bootstrap}}
\begin{proof}
The second branch of the minimum in \eqref{eq:3terms_bootstrap} follows from Proposition~\ref{eq:prop_bootstrap}. It therefore suffices to prove the first branch.

Write the recursion \eqref{eq:bootstrap} as
\begin{align*}
Y_{i,k+1}^*(\tau)=Y_{i,k}^*(\tau)+\gamma_k Z_{i,k+1}^*(\tau),
\quad\textrm{where}\ 
Z_{i,k+1}^*(\tau)=W_{k+1}\big(\tau-g_{a\gamma_k}(Y_{i,k}^*(\tau)-X_{i,k+1})\big).
\end{align*}
Recall $\F_{i,k}^*=\sigma\{X_{i,k},X_{i,k-1},\ldots,W_k,W_{k-1},\ldots,W_1\}$ and $G_{i,k}(y)=\EE\big(g_{a\gamma_k}(y-X_{i,k+1})\big)$. Since $W_{k+1}$ is independent of $(X_{i,k+1},\F_{i,k}^*)$ and $\EE W_{k+1}=1$,
\begin{align}
\label{eq:driftstar}
\EE\big(Z_{i,k+1}^*(\tau)\,|\,\F_{i,k}^*\big)=\tau-G_{i,k}\big(Y_{i,k}^*(\tau)\big).
\end{align}
Thus the conditional drift of the perturbed recursion coincides with that of the original recursion \eqref{eq:sgdcont}. Recall $\xi_{i,k+1}^*(\tau)$ in \eqref{eq:xiik1star} and $\rho_{i,k}^*(\tau)$ in \eqref{eq:rhoikstartau}. 
By \eqref{eq:driftstar}, 
$$\xi_{i,k+1}^*(\tau)=Z_{i,k+1}^*(\tau)-\EE(Z_{i,k+1}^*(\tau)|\F_{i,k}^*).$$
So $\xi_{i,k+1}^*(\tau)$, $k\geq 1$, are martingale differences with respect to $\F_{i,k}^*$, bounded by $|\xi_{i,k+1}^*(\tau)|\leq  2a+1$, and the analogue of \eqref{eq:decompyiktau}--\eqref{eq:ykdecomposition1} holds:
\begin{align}
\label{eq:decompyiktau_bootstrap}
Y_{i,k+1}^*(\tau)-Q_i(\tau)
=\big[1-\gamma_kf_i(Q_i(\tau))\big](Y_{i,k}^*(\tau)-Q_i(\tau))+\gamma_k \big(\xi_{i,k+1}^*(\tau)+\rho_{i,k}^*(\tau)\big).
\end{align}
Since $\rho_{i,k}^*(\tau)$ is the same function of $Y_{i,k}^*(\tau)$ as $\rho_{i,k}(\tau)$ in \eqref{eq:phoikreprest} is of $Y_{i,k}(\tau)$, the derivation \eqref{eq:gktau2}--\eqref{eq:rhoikbddtwoparts} yields
\begin{align}
\label{eq:rhoikbddtwoparts_bootstrap}
|\rho_{i,k}^*(\tau)|\leq 2ac_f\gamma_k+c_f(Y_{i,k}^*(\tau)-Q_i(\tau))^2.
\end{align}

The proof of Theorem~\ref{thm:tailsingleonerefine} now applies with
$Y_{i,k}(\tau)$, $\xi_{i,k+1}(\tau)$, $\rho_{i,k}(\tau)$ and $\F_{i,k}$ replaced by
their starred counterparts. Indeed, the contraction factors
$a_{i,k}=1-\gamma_kf_i(Q_i(\tau))$, and hence $b_{i,k}$, are unchanged and
deterministic, so that $H_{i,k}^*(\tau)=\gamma_kb_{i,n-1}b_{i,k}^{-1}\xi_{i,k+1}^*(\tau)$
are martingale differences with respect to $\F_{i,k}^*$, bounded in absolute value
by $(2a+1)\gamma_kb_{i,n-1}b_{i,k}^{-1}$; this changes only the constants in
\eqref{eq:freedman1}--\eqref{eq:tij}. Since
$|Y_{i,k+1}^*(\tau)-Y_{i,k}^*(\tau)|\leq 2a\gamma_k$, the bound \eqref{eq:rhobddk}
holds with $Y_{i,k}(\tau)$ replaced by $Y_{i,k}^*(\tau)$ in view of
\eqref{eq:rhoikbddtwoparts_bootstrap}. Finally, the residual term after $k_n+1$
iterations is controlled by Lemma~\ref{lem:excursion}, which is stated for the
recursion \eqref{eq:bootstrap}.
\end{proof}
\subsection{Proof of Corollary~\ref{cor:tail_bootstrap}}
\begin{proof}
By the proof of Proposition~\ref{prop:tailsingleonerefine_bootstrap}, the identities \eqref{eq:decompofbarwintauqitau}--\eqref{eq:decompyinbartauqitau} hold with starred quantities, giving the decomposition into $\I_1$--$\I_4$. For part $\I_1$, since $Y_{i,1}^*(\tau)=0$, we have $n^{-1}\gamma_1^{-1}|Y_{i,1}^*(\tau)-Q_i(\tau)|\leq M/(c_\gamma n)$, and the tail bound follows from the same argument in Proposition~\ref{prop:tailsingleonerefine_bootstrap}. Parts $\I_2$ and $\I_4$ follow by applying Proposition~\ref{prop:tailsingleonerefine_bootstrap} at each index $k$ with the same choices of $a_k$, $H_2$ and $H_4$ as in the proof of Theorem~\ref{thm:tail}. For $\I_4$ one uses \eqref{eq:rhoikbddtwoparts_bootstrap} in place of \eqref{eq:rhoikbddtwoparts} in \eqref{eq:rhokbdd}. For part $\I_3$, since $\xi_{i,k+1}^*(\tau)$ are martingale differences with respect to $\F_{i,k}^*$ and $|\xi_{i,k+1}^*(\tau)|\leq 2a+1$, Azuma's inequality gives
\begin{align*}
\PP(|\I_3|>x)\leq 2\exp\{-nx^2/(2(2a+1)^2)\}.
\end{align*}
The uniform bound \eqref{eq:maxsupitaubardiff} follows from the $\delta$-net argument \eqref{eq:qidifftj}--\eqref{eq:bdddifynbarqtau}, using the monotonicity of $Q_i(\tau)$ and of $\bar Y_{i,n}^*(\tau)$.
\end{proof}

\subsection{Proposition~\ref{prop:unifbahadur_bootstrap} (Bootstrap Bahadur representation)}

The next proposition extends the uniform Bahadur representation of Theorem~\ref{thm:unifbahadur} to the perturbed recursion.
 
\begin{proposition}
\label{prop:unifbahadur_bootstrap}
Under Assumption~\ref{asmp:condf}, let $\bar\xi_{i,n}^*(\tau)=n^{-1}\sum_{k=1}^n \xi_{i,k+1}^*(\tau)$ with $\xi_{i,k+1}^*(\tau)$ as in \eqref{eq:xiik1star}. Then for any $x>0$,
\begin{align}
\label{eq:tailforunifbahadur_bootstrap}
&\PP\Big(\max_{1\leq i\leq p}\sup_{\tau_0\leq \tau\leq \tau_1} \big|\bar Y_{i,n}^*(\tau)-Q_i(\tau)-\bar\xi_{i,n}^*(\tau)/f_i(Q_i(\tau))\big| >x\Big)\nonumber\\
\lesssim& np(1+n/x)\Big(
\exp(-c_1E(x))+\exp(-c_2E'(x))\Big),
\end{align}
where $E(x)$ and $E'(x)$ are the same as in \eqref{eq:singletailbaryintau}, and $c_1,c_2$ and the constant in $\lesssim$ are independent of $n,p$. Consequently, letting $L_n=\log(np)$, $s=(2\alpha+1-2\beta(1-\beta))/\alpha$ with $\alpha=\min\{2(1-\beta),\beta\}$, and
\begin{align}
\label{eq:Delta1_bootstrap}
\Delta_1
=
d_1\left(
n^{1/2-\beta}L_n
+
n^{-(1-\beta)/2}L_n^{1/2}
+
n^{-1/2}L_n^s
\right),
\end{align}
for any $d>0$ there exists $d_1>0$ sufficiently large such that
\begin{align*}
\PP\Big(\max_{1\leq i\leq p}\sup_{\tau_0\leq \tau\leq \tau_1}
n^{1/2}\big|f_i(Q_i(\tau))\{\bar Y_{i,n}^*(\tau)-Q_i(\tau)\}-\bar\xi_{i,n}^*(\tau)\big| >\Delta_1\Big)
\lesssim (np)^{-d}.
\end{align*}
\end{proposition}
 
\begin{proof}
We follow the proof of Theorem~\ref{thm:unifbahadur}, replacing each ingredient by its starred analogue.
By \eqref{eq:gkytau_bootstrap}, for any $\tau_0\leq \tau<\tau'\leq \tau_1$,
\begin{align*}
Y_{i,k+1}^*(\tau')-Y_{i,k+1}^*(\tau)
\leq & Y_{i,k}^*(\tau')-Y_{i,k}^*(\tau)+
\gamma_k W_{k+1}\big((\tau'-\tau)-g_{a\gamma_k}(Y_{i,k}^*(\tau')-X_{i,k+1})\\
&+g_{a\gamma_k}(Y_{i,k}^*(\tau)-X_{i,k+1}) \big)\\
\leq & Y_{i,k}^*(\tau')-Y_{i,k}^*(\tau)+
2a\gamma_k(\tau'-\tau),
\end{align*}
where the last inequality uses the monotonicity of $Y_{i,k}^*(\tau)$ (Lemma~\ref{lem:monotone_bootstrap}), the monotonicity of $g_{a\gamma_k}(\cdot)$, and $0\leq W_{k+1}\leq 2a$. By induction,
\begin{align}
\label{eq:bddyktau_bootstrap}
0\leq Y_{i,k}^*(\tau')-Y_{i,k}^*(\tau)
\leq 2a\sum_{j=1}^{k-1} \gamma_j(\tau'-\tau)
\leq 2ac_\gamma(1-\beta)^{-1} k^{1-\beta}(\tau'-\tau),
\end{align}
which is the analogue of \eqref{eq:bddyktau} with constant inflated by $2a$. Next, since $g_{a\gamma_k}(\cdot)$ is Lipschitz continuous with Lipschitz constant $(2a\gamma_k)^{-1}$, $0\leq W_{k+1}\leq 2a$, and $|G_{i,k}(x)-G_{i,k}(y)|\leq |f_i|_\infty|x-y|$, by \eqref{eq:xiik1star} and \eqref{eq:bddyktau_bootstrap},
\begin{align}
\label{eq:xitautaupri_bootstrap}
|\xi_{i,k+1}^*(\tau)-\xi_{i,k+1}^*(\tau')|
&\leq W_{k+1}\big[|\tau-\tau'|+(2a\gamma_k)^{-1}|Y_{i,k}^*(\tau)-Y_{i,k}^*(\tau')|\big]\nonumber\\
&\quad+\big[|\tau-\tau'|+c_f|Y_{i,k}^*(\tau)-Y_{i,k}^*(\tau')|\big]\nonumber\\
&\leq c_1(1+k)|\tau'-\tau|,
\end{align}
for some constant $c_1>0$ depending only on $a,c_\gamma,\beta,c_f$.
Denote
\begin{align*}
l_i^*(\tau)=\bar Y_{i,n}^*(\tau)-Q_i(\tau)-\bar\xi_{i,n}^*(\tau)/f_i(Q_i(\tau)).
\end{align*}
By \eqref{eq:bddyktau_bootstrap}, for any $\delta>0$ and $|s-t|\leq \delta$,
\begin{align*}
\big|(\bar Y_{i,n}^*(s)-Q_i(s))-(\bar Y_{i,n}^*(t)-Q_i(t))\big|
\leq 2ac_\gamma (1-\beta)^{-1}n^{1-\beta}\delta+c_L^{-1}\delta\lesssim n^{1-\beta}\delta,
\end{align*}
and by \eqref{eq:xitautaupri_bootstrap}, together with $|\bar\xi_{i,n}^*(\tau)|\leq 2a+1$ and the Lipschitz continuity of $1/f_i(Q_i(\cdot))$ under Assumption~\ref{asmp:condf},
\begin{align*}
|\bar\xi_{i,n}^*(s)/f_i(Q_i(s))-\bar\xi_{i,n}^*(t)/f_i(Q_i(t))|
\lesssim n\delta.
\end{align*}
Hence $|l_i^*(t)-l_i^*(s)|\leq c_2n\delta$ for $|s-t|\leq\delta$ and some constant $c_2>0$. Take $\delta=x/(2c_2n)$ and consider the $\delta$-net $\A_\delta=\{t_0,t_1,\ldots,t_K\}$ of the interval $[\tau_0,\tau_1]$ with $t_0=\tau_0$, $t_j=\tau_0+j\delta$ for $1\leq j\leq K-1$, and $t_K=\tau_1$, so that $K=O(1+n/x)$. Then
\begin{align}
\label{eq:discretemaxijbar_bootstrap}
\max_{1\leq i\leq p}\sup_{\tau_0\leq \tau \leq \tau_1} |l_i^*(\tau)|\leq
\max_{1\leq i\leq p}\max_{0\leq j\leq K} |l_i^*(t_j)|+x/2.
\end{align}
By the proof of Proposition~\ref{prop:tailsingleonerefine_bootstrap}, the decomposition \eqref{eq:decomi1234} holds with starred quantities:
\begin{align}
\label{eq:decomi1234_bootstrap}
f_i(Q_i(\tau))l_i^*(\tau)
=&-\Big[n^{-1}\gamma_{n}^{-1}(Y_{i,n+1}^*(\tau)-Q_i(\tau))-
n^{-1}\gamma_1^{-1}(Y_{i,1}^*(\tau)-Q_i(\tau))\Big]\nonumber\\
&+n^{-1}\sum_{k=1}^{n-1}(Y_{i,k+1}^*(\tau)-Q_i(\tau))(\gamma_{k+1}^{-1}-\gamma_k^{-1})+\bar\rho_{i,n}^*(\tau),
\end{align}
where $\bar\rho_{i,n}^*(\tau)=n^{-1}\sum_{k=1}^n \rho_{i,k}^*(\tau)$ with $\rho_{i,k}^*(\tau)$ in \eqref{eq:rhoikstartau}. By the argument of $\I_1,\I_2$ and $\I_4$ in the proof of Corollary~\ref{cor:tail_bootstrap} and $f_i(Q_i(\tau))\geq c_L$ for any $\tau_0\leq \tau\leq \tau_1$,
\begin{align*}
\PP(|l_i^*(\tau)|>x)
\lesssim
n\exp(-c_3E(x))+n\exp(-c_4E'(x)).
\end{align*}
Since
\begin{align*}
\PP\Big(\max_{1\leq i\leq p}\max_{0\leq j \leq K}|l_i^*(t_j)|>x/2\Big)
\leq \sum_{i=1}^p\sum_{j=0}^K \PP\big(|l_i^*(t_j)|>x/2\big),
\end{align*}
the bound \eqref{eq:tailforunifbahadur_bootstrap} follows by combining the above
with \eqref{eq:discretemaxijbar_bootstrap} and $K=O(1+n/x)$, the constants
$c_3,c_4$ absorbing the factor $2$ in $E(x/2)$ and $E'(x/2)$. The tail bound with
$\Delta_1$ follows from \eqref{eq:tailforunifbahadur_bootstrap} by the same
computation that yields \eqref{J27706p} from \eqref{eq:tailforunifbahadur}.
\end{proof}

The following lemma is the bootstrap analogue of Lemma~\ref{lem:diffxitilxi} and will be used in the proof of Theorem~\ref{thm:bootstrap}.
 
\begin{lemma}
\label{lem:diffxitilxi_bootstrap}
Let $\xi_{i,k+1}^\diamond(\tau)$ be as in \eqref{eq:defxidiamond}, $\xi^{*\diamond}_{i,k+1}(\tau)=W_{k+1}\xi_{i,k+1}^\diamond(\tau)$, and
\begin{align*}
D_{i,k+1}(\tau)
=\xi_{i,k+1}(\tau)-\xi^\diamond_{i,k+1}(\tau),
\qquad
D^*_{i,k+1}(\tau)
=\xi^*_{i,k+1}(\tau)-\xi^{*\diamond}_{i,k+1}(\tau).
\end{align*}
Then for any $z>0$,
\begin{align}
\label{eq:diffDstarD}
\max_{1\leq i\leq p}\sup_{\tau\in[\tau_0,\tau_1]}
\PP\Big(\Big|\sum_{k=1}^n \big(D^*_{i,k+1}(\tau)-D_{i,k+1}(\tau)\big)\Big|> z\Big)
\lesssim \exp(-cz)
+\exp(-c'z^2/M_n)
+(np)^{-4},
\end{align}
where $M_n$ is as in Lemma~\ref{lem:diffxitilxi} and the constant in $\lesssim$ and $c,c'>0$ are independent of $n,p$.
\end{lemma}
 
\begin{proof}
Since $\EE(\xi^{*\diamond}_{i,k+1}(\tau)|\F_{i,k}^*)=\EE W_{k+1}\cdot \EE\xi_{i,k+1}^\diamond(\tau)=0$, the summands $D^*_{i,k+1}(\tau)-D_{i,k+1}(\tau)$, $1\leq k\leq n$, are martingale differences with respect to $\F_{i,k}^*$, bounded in absolute value by $4a+3$. By Freedman's inequality, as in \eqref{eq:feedman}, for any $M>0$,
\begin{align*}
\PP\Big(\Big|\sum_{k=1}^n \big(D^*_{i,k+1}(\tau)-D_{i,k+1}(\tau)\big)\Big|> z\Big)
\leq 2\exp\Big(-\frac{z^2}{c_5(z+M)}\Big)+\PP(V^*\geq M),
\end{align*}
where $c_5>0$ depends only on $a$ and
\begin{align*}
V^*=\sum_{k=1}^n \EE\big[(D^*_{i,k+1}(\tau)- D_{i,k+1}(\tau))^2|\F_{i,k}^*\big]
\leq 2\sum_{k=1}^n \EE\big[(D^*_{i,k+1}(\tau))^2|\F_{i,k}^*\big]
+2\sum_{k=1}^n \EE\big[(D_{i,k+1}(\tau))^2|\F_{i,k}^*\big].
\end{align*}
For the second sum, since $D_{i,k+1}(\tau)$ does not involve the multipliers and $X_{i,k+1}$ is independent of $\F_{i,k}^*$, we have $\EE[(D_{i,k+1}(\tau))^2|\F_{i,k}^*]=\EE[(D_{i,k+1}(\tau))^2|\F_{i,k}]$, which by the derivation of \eqref{eq:Vbdd} in the proof of Lemma~\ref{lem:diffxitilxi} is bounded by $c_6(|Y_{i,k}(\tau)-Q_i(\tau)|+\gamma_k)$. For the first sum, let
\begin{align*}
A_{i,k+1}(\tau)=\1_{\{X_{i,k+1}\le Q_i(\tau)\}}-g_{a\gamma_k}\big(Y^*_{i,k}(\tau)-X_{i,k+1}\big).
\end{align*}
Then, by \eqref{eq:xiik1star} and $\EE W_{k+1}=1$,
\begin{align*}
D^*_{i,k+1}(\tau)
=W_{k+1}A_{i,k+1}(\tau)-\big(\tau-G_{i,k}(Y^*_{i,k}(\tau))\big)
=W_{k+1}A_{i,k+1}(\tau)-\EE\big[W_{k+1}A_{i,k+1}(\tau)|\F_{i,k}^*\big],
\end{align*}
and hence, using $0\leq W_{k+1}\leq 2a$ and the argument of \eqref{eq:bddgkxk2},
\begin{align}
\label{eq:D2condbdd}
&\EE\big[(D^*_{i,k+1}(\tau))^2|\F_{i,k}^*\big]\nonumber\\
\leq& 4a^2\,\EE\big[A_{i,k+1}^2(\tau)|\F_{i,k}^*\big]\nonumber\\
\leq &8a^2\Big(\EE\big[\big(\1_{\{X_{i,k+1}\le Q_i(\tau)\}}-\1_{\{X_{i,k+1}\le Y^*_{i,k}(\tau)\}}\big)^2|\F_{i,k}^*\big]
+\EE\big[\1_{\{|Y^*_{i,k}(\tau)-X_{i,k+1}|\leq a\gamma_k\}}|\F_{i,k}^*\big]\Big)\nonumber\\
\leq &8a^2\big(c_f|Y^*_{i,k}(\tau)-Q_i(\tau)|+2ac_f\gamma_k\big).
\end{align}
Combining the two bounds,
\begin{align*}
V^*\lesssim \sum_{k=1}^n |Q_i(\tau)-Y_{i,k}(\tau)|+\sum_{k=1}^n |Q_i(\tau)-Y^*_{i,k}(\tau)|+n^{1-\beta}.
\end{align*}
The rest follows the same argument as in the proof of Lemma~\ref{lem:diffxitilxi} and we complete the proof.
\end{proof}

\subsection{Proof of Theorem~\ref{thm:bootstrap}}

\begin{proof}
Write $d_i(\tau)=f_i(Q_i(\tau))$. Take $\delta=n^{-2}$ and let
$\mathcal A_\delta=\{t_0,t_1,\ldots,t_K\}$ be the $\delta$-net used in the
proof of Theorem~\ref{thm:main}, where $K\lesssim n^2$. Define
\begin{align*}
S_{i,n}(\tau)
&=
n^{1/2}
\left\{
\bar Y^*_{i,n}(\tau)-\bar Y_{i,n}(\tau)
\right\},\\
Z_{i,n}(\tau)
&=
n^{-1/2}\sum_{k=1}^n
(W_{k+1}-1)
\frac{
\tau-\1_{\{X_{i,k+1}\le Q_i(\tau)\}}
}{
d_i(\tau)
}.
\end{align*}
By \eqref{eq:bddyktau} and \eqref{eq:bddyktau_bootstrap}, for
$|s-t|\le\delta$,
\begin{align*}
|S_{i,n}(s)-S_{i,n}(t)|
\le
c_1n^{3/2-\beta}\delta
=:\Delta_0.
\end{align*}
Consequently,
\begin{align}
\label{eq:Sdecomp_bootstrap}
\left|
\max_i\sup_{\tau_0\le\tau\le\tau_1}|S_{i,n}(\tau)|
-
\max_i\max_{0\le j\le K}|S_{i,n}(t_j)|
\right|
\le\Delta_0.
\end{align}

Recall $\bar\xi_{i,n}$, $\bar\xi^*_{i,n}$,
$\xi^\diamond_{i,k+1}$, $\xi^{*\diamond}_{i,k+1}$,
$D_{i,k+1}$, and $D^*_{i,k+1}$ from
Lemma~\ref{lem:diffxitilxi_bootstrap}, and define
\begin{align*}
R_{i,n}(\tau)
&=
d_i(\tau)
\left\{\bar Y_{i,n}(\tau)-Q_i(\tau)\right\}
-
\bar\xi_{i,n}(\tau),\\
R^*_{i,n}(\tau)
&=
d_i(\tau)
\left\{\bar Y^*_{i,n}(\tau)-Q_i(\tau)\right\}
-
\bar\xi^*_{i,n}(\tau).
\end{align*}
At every net point,
\begin{align}
\label{eq:sintjzintjdr}
S_{i,n}(t_j)
&=
Z_{i,n}(t_j)
+
\frac{n^{-1/2}}{d_i(t_j)}
\sum_{k=1}^n
\left\{
D^*_{i,k+1}(t_j)-D_{i,k+1}(t_j)
\right\}
\nonumber\\
&\quad+
\frac{n^{1/2}}{d_i(t_j)}
\left\{
R^*_{i,n}(t_j)-R_{i,n}(t_j)
\right\}.
\end{align}

By Theorem~\ref{thm:unifbahadur} and
Proposition~\ref{prop:unifbahadur_bootstrap},
\begin{align*}
\PP\left(
\max_i\sup_\tau
n^{1/2}
\left\{
|R_{i,n}(\tau)|+|R^*_{i,n}(\tau)|
\right\}
>
2\Delta_1
\right)
\lesssim (np)^{-1},
\end{align*}
where $\Delta_1$ is defined in \eqref{eq:Delta1_bootstrap}. Moreover,
Lemma~\ref{lem:diffxitilxi_bootstrap} and a union bound imply
\begin{align}
\label{eq:netD_bootstrap}
\PP\left(
\max_i\max_j
n^{-1/2}
\left|
\sum_{k=1}^n
\left\{
D^*_{i,k+1}(t_j)-D_{i,k+1}(t_j)
\right\}
\right|
>
\Delta_2
\right)
\lesssim (np)^{-1},
\end{align}
where
\begin{align*}
\Delta_2
=
d_2\left(
n^{-1/2}L_n+\sqrt{M_nL_n/n}
\right).
\end{align*}
Since $d_i(\tau)\ge c_L$, the total remainder in
\eqref{eq:sintjzintjdr} is bounded by
\begin{align}
\label{eq:scaled_bootstrap_remainder}
\widetilde\Delta
=
c_L^{-1}(2\Delta_1+\Delta_2)
\end{align}
outside an event of probability $O((np)^{-1})$.

The summand vectors defining $Z_{i,n}(t_j)$ are i.i.d.\ across $k$ and
uniformly bounded. Their covariance satisfies
\begin{align*}
&\operatorname{cov}\left(
(W_{k+1}-1)
\frac{t-\1_{\{X_{i,k+1}\le Q_i(t)\}}}{d_i(t)},
(W_{k+1}-1)
\frac{s-\1_{\{X_{j,k+1}\le Q_j(s)\}}}{d_j(s)}
\right)\\
&\qquad=
\frac{
\PP\{X_{i,k+1}\le Q_i(t),X_{j,k+1}\le Q_j(s)\}-ts
}{
d_i(t)d_j(s)
}
=
\operatorname{cov}\{\mathcal H_i(t),\mathcal H_j(s)\}.
\end{align*}
Their variances are uniformly bounded above and away from zero. Therefore,
the high-dimensional Gaussian approximation of
\citet{chernozhukov2017central} gives
\begin{align}
\label{eq:GA_bootstrap}
\sup_{u\in\RR}
\left|
\PP\left(
\max_i\max_j|Z_{i,n}(t_j)|\le u
\right)
-
\PP\left(
\max_i\max_j|\mathcal H_i(t_j)|\le u
\right)
\right|
\lesssim
L_n^{7/6}n^{-1/6}.
\end{align}

By \eqref{eq:dweights_regular} and the Brownian-bridge modulus argument used
in the proof of Theorem~\ref{thm:main}, for
$\Delta_3=d_3n^{-1}L_n$ and sufficiently large $d_3$,
\begin{align*}
\PP\left(
\max_i\sup_\tau|\mathcal H_i(\tau)|
-
\max_i\max_j|\mathcal H_i(t_j)|
>
\Delta_3
\right)
\lesssim (np)^{-1}.
\end{align*}
Combining the preceding bounds with Gaussian anti-concentration yields
\begin{align*}
\sup_{u\in\RR}
\Bigg|
&\PP\left(
\max_i\sup_\tau|S_{i,n}(\tau)|\le u
\right)
-
\PP\left(
\max_i\sup_\tau|\mathcal H_i(\tau)|\le u
\right)
\Bigg|
\lesssim R_n.
\end{align*}
Here, division by $d_i(\tau)$ changes only the constants because
$c_L\le d_i(\tau)\le c_f$. Indeed, using $r\le\beta-1/2$, $r\le(1-\beta)/2$, and
$r\le\beta/4$, direct calculation shows that all the Bahadur,
discretization, and Gaussian-approximation terms above are bounded by
$CR_n$.

Finally, Lemma~\ref{lem:main_unstudentized_aux} gives
\begin{align*}
\sup_{u\in\RR}
\left|
\PP(T\le u)
-
\PP\left(
\max_i\sup_\tau|\mathcal H_i(\tau)|\le u
\right)
\right|
\lesssim R_n.
\end{align*}
Since
\[
T^*=\max_i\sup_\tau|S_{i,n}(\tau)|,
\]
the triangle inequality gives
\begin{align*}
\sup_{u\in\RR}
\left|
\PP(T^*\le u)-\PP(T\le u)
\right|
\lesssim R_n,
\end{align*}
which proves the result.
\end{proof}

\subsection{Proof of Theorem~\ref{thm:bootstrap_cond}}

Let $\delta=n^{-2}$ and let
$\mathcal A_\delta=\{t_0,\ldots,t_K\}$, with $K\lesssim n^2$, be the
$\delta$-net used in the proof of Theorem~\ref{thm:bootstrap}. Recall that
$d_i(t)=f_i(Q_i(t))$, and define
\begin{align*}
x_k(i,j)
=
\frac{
t_j-\1_{\{X_{i,k+1}\le Q_i(t_j)\}}
}{
d_i(t_j)
},
\qquad
1\le i\le p,\quad
0\le j\le K,\quad
1\le k\le n.
\end{align*}
Then
\begin{align*}
Z_{i,n}(t_j)
=
n^{-1/2}\sum_{k=1}^n(W_{k+1}-1)x_k(i,j).
\end{align*}
The variables $x_k(i,j)$ are $\mathcal D_n$-measurable and satisfy
$|x_k(i,j)|\le c_L^{-1}$.
Define the empirical and population covariance matrices, indexed by the
pairs $(i,j)$, as
\begin{align}
\label{eq:Sigmahat}
\widehat\Sigma_n\big((i,j),(i',j')\big)
&=
n^{-1}\sum_{k=1}^n
x_k(i,j)x_k(i',j'),\\
\Sigma\big((i,j),(i',j')\big)
&=
\operatorname{cov}\left(
\mathcal H_i(t_j),\mathcal H_{i'}(t_{j'})
\right).
\nonumber
\end{align}
For every $k$,
\begin{align*}
\EE\left\{x_k(i,j)x_k(i',j')\right\}
=
\Sigma\big((i,j),(i',j')\big).
\end{align*}

\begin{lemma}
\label{lem:covconc_bootstrap}
Under Assumption~\ref{asmp:condf}, there exists a constant $C_0>0$ such that
\begin{align}
\label{eq:covconc}
\PP\left(
\max_{(i,j),(i',j')}
\left|
\widehat\Sigma_n\big((i,j),(i',j')\big)
-
\Sigma\big((i,j),(i',j')\big)
\right|
>
C_0\sqrt{\frac{L_n}{n}}
\right)
\le (np)^{-1}.
\end{align}
\end{lemma}

\begin{proof}
For each fixed pair $(i,j),(i',j')$, the centered products
\begin{align*}
x_k(i,j)x_k(i',j')
-
\EE\left\{x_k(i,j)x_k(i',j')\right\},
\qquad k=1,\ldots,n,
\end{align*}
are i.i.d.\ and uniformly bounded by a constant depending only on $c_L$.
Hoeffding's inequality and a union bound over at most $(pK)^2$ pairs yield
\eqref{eq:covconc}.
\end{proof}

\begin{proof}[Proof of Theorem~\ref{thm:bootstrap_cond}]
Recall $S_{i,n}$, $Z_{i,n}$, $R_{i,n}$, $R^*_{i,n}$,
$D_{i,k+1}$, and $D^*_{i,k+1}$ from the proof of
Theorem~\ref{thm:bootstrap}. Also, let
\begin{align*}
\Delta_0&=c_1n^{-1/2-\beta},\\
\Delta_2&=
d_2\left(n^{-1/2}L_n+\sqrt{M_nL_n/n}\right),\\
\Delta_3&=d_3n^{-1}L_n,
\end{align*}
and recall $\Delta_1$ from \eqref{eq:Delta1_bootstrap}.
Define the bootstrap exception event
\begin{align*}
A_n
={}&
\left\{
\max_{1\le i\le p}\max_{0\le j\le K}
n^{-1/2}
\left|
\sum_{k=1}^n
\left\{
D^*_{i,k+1}(t_j)-D_{i,k+1}(t_j)
\right\}
\right|
>\Delta_2
\right\}\\
&\quad\cup
\left\{
\max_{1\le i\le p}
\sup_{\tau_0\le\tau\le\tau_1}
n^{1/2}|R^*_{i,n}(\tau)|
>\Delta_1
\right\}.
\end{align*}
By \eqref{eq:netD_bootstrap} and
Proposition~\ref{prop:unifbahadur_bootstrap},
\begin{align*}
\PP(A_n)\le C_1(np)^{-1}.
\end{align*}
Since
\[
\EE\left\{\PP(A_n\mid\mathcal D_n)\right\}=\PP(A_n),
\]
Markov's inequality gives
\begin{align}
\label{eq:markov_cond}
\PP\left\{
\PP(A_n\mid\mathcal D_n)>
C_1^{1/2}(np)^{-1/2}
\right\}
\le
C_1^{1/2}(np)^{-1/2}.
\end{align}

Define the $\mathcal D_n$-measurable event
\[
\mathcal E_n
=
\mathcal E_{n,1}\cap
\mathcal E_{n,2}\cap
\mathcal E_{n,3},
\]
where
\begin{align*}
\mathcal E_{n,1}
&=
\left\{
\max_{(i,j),(i',j')}
\left|
\widehat\Sigma_n\big((i,j),(i',j')\big)
-
\Sigma\big((i,j),(i',j')\big)
\right|
\le C_0\sqrt{L_n/n}
\right\},\\
\mathcal E_{n,2}
&=
\left\{
\max_{1\le i\le p}
\sup_{\tau_0\le\tau\le\tau_1}
n^{1/2}|R_{i,n}(\tau)|
\le\Delta_1
\right\},\\
\mathcal E_{n,3}
&=
\left\{
\PP(A_n\mid\mathcal D_n)
\le C_1^{1/2}(np)^{-1/2}
\right\}.
\end{align*}
Lemma~\ref{lem:covconc_bootstrap},
Theorem~\ref{thm:unifbahadur}, and \eqref{eq:markov_cond} imply
\begin{align}
\label{eq:good_event_cond}
\PP(\mathcal E_n)
\ge
1-C(np)^{-1/2}.
\end{align}
All the following bounds are established on $\mathcal E_n$.
Conditionally on $\mathcal D_n$, the vectors
\begin{align*}
\left(
(W_{k+1}-1)x_k(i,j)
\right)_{1\le i\le p,\,0\le j\le K},
\qquad k=1,\ldots,n,
\end{align*}
are independent across $k$, centered, and uniformly bounded. Since
$\operatorname{var}(W_{k+1})=1$, their conditional covariance matrix is
$\widehat\Sigma_n$. It suffices to consider the regime in which $R_n\le c_0$ for a sufficiently
small constant $c_0>0$. Indeed, if $R_n>c_0$, the conclusion follows
trivially after enlarging $C$, since the Kolmogorov distance is bounded by
one. Moreover,
\begin{align*}
L_n^{7/6}n^{-1/6}\le R_n.
\end{align*}
Thus, in the nontrivial regime, $L_n^7/n$ is sufficiently small and hence
$L_n/n$ is sufficiently small. In particular, after choosing $c_0$ small
enough,
\begin{align*}
C_0\sqrt{\frac{L_n}{n}}
\le
\frac{\tau_0(1-\tau_1)}{2c_f^2}.
\end{align*}
The finitely many remaining values of $n$ can be absorbed by enlarging the
constant $C$.
Consequently, on $\mathcal E_{n,1}$, its diagonal entries satisfy
\begin{align*}
\widehat\Sigma_n\big((i,j),(i,j)\big)
&\ge
\frac{t_j(1-t_j)}{d_i^2(t_j)}
-
C_0\sqrt{L_n/n}\\
&\ge
\frac{\tau_0(1-\tau_1)}{2c_f^2}.
\end{align*}
Thus, the conditional variances are uniformly bounded away from zero.
Conditionally on $\mathcal D_n$, the Gaussian approximation of
\citet{chernozhukov2017central} therefore gives
\begin{align}
\label{eq:condCCK}
\sup_{u\in\RR}
\Bigg|
&\PP\left(
\max_{i,j}|Z_{i,n}(t_j)|\le u
\mathrel{\big|}\mathcal D_n
\right)
-
\PP\left(
\max_{i,j}|N_{i,j}|\le u
\right)
\Bigg|
\lesssim
L_n^{7/6}n^{-1/6},
\end{align}
where
\[
(N_{i,j})\sim N(0,\widehat\Sigma_n).
\]

By the Gaussian comparison inequality of
\citet{chernozhukov2015comparison}, on $\mathcal E_{n,1}$,
\begin{align}
\label{eq:gausscomp_cond}
\sup_{u\in\RR}
\Bigg|
&\PP\left(
\max_{i,j}|N_{i,j}|\le u
\right)
-
\PP\left(
\max_{i,j}|\mathcal H_i(t_j)|\le u
\right)
\Bigg|
\nonumber\\
&\lesssim
n^{-1/6}L_n^{5/6}.
\end{align}

By \eqref{eq:Sdecomp_bootstrap} and
\eqref{eq:sintjzintjdr}, on $A_n^c\cap\mathcal E_{n,2}$,
\begin{align*}
\left|
T^*
-
\max_{i,j}|Z_{i,n}(t_j)|
\right|
\le
\Delta_0+c_L^{-1}(2\Delta_1+\Delta_2).
\end{align*}
Let
\[
\Delta
=
\Delta_0+c_L^{-1}(2\Delta_1+\Delta_2).
\]
For every $u\in\RR$,
\begin{align*}
\PP(T^*\le u\mid\mathcal D_n)
&\le
\PP\left(
\max_{i,j}|Z_{i,n}(t_j)|
\le u+\Delta
\mathrel{\big|}\mathcal D_n
\right)
+
\PP(A_n\mid\mathcal D_n)\\
&\le
\PP\left(
\max_{i,j}|\mathcal H_i(t_j)|
\le u+\Delta
\right)
+
CL_n^{7/6}n^{-1/6}
+
C(np)^{-1/2}\\
&\le
\PP\left(
\max_i\sup_{\tau_0\le\tau\le\tau_1}
|\mathcal H_i(\tau)|
\le u+\Delta+\Delta_3
\right)\\
&\quad+
CL_n^{7/6}n^{-1/6}
+
C(np)^{-1/2}.
\end{align*}
The reverse inequality follows in the same way.

The Gaussian anti-concentration inequality of
\citet{chernozhukov2015comparison}, the uniform variance bounds for
$\mathcal H_i$, and
\[
\EE\left(
\max_i\sup_\tau|\mathcal H_i(\tau)|
\right)
\lesssim L_n^{1/2}
\]
show that the shift $\Delta+\Delta_3$ contributes at most
$C(\Delta+\Delta_3)L_n^{1/2}$. By the same rate calculation used in the proof of
Theorem~\ref{thm:bootstrap}, this contribution is bounded by $CR_n$.
Moreover,
$n^{-1/6}L_n^{5/6}\le n^{-1/6}L_n^{7/6}$, and $(np)^{-1/2}$ is of
smaller order than $R_n$. Therefore, on $\mathcal E_n$,
\begin{align}
\label{eq:conditional_H_approx}
\sup_{u\in\RR}
\left|
\PP(T^*\le u\mid\mathcal D_n)
-
\PP\left(
\max_i\sup_\tau|\mathcal H_i(\tau)|\le u
\right)
\right|
\le CR_n.
\end{align}

By Lemma~\ref{lem:main_unstudentized_aux},
\begin{align*}
\sup_{u\in\RR}
\left|
\PP(T\le u)
-
\PP\left(
\max_i\sup_\tau|\mathcal H_i(\tau)|\le u
\right)
\right|
\lesssim R_n.
\end{align*}
Combining this bound with \eqref{eq:conditional_H_approx} and using the
triangle inequality gives
\begin{align*}
\sup_{u\in\RR}
\left|
\PP(T^*\le u\mid\mathcal D_n)
-
\PP(T\le u)
\right|
\le CR_n
\end{align*}
on $\mathcal E_n$. The probability bound in
\eqref{eq:good_event_cond} proves the first assertion of the theorem, and
the $O_{\PP}(R_n)$ conclusion follows immediately.
\end{proof}

\subsection{Coverage based on the conditional bootstrap quantile}
To justify the coverage statement, let
\begin{align*}
F_n(u)
&=
\PP(T\le u),\\
F_n^*(u)
&=
\PP(T^*\le u\mid\mathcal D_n),
\end{align*}
and define the deterministic generalized inverse
\begin{align*}
q_n(a)
=
\inf\{u\in\RR:F_n(u)\ge a\}.
\end{align*}
On the event in Theorem~\ref{thm:bootstrap_cond}, write
$\varepsilon_n=CR_n$. The generalized-inverse inequalities give, for all
sufficiently large $n$,
\begin{align}
\label{eq:bootstrap_quantile_bracket}
q_n(1-\eta-\varepsilon_n)
\le
c^*_{1-\eta}
\le
q_n(1-\eta+\varepsilon_n).
\end{align}
Moreover, Lemma~\ref{lem:main_unstudentized_aux} and Gaussian
anti-concentration imply
\begin{align*}
\sup_{u\in\RR}
\left\{
F_n(u)-F_n(u-)
\right\}
\lesssim R_n.
\end{align*}
Using \eqref{eq:bootstrap_quantile_bracket} and the probability bound in
Theorem~\ref{thm:bootstrap_cond}, we obtain
\begin{align*}
\left|
\PP\left(T\le c^*_{1-\eta}\right)
-
(1-\eta)
\right|
\lesssim
R_n+(np)^{-1/2}.
\end{align*}

\section{Proofs for the local nonparametric quantile-regression results}

\label{app:local-sgd-details}
\subsection{Technical details for Section~\ref{sec:conditional}}

\subsubsection{Monotonicity of the local recursion}
\label{sec:mono_lnqr}
If $|z_i-Z_k|>h_n$, both $\widehat q_{i,k+1}(\tau)$ and
$\widehat q_{i,k+1}(\tau')$ equal their step-$k$ values.  If
$|z_i-Z_k|\le h_n$ and
$\tau\ge\tau'$, the Lipschitz property of $g_{a\gamma_{N_{i,k}}}$ gives
\[
\widehat q_{i,k+1}(\tau)-\widehat q_{i,k+1}(\tau')
\ge \gamma_{N_{i,k}}(\tau-\tau')
+\left(1-\frac{1}{2a}\right)
 \{\widehat q_{i,k}(\tau)-\widehat q_{i,k}(\tau')\}.
\]
Induction, starting from $\widehat q_{i,1}(\tau)=y_i$, proves that
$\tau\mapsto\widehat q_{i,k}(\tau)$ is nondecreasing.  The same calculation
also bounds its active increments by $\gamma_{N_{i,k}}$, as in
\eqref{eq:sgdcont}.

\subsubsection{Window regularity and local constant bias}

\begin{lemma}[Window regularity and local constant bias]
\label{lem:window-bias}
Under Assumption~\ref{asm:local-quantile}, there exist constants
$0<c_\pi<C_\pi<\infty$ and $0<c_d<C_d<\infty$, independent of
$n$, $p$, and $i$, such that, for all sufficiently large $n$,
\begin{align}
&c_\pi h_n
\le \pi_{i,h}\le C_\pi h_n,
\qquad 
\sup_i
\left\{
\|f_{i,h}^{\mathrm{win}}\|_\infty
+
\|\partial_y f_{i,h}^{\mathrm{win}}\|_\infty
\right\}
\le C_d,
\label{eq:window-regularity-1}\\
&\inf_i\inf_{\tau_0\le\tau\le\tau_1}
d_{i,h}(\tau)
\ge c_d,
\qquad
\sup_i\sup_{\tau_0\le\tau\le\tau_1}
|q_{i,h}(\tau)|
\le M+1.
\label{eq:window-regularity-2}
\end{align}
Moreover,
\begin{equation}
\sup_i\sup_{\tau_0\le\tau\le\tau_1}
\left\{
|q_{i,h}(\tau)-q_i(\tau)|
+
|d_{i,h}(\tau)-d_i(\tau)|
\right\}
\lesssim h_n^2.
\label{eq:window-bias-order}
\end{equation}
\end{lemma}

\begin{proof}
Suppress the subscript $n$ on $h_n$. Since
$c_Z\le f_Z(z)\le C_Z$ on the neighborhood containing all the windows,
\[
2c_Zh\le\pi_{i,h}\le2C_Zh.
\]
Differentiating the density under the integral gives
\[
\partial_y f_{i,h}^{\mathrm{win}}(y)
=
\frac{1}{\pi_{i,h}}
\int_{z_i-h}^{z_i+h}
\partial_yf(y\mid z)f_Z(z)\,\mathrm{d}z.
\]
The uniform bounds on $f(y\mid z)$ and $\partial_yf(y\mid z)$ therefore
imply
\[
\sup_i
\left\{
\|f_{i,h}^{\mathrm{win}}\|_\infty
+
\|\partial_yf_{i,h}^{\mathrm{win}}\|_\infty
\right\}
\le C_d
\]
for some constant $C_d$. This proves
\eqref{eq:window-regularity-1}.

For $H_{F,y}(z)=F(y\mid z)f_Z(z)$, a second-order Taylor expansion in
$z$ and the symmetry of the window give, uniformly in $i$ and for $y$ in
the fixed neighborhood specified in Assumption~\ref{asm:local-quantile},
\[
\int_{z_i-h}^{z_i+h}H_{F,y}(z)\,\mathrm{d}z
=
2hH_{F,y}(z_i)+O(h^3).
\]
Similarly,
\[
\pi_{i,h}=2hf_Z(z_i)+O(h^3).
\]

Let $\mathcal Y$ denote the fixed neighborhood of the conditional quantile
values specified in Assumption~\ref{asm:local-quantile}. Since $f_Z(z_i)$ is bounded away from zero, 
\begin{equation}
\sup_i\sup_{y\in\mathcal Y}
\left|
F_{i,h}^{\mathrm{win}}(y)-F(y\mid z_i)
\right|
\lesssim h^2.
\label{eq:window-cdf-order}
\end{equation}

Applying the same argument to
$H_{f,y}(z)=f(y\mid z)f_Z(z)$ yields
\begin{equation}
\sup_i\sup_{y\in\mathcal Y}
\left|
f_{i,h}^{\mathrm{win}}(y)-f(y\mid z_i)
\right|
\lesssim h^2.
\label{eq:window-density-order}
\end{equation}

Fix $\tau\in[\tau_0,\tau_1]$ and write
$q_i=q_i(\tau)$ and $q_{i,h}=q_{i,h}(\tau)$. The density lower bound and
the uniform bound on $\partial_yf(y\mid z)$ imply that there exists a fixed
neighborhood of $q_i$ on which
\[
f(y\mid z_i)\ge \frac{c_L}{2}
\]
uniformly in $i$ and $\tau$. Equation~\eqref{eq:window-cdf-order} then
ensures that $q_{i,h}$ lies in this neighborhood for all sufficiently large
$n$. Since
\[
F_{i,h}^{\mathrm{win}}(q_{i,h})
=
F(q_i\mid z_i)
=
\tau,
\]
the mean-value theorem and \eqref{eq:window-cdf-order} give
\[
|q_{i,h}-q_i|\lesssim h^2
\]
uniformly in $i$ and $\tau$. Since $|q_i|\le M$, this also gives
\[
\sup_i\sup_{\tau_0\le\tau\le\tau_1}
|q_{i,h}(\tau)|\le M+1
\]
for all sufficiently large $n$.

Finally, by \eqref{eq:window-density-order} and the uniform bound on
$\partial_yf(y\mid z)$,
\begin{align*}
|d_{i,h}(\tau)-d_i(\tau)|
&\le
\left|
f_{i,h}^{\mathrm{win}}\{q_{i,h}(\tau)\}
-
f\{q_{i,h}(\tau)\mid z_i\}
\right|\\
&\quad+
\left|
f\{q_{i,h}(\tau)\mid z_i\}
-
f\{q_i(\tau)\mid z_i\}
\right|\\
&\lesssim
h^2+|q_{i,h}(\tau)-q_i(\tau)|
\lesssim h^2.
\end{align*}
It also follows that
$d_{i,h}(\tau)\ge c_L/2$ for all sufficiently large $n$. This proves
\eqref{eq:window-regularity-2} and
\eqref{eq:window-bias-order}.
\end{proof}

\subsubsection{Conditioning argument and scope of the approximation}

Condition on the complete window-allocation sequence.  Within window $i$, the
active responses are then i.i.d.\ from $F_{i,h}^{\mathrm{win}}$, and the active
samples from distinct disjoint windows are conditionally independent.  A
uniform Chernoff bound gives
\[
 \PP\{c_0nh_n\le N_{i,n}\le C_0nh_n\text{ for every }i\}
 \ge1-Cp\exp(-cnh_n).
\]
On this event, Lemma~\ref{lem:heterogeneous-GA} applies with effective
sample size $nh_n$.  This yields
Theorem~\ref{thm:local-window-gaussian}.  Gaussian anti-concentration and
Lemma~\ref{lem:window-bias} then yield
Corollary~\ref{cor:local-point-gaussian}.

Disjointness is essential for the independent-bridge reference law.  With
overlapping windows, define
\[
 \psi_{i,t}(Z,Y)=\1_{\{Z\in A_{i,h}\}}
 \left[t-\1_{\{Y\le q_{i,h}(t)\}}\right].
\]
The first-order Gaussian covariance becomes
\[
 \operatorname{Cov}\{\B_i^{\mathrm{loc}}(t),\B_j^{\mathrm{loc}}(s)\}
 =\frac{\EE\{\psi_{i,t}(Z,Y)\psi_{j,s}(Z,Y)\}}
 {(\pi_{i,h}\pi_{j,h})^{1/2}},
\]
which is generally nonzero for $i\ne j$.  Independent-bridge critical values
are therefore valid only under disjointness.

Finally, the leading term in \eqref{eq:local-GA-rate} is retained for direct
comparison with Theorem~\ref{thm:main}, whose proof uses
\citet{chernozhukov2017central}.  Newer high-dimensional central limit
theorems may improve that term under their additional assumptions.  The SGD
remainder and the local smoothing bias are unchanged.

\subsubsection{Rate specialization}

When $\beta=2/3$, we have $r=1/6$ and $s=17/6$, so
\[
 \mathcal R_{n,h,p}
 =L_{n,h}^{7/6}(nh_n)^{-1/6}
 +L_{n,h}^{3/2}(nh_n)^{-1/6}
 +L_{n,h}^{10/3}(nh_n)^{-1/2}.
\]
A convenient sufficient condition is
\begin{equation}
 L_{n,h}=o\{(nh_n)^{1/9}\},
 \qquad nh_n^5\lambda_p\to0.
\label{eq:local-beta-two-thirds-condition}
\end{equation}
For fixed $p$ and $h_n\asymp n^{-\kappa}$, these conditions hold when
$1/5<\kappa<1$.  Hence $h_n\asymp n^{-1/4}$ is admissible.  On a fixed
one-dimensional compact interval, the disjointness condition also imposes
$p\lesssim h_n^{-1}$.

\subsection{A deterministic heterogeneous-sample-size extension}

We first isolate the only extension of Theorem~\ref{thm:main} needed for the random local
counts.

\begin{lemma}[Heterogeneous effective sample sizes]
\label{lem:heterogeneous-GA}
Let $n_1,\ldots,n_p$ be deterministic integers satisfying
\begin{align}
\label{eq:het-counts}    
 c_0m\le n_i\le C_0m,
        \qquad 1\le i\le p,
\end{align}
for constants $0<c_0<C_0<\infty$.  For every $i$, let
$V_{i,1},\ldots,V_{i,n_i}$ be i.i.d.\ from a distribution $H_i$ with density
$h_i$, and suppose the distributions satisfy Assumption~\ref{asmp:condf} uniformly in $i$.
Assume further that the samples are independent across $i$.  Run the smoothed
SGD recursion with learning rate $\gamma_r=c_\gamma r^{-\beta}$ on each
coordinate from step-$1$ initial values satisfying
$\max_{1\leq i\leq p}|\theta_{i,1}|\leq c_y<\infty$.  Let
\[
 \bar\theta_{i,n_i}(\tau)
 =n_i^{-1}\sum_{r=1}^{n_i}\theta_{i,r}(\tau)
\]
be the average estimator.  If
$q_i(\tau)=H_i^{-1}(\tau)$ and
$\delta_i(\tau)=h_i\{q_i(\tau)\}$, then
\begin{align}
\label{eq:het-GA}
 \sup_{u\in\RR}\left|
 \PP\left\{
  \max_i\sup_{\tau_0\le\tau\le\tau_1}
  n_i^{1/2}\delta_i(\tau)
  |\bar\theta_{i,n_i}(\tau)-q_i(\tau)|\le u
 \right\}
 -\PP(G_p^{\mathrm{loc}}\le u)
 \right|
 \lesssim \mathcal R_{m,p},
\end{align}
where
\[
 \mathcal R_{m,p}
 =\ell^{7/6}m^{-1/6}
  +\ell^{3/2}m^{-r}
  +\ell^{s+1/2}m^{-1/2},
 \qquad
 \ell=\log(epm).
\]
Here $r$ and $s$ are the quantities defined in Theorem~\ref{thm:main}.
\end{lemma}

\begin{proof}
We explain the changes relative to the proof of Theorem~\ref{thm:main}.
All tail bounds, the uniform Bahadur representation, the discretization
argument, and the Gaussian anti-concentration step used in the proof of
Theorem~\ref{thm:main} remain valid uniformly over $i$ after replacing $n$
by $m$, because \eqref{eq:het-counts} makes all $n_i$ comparable to $m$.
These arguments yield the second and third terms in $\mathcal R_{m,p}$.

It remains to verify that the leading empirical quantile processes admit the
same joint Gaussian approximation despite the unequal sample sizes. Let
$\mathcal T_m$ be a regular grid of $[\tau_0,\tau_1]$ with mesh $m^{-2}$.
Then $|\mathcal T_m|\lesssim m^2$.  Define
\begin{align}
 \label{eq:Ui-def} 
 U_i(t)=n_i^{-1/2}\sum_{r=1}^{n_i}
       \left[t-\1_{\{V_{i,r}\le q_i(t)\}}\right],
 \qquad t\in\mathcal T_m.
\end{align}
 
Let $n_+=\max_i n_i$ and, for $1\le r\le n_+$, set
\[
 Z_r(i,t)
 =\left(\frac{n_+}{n_i}\right)^{1/2}
   \1_{\{r\le n_i\}}
   \left[t-\1_{\{V_{i,r}\le q_i(t)\}}\right].
\]
The vectors $\{Z_r(i,t):1\le i\le p,t\in\mathcal T_m\}$ are independent over
$r$, centered, and uniformly bounded because $n_+/n_i\le C_0/c_0$.  Also,
\[
 U_i(t)=n_+^{-1/2}\sum_{r=1}^{n_+}Z_r(i,t).
\]
For $i,j\le p$ and $s,t\in\mathcal T_m$, independence across coordinates and
the probability integral transform give
\begin{align*}
 \frac1{n_+}\sum_{r=1}^{n_+}
 \EE\{Z_r(i,t)Z_r(j,s)\}
 =\1_{\{i=j\}}\{\min(t,s)-ts\}.
\end{align*}
Thus the comparison Gaussian vector consists of independent Brownian bridges
evaluated on $\mathcal T_m$.  Applying the high-dimensional central limit
theorem of \citet{chernozhukov2017central} to the
$2p|\mathcal T_m|$ coordinates obtained by adjoining their negatives gives
\[
 \sup_u\left|
 \PP\left\{ \max_i\max_{t\in\mathcal T_m}|U_i(t)|\le u\right\}
 -\PP\left\{ \max_i\max_{t\in\mathcal T_m}|\B_i^{\mathrm{loc}}(t)|\le u\right\}
 \right|
 \lesssim \ell^{7/6}m^{-1/6}.
\]
The replacement of the smoothed martingale differences by the indicator scores
in \eqref{eq:Ui-def}, the discretization of the SGD paths, and the Brownian-bridge modulus
of continuity are unchanged from the proof of Theorem~\ref{thm:main} because all $n_i$ are
comparable to $m$.  This gives the first term in $\mathcal R_{m,p}$ and completes the proof of
\eqref{eq:het-GA}.
\end{proof}

\subsection{Proof of Theorem~\ref{thm:local-window-gaussian}}

\begin{proof}
Let
\[
 J_k=
 \begin{cases}
 i,&Z_k\in A_{i,h}\text{ for some }1\le i\le p,\\
 0,&Z_k\notin\bigcup_{i=1}^pA_{i,h},
 \end{cases}
\]
which is well-defined by the disjointness condition, and let
$\mathcal A_n=\sigma(J_1,\ldots,J_n)$.  For each $i$, define the active times
\[
 T_{i,r}=\inf\{k:N_{i,k}=r\},
 \qquad 1\le r\le N_{i,n},
\]
and active observations $Y_{i,r}^{\circ}=Y_{T_{i,r}}$.  Set
$\theta_{i,1}(\tau)=y_i$.  
At the $r$th active observation,
$\widehat q_{i,T_{i,r}}(\tau)=\theta_{i,r}(\tau)$, and equation
\eqref{eq:condsgdN} becomes
\begin{equation}
\label{eq:active-recursion}
 \theta_{i,r+1}(\tau)
 =\theta_{i,r}(\tau)
  +\gamma_r\left[
       \tau-g_{a\gamma_r}
       \{\theta_{i,r}(\tau)-Y_{i,r}^{\circ}\}
    \right].
\end{equation}
Moreover,
\[
 \bar q_{i,n}(\tau)
 =N_{i,n}^{-1}\sum_{r=1}^{N_{i,n}}\theta_{i,r}(\tau).
\]

Conditionally on $\mathcal A_n$, the variables
$Y_{i,1}^{\circ},\ldots,Y_{i,N_{i,n}}^{\circ}$ are i.i.d.\ with distribution
$F_{i,h}^{\mathrm{win}}$.  The active samples belonging to different $i$ are conditionally
independent.  This follows directly from the independence of the pairs
$(Z_k,Y_k)$: conditioning on $J_k=i$ replaces the law of $Y_k$ by the law of
$Y$ given $Z\in A_{i,h}$, and conditioning on all $J_k$ preserves independence
across $k$.

By Lemma~\ref{lem:window-bias},
$c_\pi h_n\le\pi_{i,h}\le C_\pi h_n$.  A multiplicative Bernstein or Chernoff
bound gives
\[
 \PP\left\{
   |N_{i,n}-n\pi_{i,h}|>\frac12n\pi_{i,h}
 \right\}
 \le2\exp(-c n\pi_{i,h})
 \le2\exp(-c'nh_n).
\]
Hence, for suitable constants $0<c_0<C_0<\infty$, the event
\begin{align}
 \label{eq:count-event}   
  \mathcal E_n=
 \{c_0nh_n\le N_{i,n}\le C_0nh_n\text{ for every }1\le i\le p\}
\end{align}
satisfies
\begin{equation}
\label{eq:count-event-prob}
       \PP(\mathcal E_n^c)\le2p\exp(-cnh_n).
\end{equation}

Fix a realization of $\mathcal A_n$ for which $\mathcal E_n$ holds.  The
conditional active recursions \eqref{eq:active-recursion} satisfy the hypotheses of
Lemma~\ref{lem:heterogeneous-GA} with $m=nh_n$, in view of
\eqref{eq:window-regularity-1}--\eqref{eq:window-regularity-2}.  Therefore,
uniformly over every such realization,
\begin{align}
 \label{eq:cond-GA}   
 \sup_u\left|
 \PP\{T_n^{\mathrm{win}}\le u\mid\mathcal A_n\}-\PP(G_p^{\mathrm{loc}}\le u)
 \right|
 \le C\mathcal R_{n,h,p}.
\end{align}
Integrating \eqref{eq:cond-GA} over $\mathcal A_n$ and using \eqref{eq:count-event-prob} proves
\eqref{eq:window-GA}.
\end{proof}

\subsection{Proof of Corollary~\ref{cor:local-point-gaussian}}

\begin{proof}
It is useful to separate the centering error from the replacement of the
window density.  Define
\[
 T_n^{\mathrm{cen}}
 =\max_{1\le i\le p}\sup_{\tau_0\le\tau\le\tau_1}
   N_{i,n}^{1/2}d_{i,h}(\tau)
   |\bar q_{i,n}(\tau)-q_i(\tau)|.
\]
On the count event $\mathcal E_n$ in \eqref{eq:count-event}, Lemma~\ref{lem:window-bias}
gives
\begin{equation}
\label{eq:cen-win-diff}
 |T_n^{\mathrm{cen}}-T_n^{\mathrm{win}}|
 \le C\max_iN_{i,n}^{1/2}
       \sup_\tau|q_{i,h}(\tau)-q_i(\tau)|
 \le C(nh_n)^{1/2}h_n^2=:b_n.
\end{equation}

Since $\tau_0$ and $\tau_1$ are fixed interior quantile levels,
\[
0<
\inf_{\tau_0\leq\tau\leq\tau_1}\tau(1-\tau)
\leq
\sup_{\tau_0\leq\tau\leq\tau_1}\tau(1-\tau)
<\infty.
\]
Therefore, the Gaussian anti-concentration inequality for separable
Gaussian processes with marginal variances uniformly bounded away from
zero yields
\begin{equation}
\sup_{u\in\mathbb R}
\PP\left(
|G_p^{\mathrm{loc}}-u|\leq\varepsilon
\right)
\leq
C\varepsilon
\left\{
\mathbb E(G_p^{\mathrm{loc}})+1
\right\}
\leq
C\varepsilon\lambda_p^{1/2}.
\label{eq:Gp-anticon}
\end{equation}

Combining \eqref{eq:cen-win-diff}, Theorem~\ref{thm:local-window-gaussian}, \eqref{eq:count-event-prob}, and
\eqref{eq:Gp-anticon}, in both directions, gives
\begin{align}
\label{eq:Tcen-GA}
 \sup_u|\PP(T_n^{\mathrm{cen}}\le u)-\PP(G_p^{\mathrm{loc}}\le u)|
 &\lesssim \mathcal R_{n,h,p}+p\exp(-cnh_n)
             +b_n\lambda_p^{1/2}\notag\\
 &\lesssim \mathcal R_{n,h,p}+p\exp(-cnh_n)
             +\{nh_n^5\lambda_p\}^{1/2}.
\end{align}

We next replace $d_{i,h}$ with $d_i$.  On $\mathcal E_n$, the uniform density
expansion in Lemma~\ref{lem:window-bias} implies
\begin{equation}
 |T_n^{\mathrm{loc}}-T_n^{\mathrm{cen}}|
 \le Ch_n^2T_n^{\mathrm{cen}}.
\label{eq:loc-cen-diff}
\end{equation}
Let $v_n=C_0L_{n,h}^{1/2}$, where $C_0$ is sufficiently large.  By \eqref{eq:Tcen-GA}, the
sub-Gaussian tail of $G_p^{\mathrm{loc}}$, and $L_{n,h}=\log(epnh_n)\ge\log(ep)$,
\begin{equation}
\label{eq:Tcen-tail}
 \PP(T_n^{\mathrm{cen}}>v_n)
 \lesssim \mathcal R_{n,h,p}+p\exp(-cnh_n)
 +\{nh_n^5\lambda_p\}^{1/2}+(epnh_n)^{-c_0},
\end{equation}
where $c_0$ can be made arbitrarily large by increasing $C_0$.  On
$\mathcal E_n\cap\{T_n^{\mathrm{cen}}\le v_n\}$, \eqref{eq:loc-cen-diff} is bounded by
$Ch_n^2L_{n,h}^{1/2}$.  A second application of \eqref{eq:Gp-anticon}, together with
\eqref{eq:count-event-prob}, \eqref{eq:Tcen-GA}, and \eqref{eq:Tcen-tail}, therefore gives
\begin{align*}
 \sup_u|\PP(T_n^{\mathrm{loc}}\le u)-\PP(G_p^{\mathrm{loc}}\le u)|
 \lesssim{}&\mathcal R_{n,h,p}+p\exp(-cnh_n)
 +\{nh_n^5\lambda_p\}^{1/2}\\
 &+h_n^2(L_{n,h}\lambda_p)^{1/2}+(epnh_n)^{-c_0}.
\end{align*}
The final polynomial term is absorbed by
$L_{n,h}^{7/6}(nh_n)^{-1/6}$ after choosing $c_0$ large.  This proves
\eqref{eq:local-point-GA}.

It remains to verify the asserted sufficient conditions.  The condition
$\lambda_p/(nh_n)\to0$ gives $p\exp(-cnh_n)\to0$.  Also,
\[
 h_n^2(L_{n,h}\lambda_p)^{1/2}
 =\{nh_n^5\lambda_p\}^{1/2}
   \{L_{n,h}/(nh_n)\}^{1/2}.
\]
The first factor tends to zero by assumption.  The first term of
$\mathcal R_{n,h,p}$ tending to zero implies
$L_{n,h}^7/(nh_n)\to0$, hence $L_{n,h}/(nh_n)\to0$.  Thus every term in
\eqref{eq:local-point-GA} vanishes.
\end{proof}

\subsection{Proof of Corollary~\ref{cor:local-feasible-bands}}

\begin{proof}
Let
\[
 \widehat T_n
 =\max_i\sup_{\tau_0\le\tau\le\tau_1}
  N_{i,n}^{1/2}\widehat d_i(\tau)
  |\bar q_{i,n}(\tau)-q_i(\tau)|.
\]
On the event $\Delta_{d,n}\le\eta_n$,
\begin{equation}
       |\widehat T_n-T_n^{\mathrm{loc}}|
       \le\eta_n T_n^{\mathrm{loc}}.
\label{eq:That-diff}
\end{equation}
Corollary~\ref{cor:local-point-gaussian}, the bound
\(\mathbb E(G_p^{\mathrm{loc}})\lesssim\lambda_p^{1/2}\) used in
\eqref{eq:Gp-anticon}, and Gaussian concentration imply
\begin{align}
\label{eq:Tloc-order}  
T_n^{\mathrm{loc}}=O_{\PP}(\lambda_p^{1/2}).
\end{align}
Choose a deterministic $\eta_n\downarrow0$ such that
$\PP(\Delta_{d,n}>\eta_n)\to0$ and $\lambda_p\eta_n\to0$.  This is possible by
\eqref{eq:density-relative-rate}.  For any fixed $C>0$, on the event
$\{\Delta_{d,n}\le\eta_n,\ T_n^{\mathrm{loc}}\le C\lambda_p^{1/2}\}$,
the difference in \eqref{eq:That-diff} is at most $C\eta_n\lambda_p^{1/2}$.  By the
anti-concentration bound \eqref{eq:Gp-anticon}, the corresponding change in distribution is
at most $C'\eta_n\lambda_p=o(1)$.  The complement of this event can be made
arbitrarily small by first choosing $C$ large, using \eqref{eq:Tloc-order}, and then letting
$n\to\infty$.  It follows that
\[
 \sup_u|\PP(\widehat T_n\le u)-\PP(G_p^{\mathrm{loc}}\le u)|\to0.
\]
The asserted simultaneous coverage follows from the definition of
$c_{1-\alpha,p}^{\mathrm{loc}}$.  Continuity of the law of $G_p^{\mathrm{loc}}$ follows from \eqref{eq:Gp-anticon}.
\end{proof}

\section{Additional Simulation and Illustration Results}\label{sec:numerical_app}

\setcounter{table}{0}
\renewcommand{\thetable}{E\arabic{table}}
\setcounter{figure}{0}
\renewcommand{\thefigure}{E\arabic{figure}}

\subsection{Additional Simulation Results}

\begin{table}[H]
\footnotesize
\centering
\begin{tabular}{r|rrrr|rrrr}
	\hline 
  \multicolumn{1}{r}{} & \multicolumn{4}{c}{Normal distribution} & \multicolumn{4}{c}{$t_{10}$-distribution} \\
  $n$ & $15\%$ & $10\%$ & $5\%$ & $1\%$ & $15\%$ & $10\%$ & $5\%$ & $1\%$ \\
  \hline
  250 & 0.102 & 0.072 & 0.032 & 0.005 & 0.114 & 0.070 & 0.032 & 0.004 \\ 
  500 & 0.132 & 0.086 & 0.050 & 0.010 & 0.134 & 0.091 & 0.047 & 0.012 \\ 
  1000 & 0.153 & 0.100 & 0.055 & 0.010 & 0.146 & 0.103 & 0.052 & 0.011 \\ 
  2000 & 0.152 & 0.102 & 0.047 & 0.014 & 0.139 & 0.096 & 0.046 & 0.012 \\ 
  4000 & 0.151 & 0.102 & 0.050 & 0.008 & 0.169 & 0.111 & 0.054 & 0.006 \\ 
  8000 & 0.165 & 0.108 & 0.054 & 0.008 & 0.162 & 0.103 & 0.042 & 0.007 \\ 
 \hline 
\end{tabular}
\caption{\footnotesize Empirical size of the asymptotic test under estimated sparsity, at nominal levels $\alpha\in\{15\%,10\%,5\%,1\%\}$ (columns), for standard normal and $t_{10}$ data (column blocks) and increasing sample size $n$ (rows). Critical values are taken from the maximum of the approximating Brownian bridges, the sparsity $f(Q(\tau))$ is estimated by kernel density estimation, and $\beta=2/3$. Results are averaged over $2000$ Monte Carlo replications.}
\label{table:simulation_kde}
\end{table}

\begin{figure}[H]
        \centering
        \begin{subfigure}
            \centering
            \includegraphics[width=0.45\textwidth]{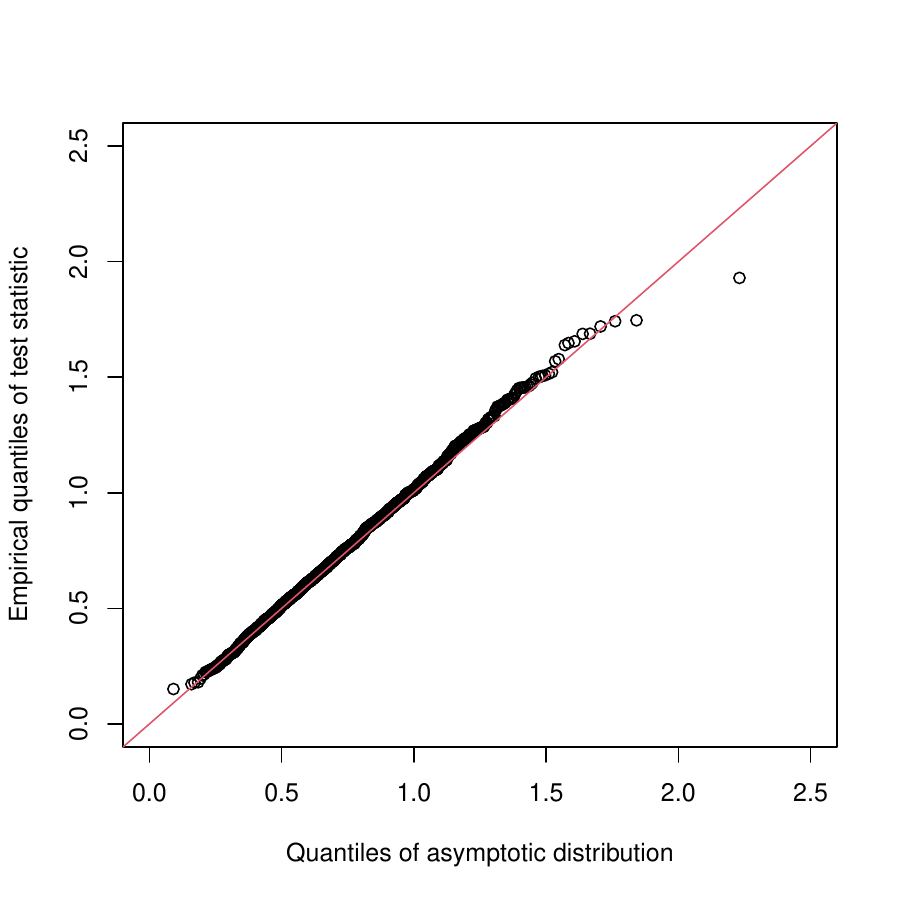}
        \end{subfigure}
        \hfill
        \begin{subfigure}
            \centering
            \includegraphics[width=0.45\textwidth]{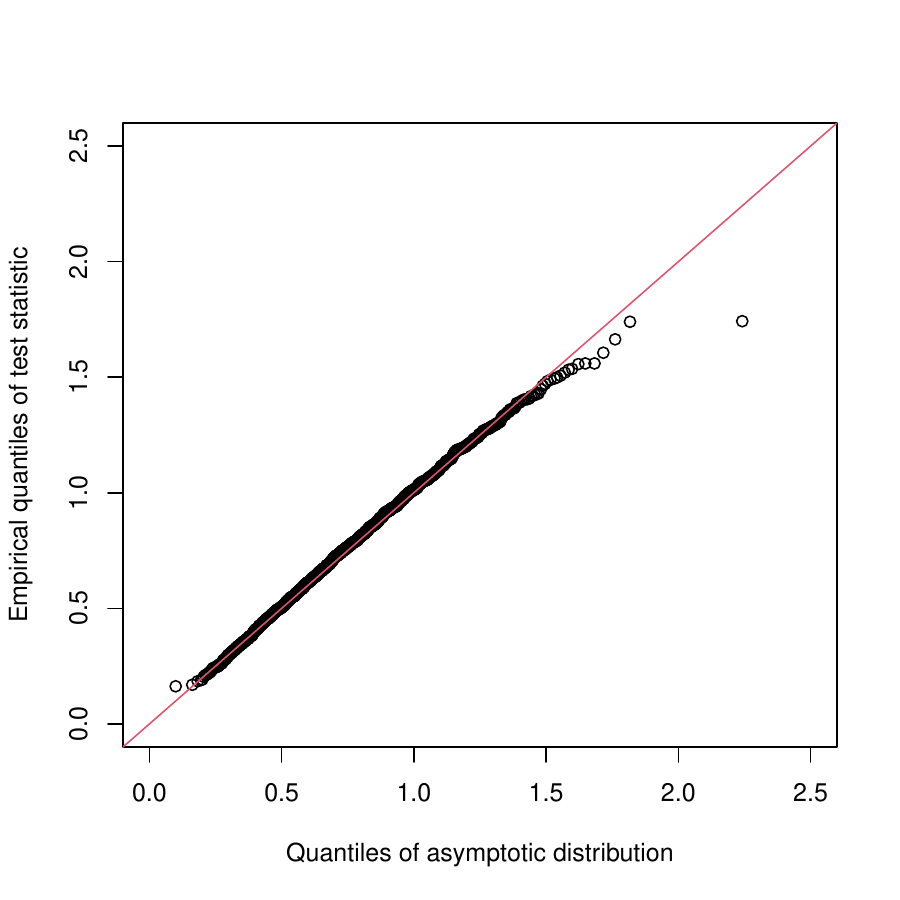}
        \end{subfigure}
    \caption{\footnotesize QQ plots of the test statistic against its asymptotic reference distribution (the maximum of Brownian bridges) under known sparsity, for standard normal data (left panel) and $t_{10}$ data (right panel).}
        \label{figure:qq}
\end{figure}

\begin{table}[H]
\footnotesize
\centering
\begin{tabular}{r|rrrr|rrrr|rrrr}
	\hline
  \multicolumn{1}{r}{} & \multicolumn{4}{c}{$\beta=1/2$} & \multicolumn{4}{c}{$\beta=2/3$} & \multicolumn{4}{c}{$\beta=3/4$} \\
  $n$ & $15\%$ & $10\%$ & $5\%$ & $1\%$ & $15\%$ & $10\%$ & $5\%$ & $1\%$ & $15\%$ & $10\%$ & $5\%$ & $1\%$ \\
  \hline
  250 & 0.167 & 0.117 & 0.060 & 0.016 & 0.184 & 0.132 & 0.073 & 0.016 & 0.203 & 0.144 & 0.080 & 0.018 \\ 
  500 & 0.174 & 0.117 & 0.056 & 0.011 & 0.175 & 0.121 & 0.060 & 0.014 & 0.210 & 0.148 & 0.078 & 0.016 \\ 
  1000 & 0.170 & 0.118 & 0.064 & 0.017 & 0.166 & 0.110 & 0.060 & 0.014 & 0.201 & 0.144 & 0.072 & 0.018 \\ 
  2000 & 0.192 & 0.122 & 0.062 & 0.015 & 0.152 & 0.104 & 0.051 & 0.011 & 0.190 & 0.130 & 0.064 & 0.014 \\ 
  4000 & 0.207 & 0.141 & 0.074 & 0.015 & 0.164 & 0.113 & 0.056 & 0.018 & 0.188 & 0.129 & 0.066 & 0.013 \\ 
  8000 & 0.187 & 0.125 & 0.064 & 0.011 & 0.173 & 0.112 & 0.051 & 0.010 & 0.170 & 0.124 & 0.064 & 0.018 \\ 
 \hline
\end{tabular}
\caption{\footnotesize Empirical size of the asymptotic test for $t_{10}$ data, at nominal levels $\alpha\in\{15\%,10\%,5\%,1\%\}$ (columns), for learning rates $\beta\in\{1/2,2/3,3/4\}$ and increasing sample size $n$ (rows). Critical values are taken from the maximum of the approximating Brownian bridges, and the sparsity $f(Q(\tau))$ is treated as known. Results are averaged over $2000$ Monte Carlo replications.}
\label{table:simulation_t}
\end{table}

\begin{figure}[tbp]
        \centering
        \vskip\baselineskip
        \begin{subfigure}
            \centering
            \includegraphics[width=0.45\textwidth]{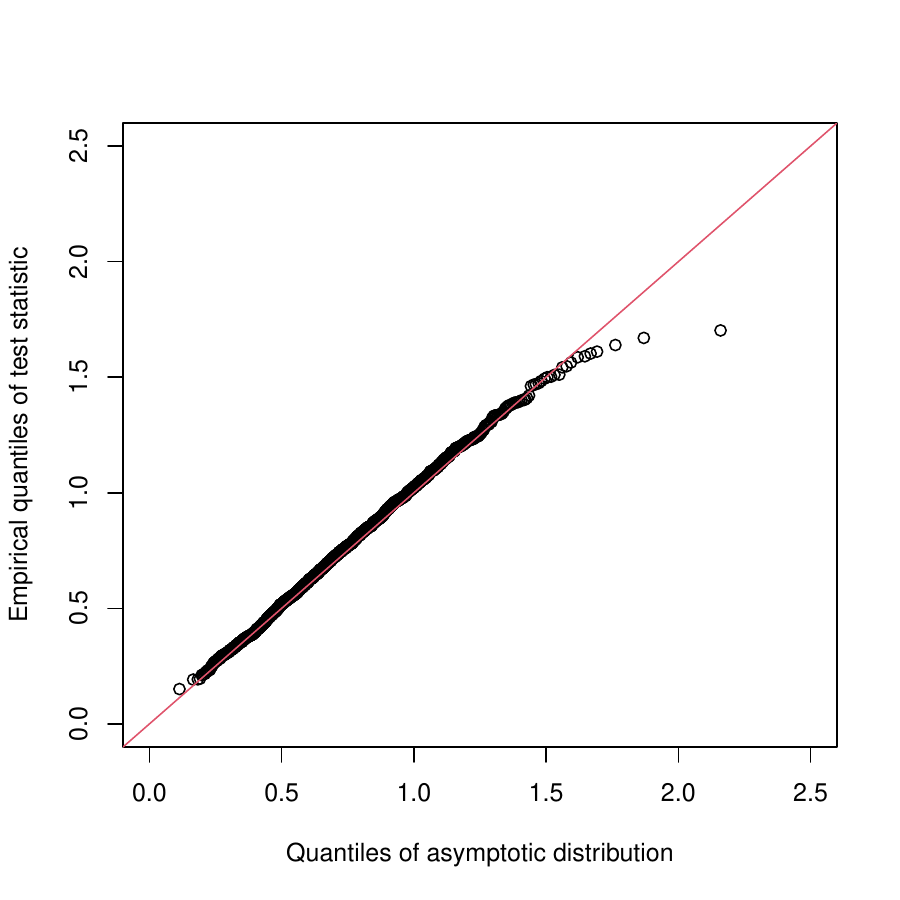}
        \end{subfigure}
        \hfill
        \begin{subfigure}
            \centering
            \includegraphics[width=0.45\textwidth]{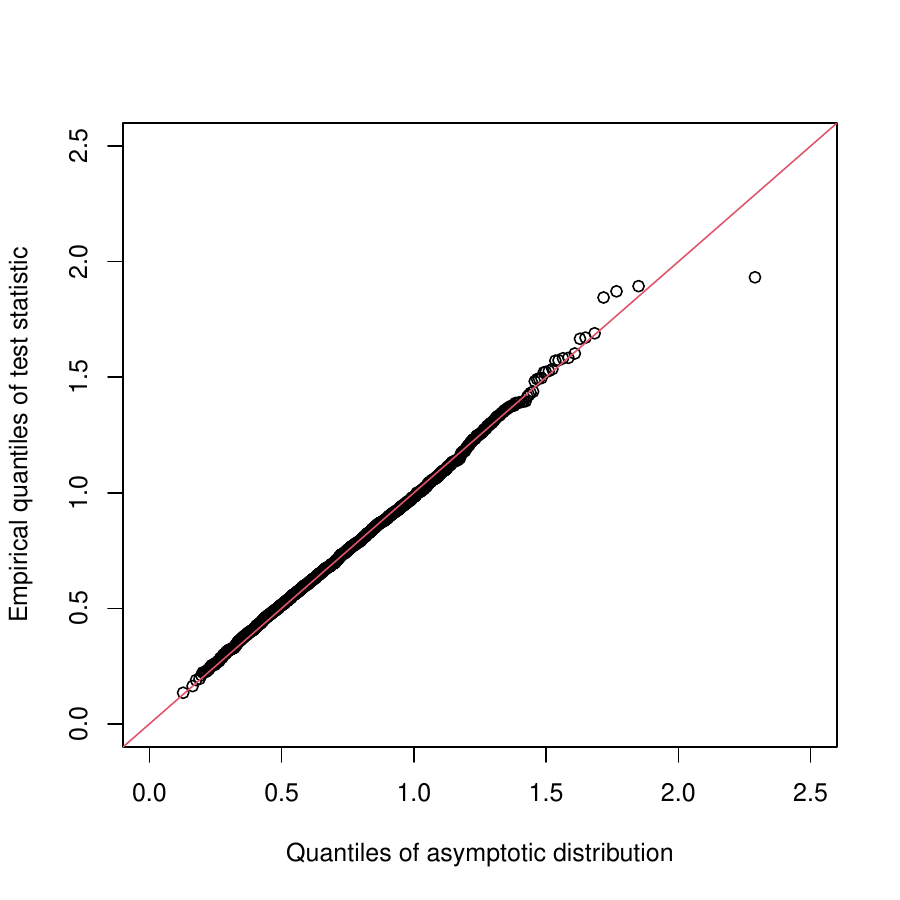}
        \end{subfigure}
        \caption{\footnotesize QQ plots of the test statistic against its asymptotic reference distribution (the maximum of Brownian bridges) under estimated sparsity, for standard normal data (left panel) and $t_{10}$ data (right panel).}
        \label{figure:qq2}
\end{figure}

\subsubsection{Conditional quantile simulation}

\begin{figure}[H]
        \centering
        \begin{subfigure}
            \centering
            \includegraphics[width=0.45\textwidth]{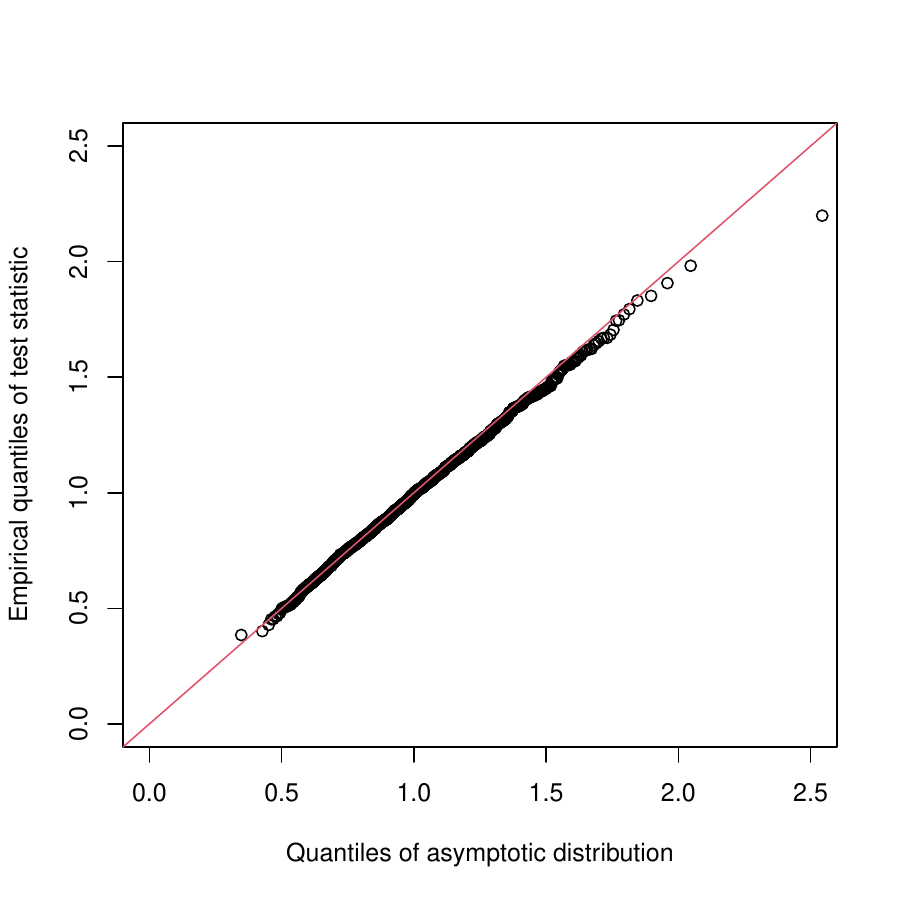}
        \end{subfigure}
        \hfill
        \begin{subfigure}
            \centering
            \includegraphics[width=0.45\textwidth]{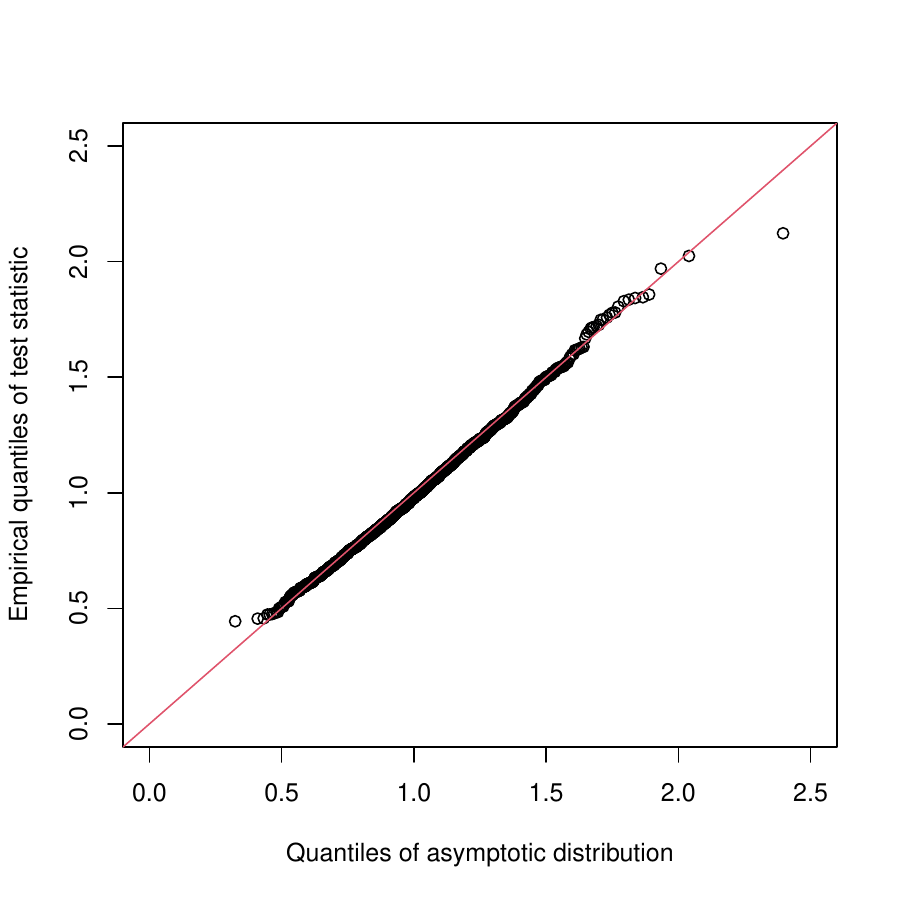}
        \end{subfigure}
        \caption{\footnotesize QQ plots of the test statistic against its asymptotic reference distribution for nonparametric conditional quantile estimation under estimated sparsity, for sample sizes $n=5000$ (left panel) and $n=10000$ (right panel).}
        \label{figure:cond}
\end{figure}

We now turn to the conditional quantile case, to confirm whether the inference procedure from Section~\ref{sec:conditional} delivers reliable coverage. The regressor $Z$ is uniformly distributed on
$[0,1]$ in our simulation setting. The conditional distribution of the dependent variable is
$Y|Z=z\sim N(0,z)$. We consider a grid of quantile levels,
$\tau\in\{0.1,0.2,\ldots,0.9\}$, and grid points $z_i\in\{0.2,0.4,0.6,0.8\}$. As
before, the learning rate parameter $\beta$ is set to $2/3$. The bandwidth is
selected by $h_n=0.7\,n^{-1/4}$ and $K(\cdot)$ is the uniform kernel, so that
$nh_n\to\infty$, $nh_n^5\to0$ and $2h_n<\min_{i\neq j}|z_i-z_j|$ at both sample
sizes considered. The QQ plots in Figure~\ref{figure:cond} show that for
large sample sizes we can achieve an excellent fit, indicating that the Gaussian
approximation remains accurate in the conditional setting as well.

\subsection{Additional Illustration Results}

\begin{figure}[tbp]
        \centering
            \includegraphics[width=0.8\textwidth]{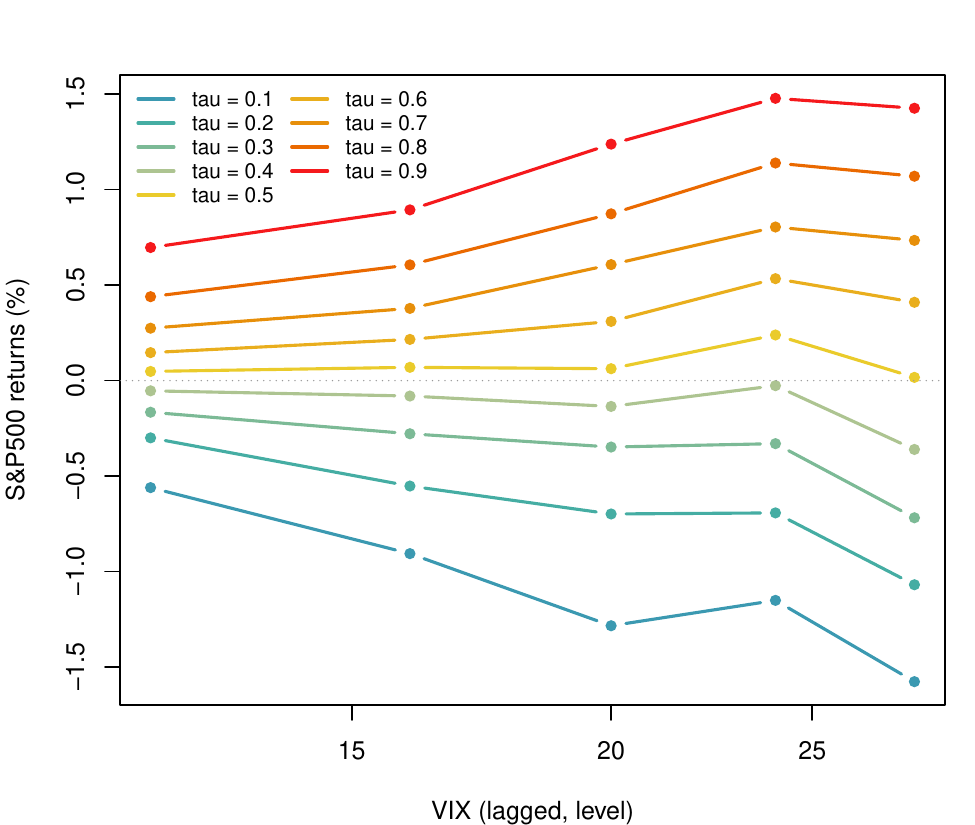}
        \caption{\footnotesize Estimated conditional quantiles of S\&P500 returns for different quantile levels.}
        \label{figure:cond_tau}
\end{figure}

\begin{figure}[tbp]
        \centering
            \includegraphics{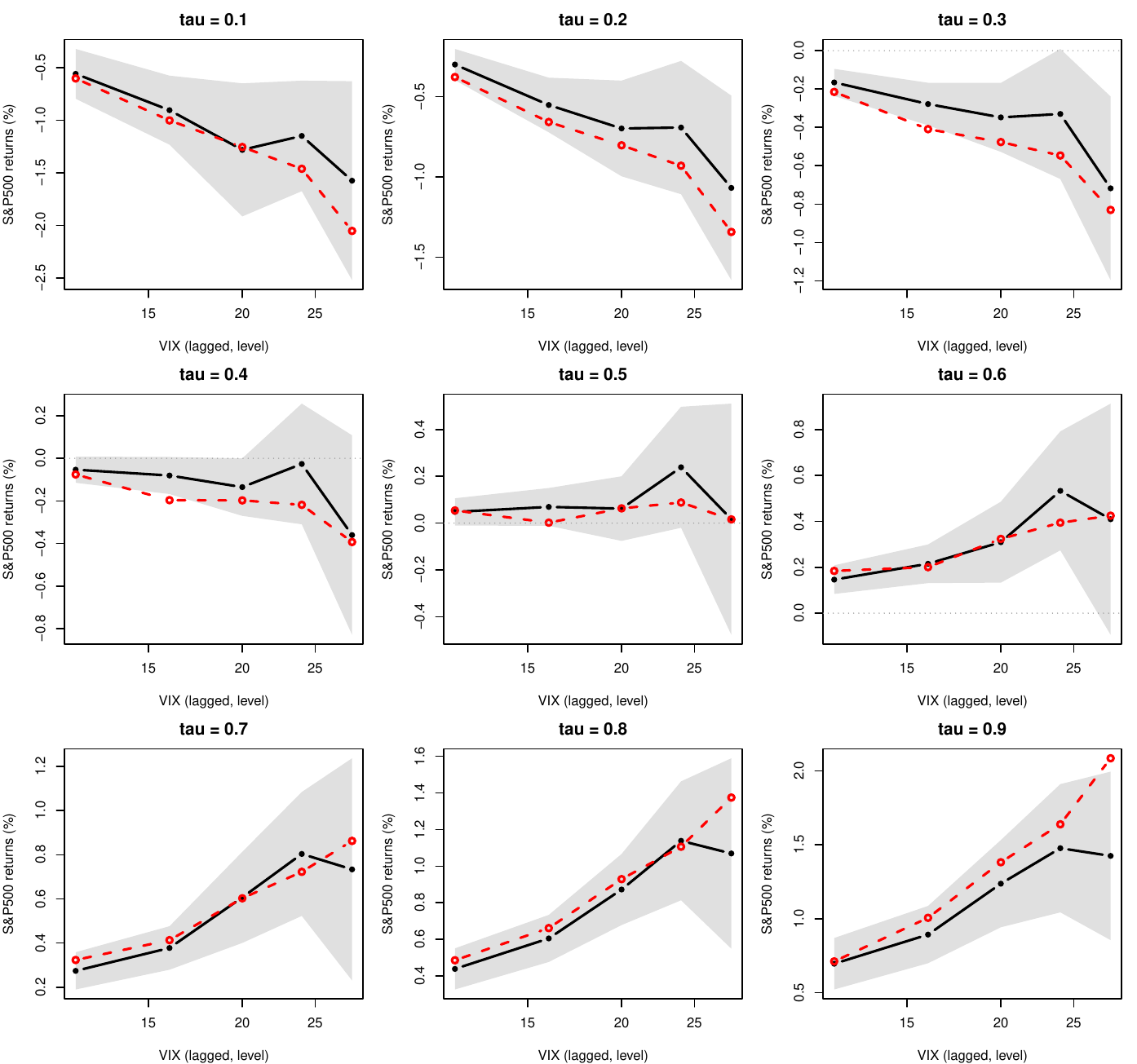}
        \caption{\footnotesize Estimated conditional quantiles of the S\&P500 returns (solid black line) with $95\%$-confidence bands for given quantile level $\tau$, and estimated location-scale model (dashed red line).}
        \label{figure:cond_tau_bands}
\end{figure}

\begin{figure}[tbp]
        \centering
            \includegraphics{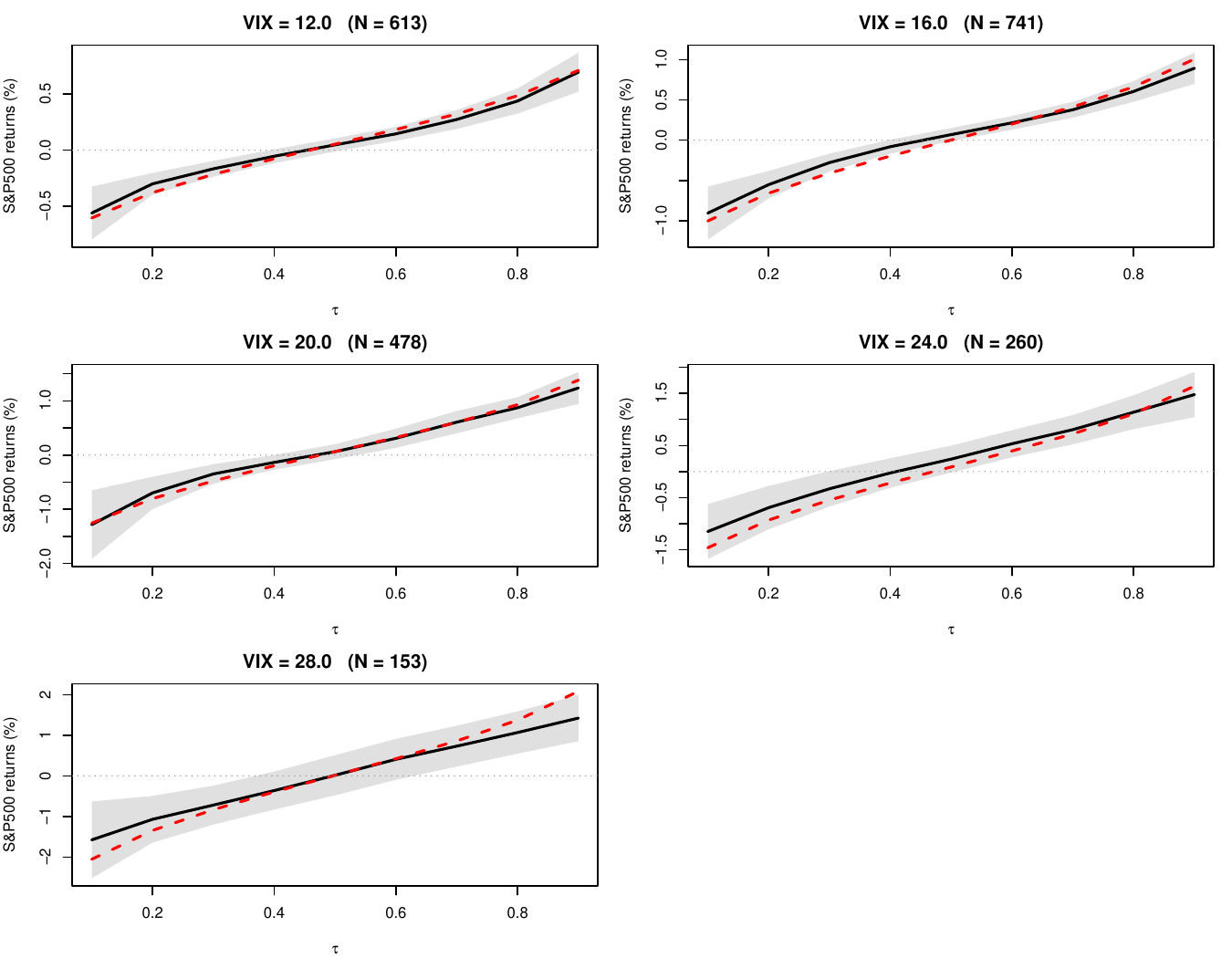}
        \caption{\footnotesize Estimated conditional quantiles of the S\&P500 returns (solid black line) with $95\%$-confidence bands for a given grid point $z_i$, and estimated location-scale model (dashed red line).}
        \label{figure:cond_vix_bands}
\end{figure}

Figure~\ref{figure:cond_tau} shows the estimated conditional quantiles for $\tau\in\G$ in one plot and demonstrates the curves do not cross. Figure~\ref{figure:cond_tau_bands} shows the estimated conditional quantiles for a fixed quantile level as a function of the volatility index. Figure~\ref{figure:cond_vix_bands} does the same for fixed volatility levels as a function of the quantile level. Table~\ref{table:deviations} presents the studentized deviations from the null.

\begin{table}[tbp]
\centering 
\begin{tabular}{lrrrrrrrrr}
  \toprule
 & 0.1 & 0.2 & 0.3 & 0.4 & 0.5 & 0.6 & 0.7 & 0.8 & 0.9 \\ 
  \midrule
12.0 & 0.266 & 1.160 & 1.009 & 0.525 & 0.155 & 0.877 & 0.842 & 0.607 & 0.131 \\ 
  16.0 & 0.428 & 0.899 & 1.704 & 1.920 & 1.203 & 0.252 & 0.519 & 0.634 & 0.846 \\ 
  20.0 & 0.062 & 0.509 & 1.033 & 0.666 & 0.010 & 0.117 & 0.028 & 0.422 & 0.709 \\ 
  24.0 & 0.859 & 0.827 & 0.921 & 0.978 & 0.841 & 0.772 & 0.417 & 0.143 & 0.538 \\ 
  28.0 & 0.729 & 0.689 & 0.340 & 0.100 & 0.002 & 0.044 & 0.370 & 0.848 & 1.673 \\  
   \bottomrule
\end{tabular}
\caption{Studentized deviations $\sqrt{N_i}\,\hat d_i(\tau)\,|\bar q_i(\tau)-q_0(z_i,\tau)|$ from the Gaussian null, by lagged VIX level (rows) and quantile level (columns). The maximum is $T_n = 1.920$; the critical value at $5\%$ significance level is $1.444$.}
\label{table:deviations}
\end{table}

\end{document}